\documentclass[11pt]{amsart}
\usepackage{amsbsy,amssymb,amscd,amsfonts,latexsym,amstext,delarray,
amsmath,graphicx, synttree, xcolor} 
\usepackage{float}
\usepackage[caption = false]{subfig}
\usepackage{qtree}

\usepackage{wrapfig}
\usepackage{lscape}
\usepackage{rotating}

\usepackage{color}

\numberwithin{equation}{section}

\def\R{{\mathbb R}}

\title[Topological Analysis of Syntax]{Topological Analysis of Syntactic Structures}
\author{Alexander Port, Taelin Karidi, Matilde Marcolli}
\address{University of Southern California \\ USA}
\email{portam@usc.edu}
\address{California Institute of Technology \\ USA}
\email{tkaridi@caltech.edu}
\address{California Institute of Technology \\ USA \newline \indent
Perimeter Institute for Theoretical Physics \\ Canada \newline \indent
University of Toronto \\ Canada}
\email{matilde@caltech.edu}
\date{}

\begin{document}
\maketitle

\begin{abstract}
We use the persistent homology method of topological data analysis and dimensional
analysis techniques to study data of syntactic structures of world languages. We analyze
relations between syntactic parameters in terms of dimensionality, of hierarchical clustering
structures, and of non-trivial loops. We show there are relations that hold across language
families and additional relations that are family-specific. We then analyze the trees describing
the merging structure of persistent connected components for languages in different language
families and we show that they partly correlate to historical phylogenetic trees but with
significant differences. We also show the existence of interesting non-trivial persistent
first homology groups in various language families. We give examples where explicit
generators for the persistent first homology can be identified, some of which appear to
correspond to homoplasy phenomena, while others may have an explanation in terms
of historical linguistics, corresponding to known cases of syntactic borrowing across
different language subfamilies.  
\end{abstract}

\section{Introduction}

The use of methods of Topological Data Analysis in linguistics was
introduced in \cite{Port}, where it was shown that the persistent homology
of the SSWL data of syntactic structures of world languages reveals the
presence of additional structures in the data, such as persistent first homology, 
that are not identifiable via other more traditional methods of computational 
measurements of language relatedness. 

\smallskip

In the present paper we carry out a much more in depth analysis, using
topological methods, of the data of syntactic structures (both from the SSWL 
database \cite{SSWL} and from the data of syntactic parameters collected 
by Longobardi and collaborators of the LanGeLin project, \cite{Longo1}, \cite{Longo2}).  

\smallskip

When we look at the data points as syntactic features or syntactic parameters, with coordinates
given by the values of the parameter over a given set of languages, we focus on the question of identifying
relations between these syntactic variables. This is a main open question already investigated by
other methods in \cite{Kazakov}, \cite{OBM}, \cite{ParkMa}, \cite{ShuMar}. We analyze the
clustering structure between syntactic parameters by analyzing the persistent connected
components of the data at various scales and the resulting tree that follows the order in which
the components merge as the scale parameter increases. We compare the detected cluster structure
obtained in this way with those discussed in \cite{Kazakov} and \cite{OBM}. We also compute the
persistent $H_1$ and we show that there are further relations between syntactic parameters corresponding
to non-trivial persistent $H_1$-generators that are not detectable
by cluster information. Moreover, we also compute estimates of
dimensionality of these data sets. We compare the dimensionality analysis
for balls and spheres of varying dimensions with the dimensionality 
analysis of our sets of data point of syntactic parameters and we identify peaks around the most likely
dimension estimates. We also perform dimension estimates for the data of
syntactic parameters with coordinates evaluated only over certain language families,  
in order to detect the presence of relations between syntactic parameters that may be language family
specific and not universal across families. We find that indeed the dimension drops when coordinates are
restricted to subfamilies indicating the presence of family-specific relations between the syntactic variables
in addition to universal ones. 

\smallskip

When we look at the data points as languages, with coordinates given by the
values of their syntactic parameters (LanGeLin data) or binary syntactic 
variables (SSWL), we focus on three main questions. The first question is to what extent the
persistent $H_0$ (the persistent connected components) of the data set
can be used as an alternative method for the reconstruction of phylogenetic
trees of language families. In \cite{Port}, based on some preliminary cases
of topological analysis of syntactic structures, we conjectured that the barcode
diagram of the persistent connected components could provide a reliable 
reconstruction of the phylogenetic tree of the languages involved. In the more
detailed analysis that we present here we show that, while this is indeed sometimes the
case, and the trees derived from the merging and clustering structure of the
persistent connected components correlate with the phylogenetic
trees of language families in terms of grouping together or languages by subfamilies, 
the information contained in the persistent components trees and in the phylogenetic trees is not
always identical. We will show examples where the structure of the tree
of the persistent connected components significantly differs from the phylogenetic tree 
while still retaining much of the information on the grouping into subfamilies. Investigating the discrepancies
between phylogenetic trees and persistent components trees will provide better
insight on what information about language relatedness is captured by the
persistent connected components that differs from historical family relatedness.
The main information carried by persistent components trees is a hierarchical organization of
the spreading of syntactic features across languages. 

\smallskip

\begin{figure}
	\subfloat[Austronesian]{\includegraphics[width = 4.5in]{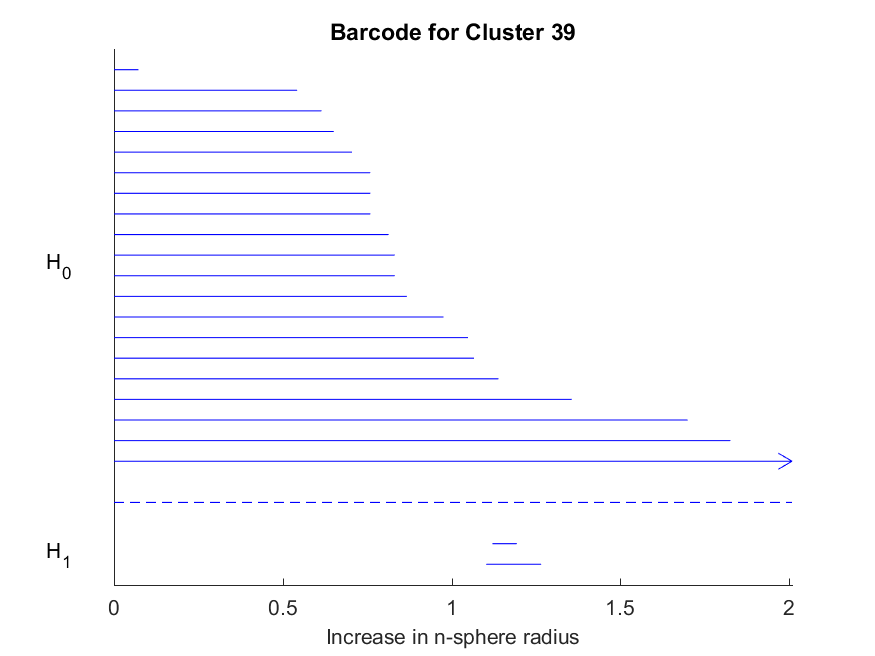}} \\
	\subfloat[Niger-Congo]{\includegraphics[width = 4.5in]{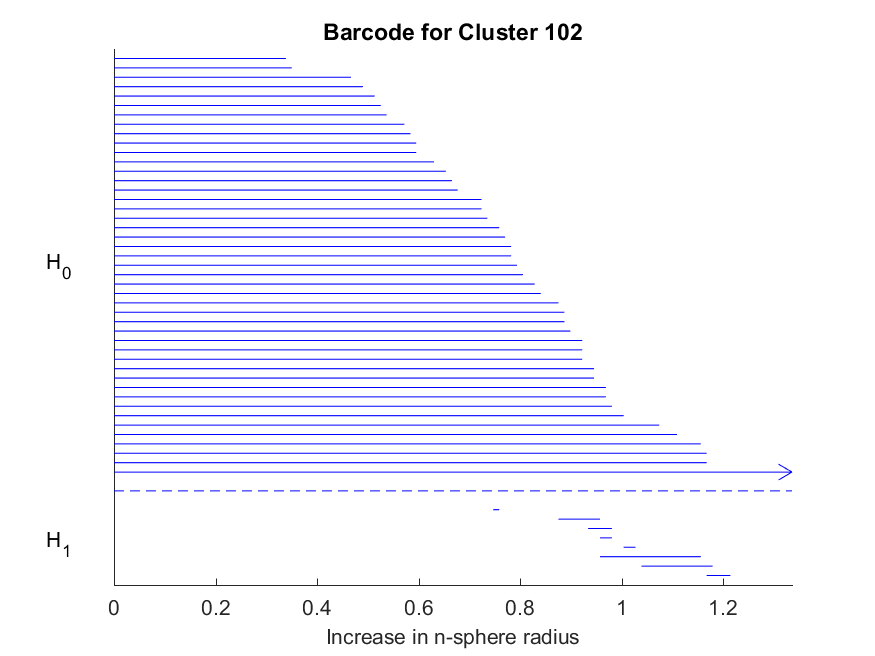}}\\
	
	\caption{Barcode graphs of the Austronesian and Niger-Congo families from the SSWL data set. 
	\label{barcodes_AC_Austro} }
\end{figure}

The second question we investigate is the detection of higher dimensional
topological structures, in particular the persistent first homology group $H_1$,
and the meaning of the resulting structures from the point of view of
historical linguistics. Unlike typical random simplicial complexes, the Vietoris-Rips
complexes at varying scales associated to data of syntactic parameters tend to
exhibit no higher dimensional homology, that is, no non-trivial persistent
generators of the $H_k$ homology groups for $k\geq 2$. This seems to indicate
that passing from trees to networks given by more general graphs (to account
for the presence of non-trivial persistent $H_1$ homology) suffices to describe
relatedness between languages, without the need to introduce higher dimensional 
geometries. The structure of the persistent $H_1$ that we see in the case of
syntactic data also differs from random simplicial sets in the fact that non-trivial
$H_1$-generators appear only in the larger clusters, and there is less overlap
between them in the barcode diagram than expected in a random setting.
As we will show in specific examples, non-trivial generators of the
persistent $H_1$ homology can sometimes consist of a set of languages not
belonging to the same subfamilies, but for which it is known that there have
been historical interactions and possible influences at the syntactic level
(for example between the Hellenic and the Slavic languages). In other cases
the set of languages that provide explicit $H_1$-generators do not seem to
be plausibly related through historical influence that could lead to 
borrowing at the syntactic level.  There are two possible explanation for
this second type of $H_1$-generator. Either the $H_1$-generator is not
the historically relevant one and a homologous one (differing by a boundary
and still generating the same $H_1$-class) would be the one that can
be interpreted in historical linguistic terms, or else the $H_1$-generator is
only detecting homoplasy phenomena in syntax. In phylogenetics 
homoplasy refers to those traits that are independently gained in separate
branches of the phylogenetic tree and are not due to a common ancestor.
Homoplasy phenomena in syntax are observed when languages that are
not closely related exhibit syntactic similarities. When such phenomena
occur one can expect that an $H_1$-generator may appear in the topological
analysis where the languages involved have no clear historical record of
mutual influence involving the possibility of syntactic borrowing. We will
analyze examples of $H_1$-generators in a case 
for which we can propose at least a conjectural historical linguistic
explanation and in a case that seems to be due exclusively to homoplasy. 
While the existence of non-trivial $H_1$-generators was already observed
in \cite{Port}, in the analysis of Indo-European language family data from
the SSWL syntactic features, in this paper we carry out a much more
extensive analysis throughout all the clusters and four different main
language families in the SSWL database, as well as for the LanGeLin
data, hence we are able to identify several more cases of non-trivial 
$H_1$-generators, also outside of the Indo-European case (for example
we show there are clusters where the Niger-Congo languages also exhibit
persistent $H_1$-generators, see Figure~\ref{barcodes_AC_Austro}, while those
sub-clusters of the Niger-Congo family that we had previously analyzed in \cite{Port}
showed no non-trivial $H_1$).

\smallskip

The third question is an estimation of dimensionality for different language
families as a measure of how spread out the syntactic features are across 
languages in a given family. We show that language families like the Niger-Congo
or the Indo-European family have an estimated dimensionality that significanty
exceeds the dimensionality computed over the entire database of languages,
while other language families like the Afroasiatic or Austronesian family have
an estimated dimensionality that is significantly smaller than that of the full set
of languages. We also consider the Ural-Altaic hypothesis from this point of
view of dimensionality estimates and we show that the combined estimate for
the two sets of languages shows two distinct peaks one closer to the peak for
the Altaic languages alone and one closer to that of the Uralic.  (This is
consistent with recent observations of a similar nature in \cite{LongoCeo}.)

\smallskip

We discuss also the use of the principal components in our data analysis, 
both in view of identifying possible linguistic interpretations for the different 
weights assigned to the syntactic parameters by the PCA method and in order
to understand the effect that the variance of the PCA can have on the details
of the reconstruction of the tree of the persistent components and the 
$H_1$-generators.

\smallskip

\begin{figure}
	\subfloat[cluster 105]{\includegraphics[width = 2.5in]{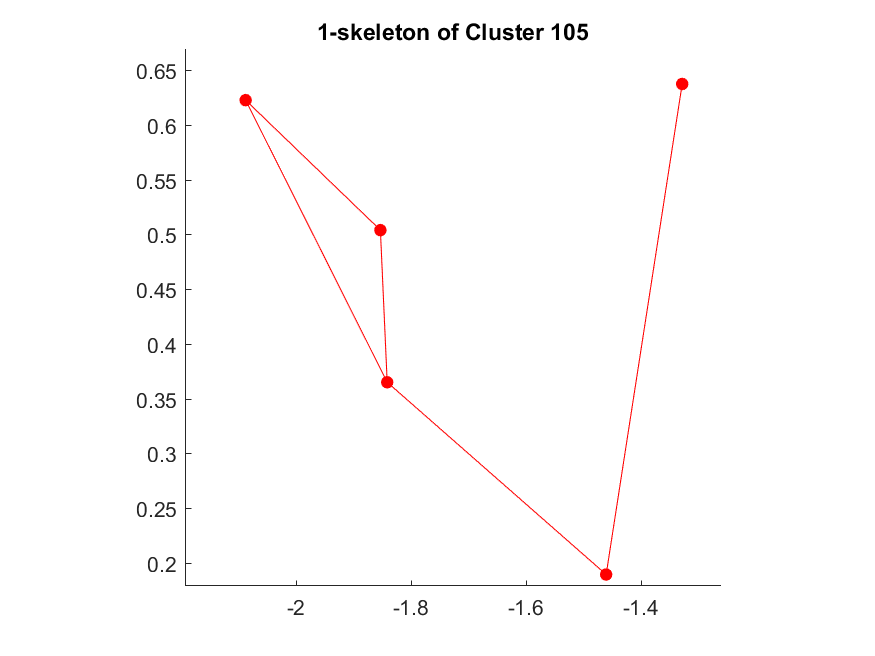}} 
	\subfloat[cluster 111]{\includegraphics[width = 2.5in]{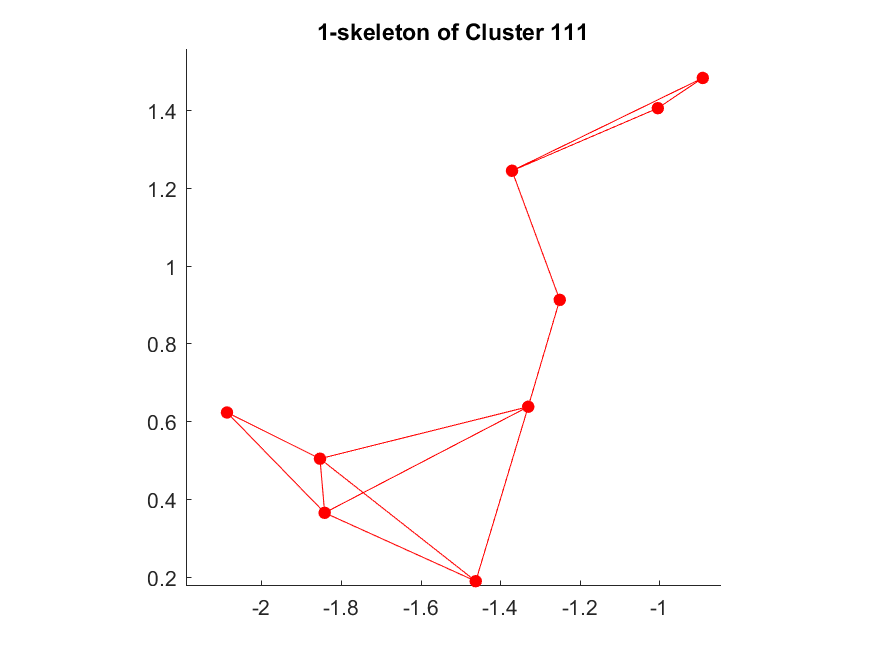}} \\
	\subfloat[cluster 116]{\includegraphics[width = 2.5in]{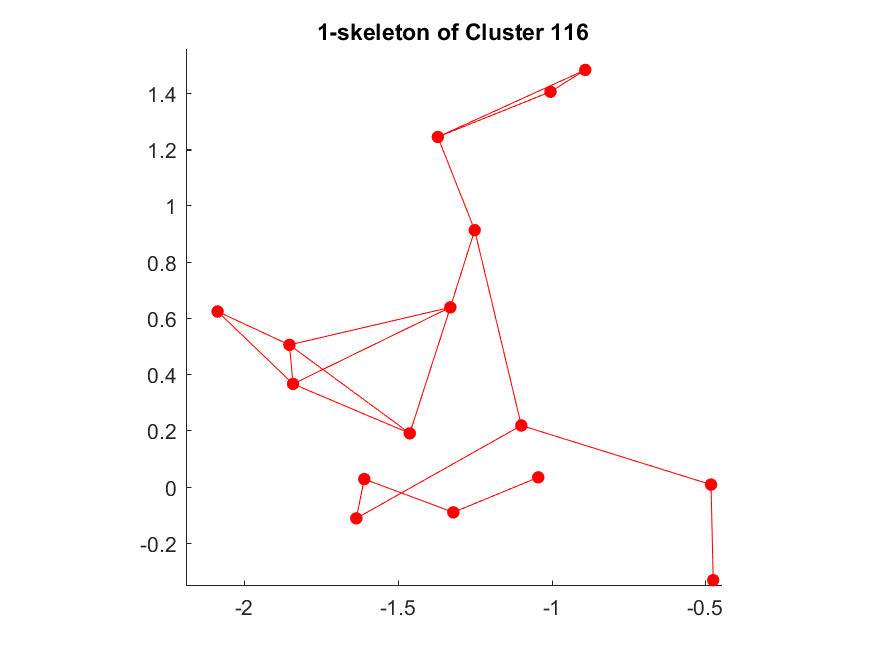}}
	\subfloat[cluster 124]{\includegraphics[width = 2.5in]{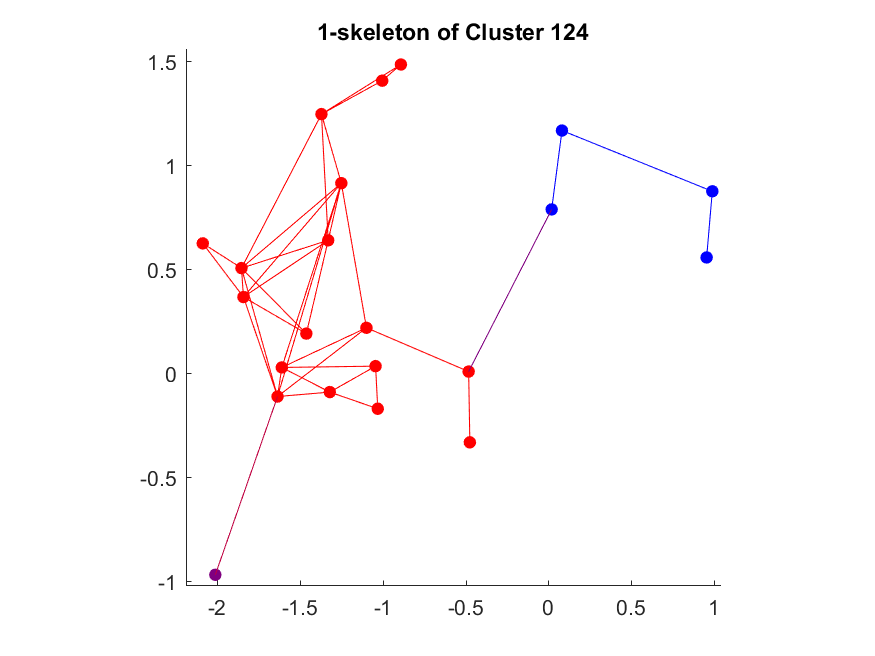}} \\
	\subfloat[cluster 138]{\includegraphics[width = 2.5in]{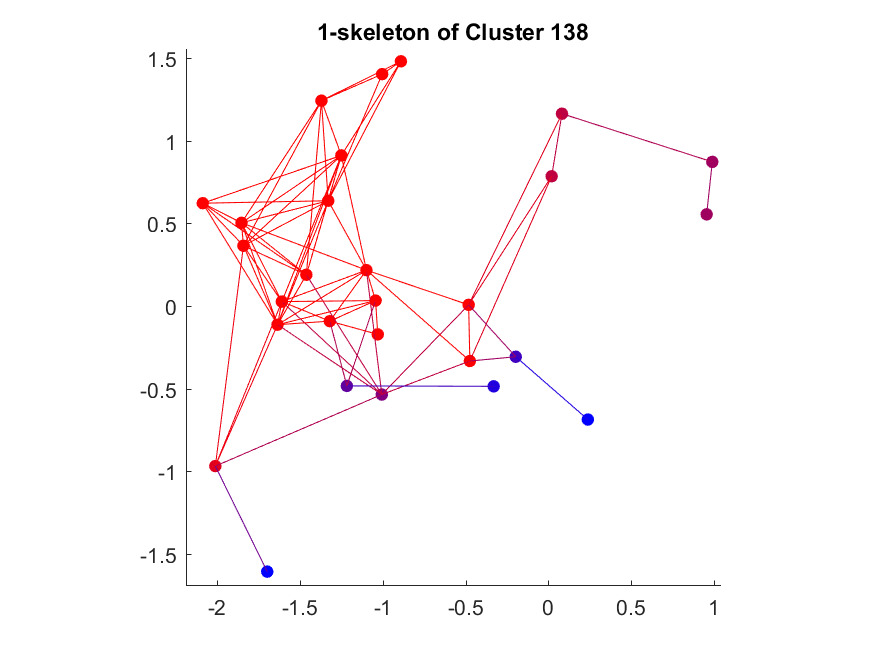}} 
	\subfloat[cluster 155]{\includegraphics[width = 2.5in]{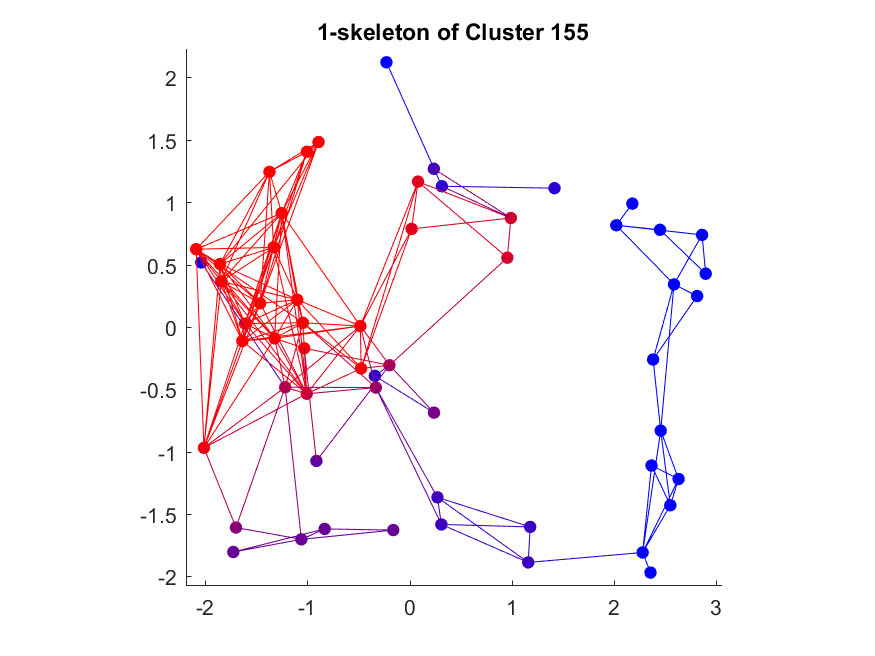}} \\
	\subfloat[cluster 167]{\includegraphics[width = 2.5in]{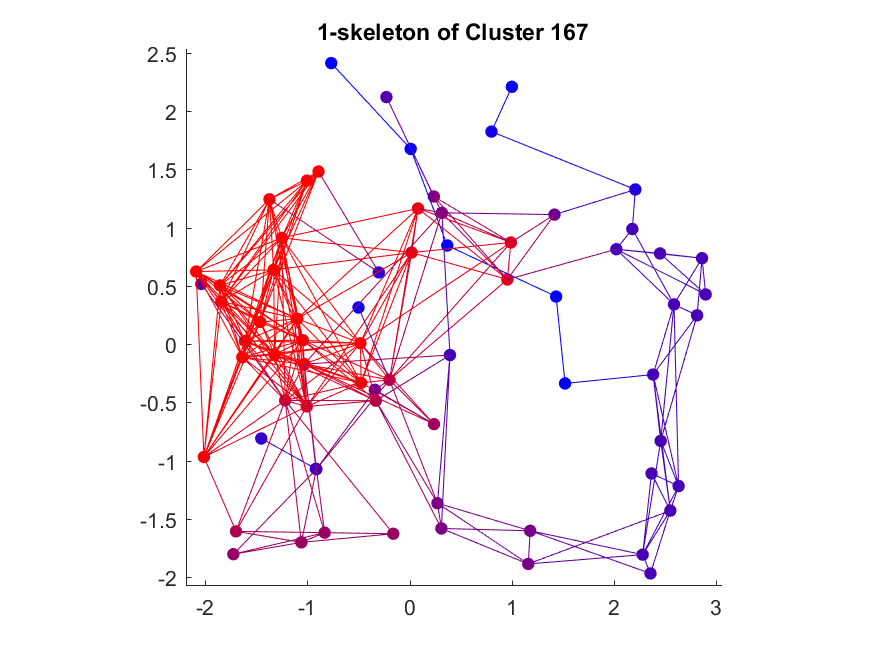}}
	\subfloat[cluster 177]{\includegraphics[width = 2.5in]{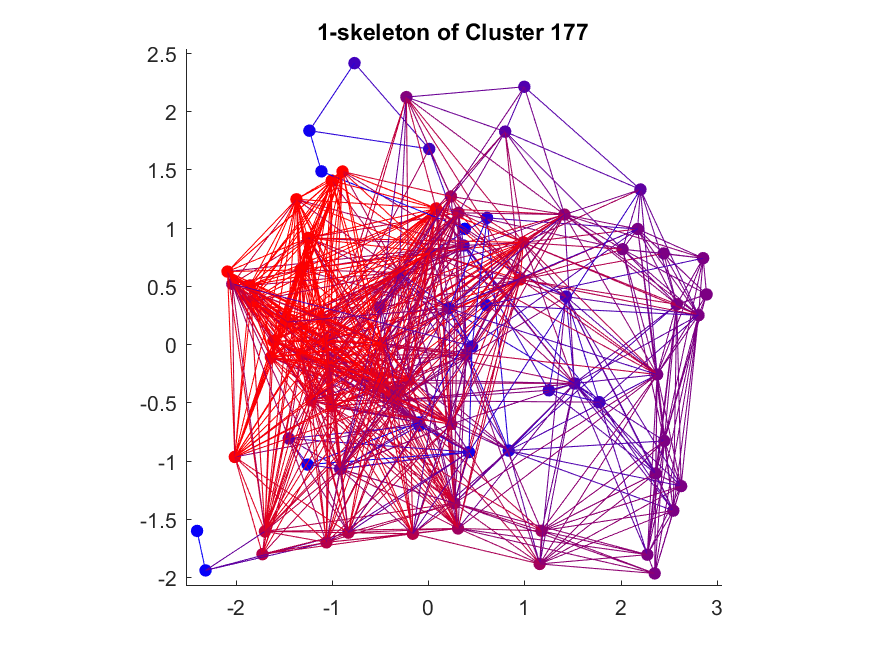}} 
	\caption{$1$-skeletons of the Indo-European family in varying radii, SSWL. \label{skeletons_SSWL}}
\end{figure}

%\begin{figure}
%	\subfloat[cluster 84]{\includegraphics[width = 2.5in]{1-skeleton_cluster_84.png}} 
%	\subfloat[cluster 105]{\includegraphics[width = 2.5in]{1-skeleton_cluster_105.png}} \\
%	\subfloat[cluster 109]{\includegraphics[width = 2.5in]{1-skeleton_cluster_109.png}}
%	\subfloat[cluster 121]{\includegraphics[width = 2.5in]{1-skeleton_cluster_121.png}} \\
%	\subfloat[cluster 133]{\includegraphics[width = 2.5in]{1-skeleton_cluster_133.png}} 
%	\subfloat[cluster 151]{\includegraphics[width = 2.5in]{1-skeleton_cluster_151.png}} \\
%	\subfloat[cluster 159]{\includegraphics[width = 2.5in]{1-skeleton_cluster_159.png}}
%	\subfloat[cluster 161]{\includegraphics[width = 2.5in]{1-skeleton_cluster_161.png}} 
%	\caption{$1$-skeletons of the Indo-European family in varying radii, SSWL. \label{skeletons_SSWL}}
%\end{figure}

\subsection{Persistent topology: a quick review}

In the field of data analysis researchers often come across very large data sets. 
The first question that arises when analyzing big data is how to make sense of it. 
More precisely, this generally means trying to identify certain lower dimensional
loci (manifolds or other kinds of geometric and topological objects) the data points 
lie on (or sufficiently near) inside a much higher dimensional ambient space. 
In other words, one wants to know what is the ``shape" of the data. Topology is 
the branch of mathematics that studies shapes, hence topological data analysis 
is especially suited for this task of analyzing high dimensional and complex data 
sets and identify the topological structures the data exhibit. In particular, persistent 
homology is a fundamental tool at the heart of topological data analysis, see
\cite{Carlsson} and also \cite{BoChaYv}, \cite{EdelHar}, \cite{Ghrist}, \cite{Zomo}.
Persistent homology gained much popularity in recent years as the primary method
of topological data analysis and found various applications, ranging from the analysis 
of protein structures in biology, to analyzing $3D$ images in image recognition, to 
computational neuroscience, and more. 
A more theoretical and categorical viewpoint on persistent homology was developed
in \cite{BuSco}, \cite{KaSha}, \cite{ManMar}. Barcode diagrams were also 
previously studied as `canonical forms' in \cite{Bara}.

\smallskip

More precisely, topological data analysis aims at extracting topological features 
from big data sets, by computing topological invariant of a corresponding space 
at various scales. The scale dependence of the topological structures can be
used to separate the meaningful characteristics of the underlying space, which are
assumed to be persistent over a larger range of scales, from effects caused by 
noise in the data, which are transitory and only appear within a small range of scales. 
In order to compute the homology groups at different scales one first constructs a
scale-dependent simplicial set or chain-complex from the data points, using a suitable 
distance function in a large ambient space where the data points are embedded. 
This construction converts the discrete data set into a global topological object. The
scale dependence gives rise to a family of simplicial sets organized as a filtration,
where each simplicial set at a given scale is nested into the ones at larger scales.
A sampling of this nested family of simplicial sets can be seen, in the case of the 
data set given by the SSWL database of syntactic parameters, in 
Figure~\ref{skeletons_SSWL}. In the family of simplicial sets so constructed, the
scale, which is also referred to as the \textit{proximity parameter}, ranges between
zero and the maximum of the distances between any two points in the data set. 
At intermediate scales, one considers those subsets of points with mutual distances
smaller than the given scale and joins them by simplexes. We review the construction
more in detail in Section~\ref{VRsec}. From the filtered simplicial complexes we then 
produce a \textit{barcode graph} by computing the homology groups at various scales. 
The barcode graph represents the persistent homology of a chain-complex uniquely,
see \cite{Carlsson}, \cite{Zomo}. Each horizontal line corresponds to a generator of 
a homology group, where its starting point is the scale of its birth and its ending point 
is the scale of its death (we will discuss this more in detail below). This barcode 
graph encodes the information regarding the persistent homology of the data and is 
beneficial in the task of distinguishing the significant features from the noise.

\subsubsection{The Vietoris-Rips Complex.} \label{VRsec}
Data sets are a discrete set of points and therefore have a trivial topology. In order to convert 
the data into a more interesting topological object, one first needs to construct a simplicial 
complex (in fact a scale-dependent family of simplicial complexes) from the data. Then, for each 
complex we can compute the corresponding homology groups. Loosely speaking, homology 
computes the number of $n$-dimensional holes in the shape of the data. 

\smallskip

Given a finite set of points $X \subset {\R}^N$ and a scale $\epsilon > 0$, the 
Vietoris-Rips complex $R_{\epsilon}(X)$ over a field $\mathbb{F}$ is an abstract 
simplicial complex whose space of $n$-simplices $\R_n(X, \epsilon)$ is the $\mathbb{F}$-vector space  
spanned by all the unordered $(n+1)$-tuples $\{x_0, \dots, x_n\}$ in the data set such that for every 
$0 \leq i \neq j \leq n$ we have $d(x_i, x_j ) \leq \epsilon$, where $d:\R^N \times \R^N \to \R$ is 
the Euclidean distance function.
We have a filtration, i.e.~a sequence of nested simplicial complexes with the inclusion maps $R_n(X,\epsilon) \to R_{n+1}(X, \epsilon)$. Moreover, we have the usual boundary maps $\partial_n : R_n(X,\epsilon) \to R_{n+1}(X, \epsilon)$ that satisfy the condition $\partial_n \circ \partial_{n+1} = 0$ for every $n > 0$. Therefore $\text{Im}(\partial_{n+1}) \subseteq \ker(\partial_n)$ so the quotient between $\ker(\partial_n)$ and $\text{Im}(\partial_{n+1})$ is well defined. Elements in $\ker({\partial_n})$ are called \textit{$n$-cycles} and elements in $\text{Im}(\partial_{n+1})$ are called \textit{$n$-boundaries}. Now we can define the $n$-th \textit{homology} group:  
$$H_n(X, \epsilon) :=  \ker(\partial_n) / \text{Im}(\partial_{n+1}) .$$

\smallskip

The dimension of the $n$-th homology, $\beta_n(X, \epsilon) := \dim(H_n(X, \epsilon)) = \dim(\ker(\partial_n) ) - \dim(\text{Im}(\partial_{n+1}))$ is called the $n$-th Betti number and counts the number of holes in $X$. For example, $\beta_0$ is the number of connected components, $\beta_1$ is the number of holes, and $\beta _2$ is the number of voids. To summarize, for every $\epsilon$ we can define a sequence of Vitories-Rips complexes, but one might ask himself what is a "good" scale $\epsilon$, or whether there is an ultimate $\epsilon$ in the way that it reveals significant features of the data. In view of this we will define the notion of persistent homology. 

\subsubsection{Persistent Homology} Let $X \subseteq {\R}^n$, $\epsilon_1 < \epsilon_2 < \dots < \epsilon_n$ and $R(X, \epsilon_1) \subseteq \dots \subseteq R(X, \epsilon_n) = R(X)$ the corresponding filtered Vietoris-Rips simplicial complex. Note that, by functoriality of homology, for any $ 1 \leq k \leq n$ the inclusion 
maps $R(X, \epsilon_i) \hookrightarrow R(X, \epsilon_j)$ induces linear maps $h_{i,j}: H_k(X, \epsilon_i) \to H_k(X, \epsilon_j)$ for every $i < j$. The  \textit{$k$-th persistent homology} of 
$R(X)$ is the pair $(H_k(X, \epsilon_i)_{1 \leq i \leq n}, (h_{i,j})_{1 \leq i \leq j \leq n})$. 

\smallskip 

By computing the homology as the value of the scale $\epsilon$ increases, the persistent homology detects which topological features of the data (i.e non trivial generators of the homology groups) persist over relatively 
long intervals and therefore are significant. 

\smallskip

One way to formalize this property is by considering the range of the maps 
$h_{\epsilon,\epsilon'}: H_k(X, \epsilon) \to H_k(X, \epsilon')$ for $\epsilon< \epsilon'$,
which detects the part of the homology $H_k(X, \epsilon)$ at scale $\epsilon$ that
persists at scale $\epsilon'$. By adjusting the length of the persistence interval $[\epsilon, \epsilon']$
one can distinguish between the part of the homology that persists for a longer range of
scales (structure) from the part that disappears within a very short range of scales (noise). 

\smallskip

There is a visual way to encode this information, called a \textit{barcode graph} (see Figure~\ref{barcodes_AC_Austro} for examples taken from the SSWL data of syntactic structures). For every $1 \leq k \leq n$ the number of lines in the graph is the Betti number $\beta_k$, where each line corresponds to a non trivial generator of $H_k(X)$. There is a line between filtration step $i$ and filtration step $i+1$ if a generator of $H_k(X, \epsilon_i)$ is mapped to a non trivial generator of $H_k(X, \epsilon_{i+1})$, under the map $h_{i,i+1}$ defined earlier. If a non-trivial generator of $H_k(X, \epsilon_i)$ is sent to zero in $H_k(X, \epsilon_{i+1})$, the line ends at $\epsilon_{i+1}$.

\medskip

\subsection{Syntactic structures of world languages} \label{SyntaxSec}

The data we are interested in analyzing encode syntactic structures of different world languages. 
A fundamental idea in modern linguistics, developed within the setting of Chomsky's Principles and
Parameters model \cite{Chomsky}, \cite{ChoLa} postulates that the syntax of any human language 
can be encoded in a universal set of binary variables, or syntactic parameters. A general introduction
to syntactic parameters aimed at readers without a linguistics background can be found in \cite{Baker}.
The main idea is that the syntactic parameters should provide coordinates for the space of 
possible human languages. The values for the set of syntactic parameters, each viewed as a 
yes/no answer to whether a certain syntactic construction is possible in a given language, 
distinguish between different languages. For an overview of the recent state of the art in
linguistic research on syntactic parameters, see the collection of papers in the volume \cite{LingAn}.

\smallskip

Open questions regarding syntactic parameters
that are especially suitable for a mathematical approach include:
\begin{itemize}
\item identifying dependencies between parameters, in other words
identifying the lower dimensional locus inside the binary space of all
possible parameter assignments that is occupied by actual human languages;
\item identifying relatedness between languages on the basis of the distribution of
their syntactic features. 
\end{itemize}
With respect to the second question, the use of syntactic parameters
as a method for computational reconstruction of phylogenetic trees of
language families was developed in \cite{Longo3}, \cite{Longo2}, \cite{Longo1},
\cite{LongoGua2} and recently also considered from the perspective of
phylogenetic algebraic geometry in \cite{OSBM} and from the topological
perspective we consider here in \cite{Port}. The first question has been
approached with mathematical methods in \cite{Kazakov}, 
\cite{Mar}, \cite{OBM}, \cite{ParkMa}, \cite{ShuMar}, \cite{SiTaMa}. 

\smallskip

\subsubsection{Databases of syntactic structures} The main sources
currently available that record binary data of syntactic structures of
languages are the SSWL database \cite{SSWL} (Syntactic Structures
of World Languages) and the data of Longobardi and the LanGeLin
collaboration, \cite{Longo1}, \cite{Longo2}. 

\smallskip

Several of the binary variables recorded in the SSWL database should 
not be regarded as genuine ``syntactic parameters"
in the sense that linguists define, although they still encode useful
information about syntactic structures. 
Thus, in the following, since we will include all these variables in our analysis, 
we will refer more loosely to ``syntactic structures" or ``binary syntactic variables" 
instead of using the more specific term  ``syntactic parameters". 

\smallskip

The $116$ binary variables recorded in the SSWL database include:
\begin{itemize}
\item variables describing word order properties, 
from {\em 01}--Subject Verb to {\em 22}--Noun Pronomial Possessor
\item variables {\em A01--A04} describing relations of adjectives to nouns and degree words
\item variable {\em AuxSel01} about the selection of auxiliary verbs
\item variables {\em C01--C04} related to word order properties of complementarizer 
and clause and adverbial subordinator and clause
\item variables {\em N201--N211} on properties of numerals
\item variables  {\em Neg01--Neg14} on negation
\item variables {\em OrderN301--OrderN312} on word order properties involving 
demostratives, adjectives, nouns, and numerals
\item variables {\em Q01--Q15} regarding the structure of questions
\item variables {\em Q16Nega--Q18Nega} and
{\em Q19NegQ--Q22NegQ} on answers to negative questions
\item variables {\em V201-V202} on declarative and interrogative Verb-Second
\item variables {\em w01a--w01c} on indefinite mass nouns in object position 
\item variables {\em w02a--w02c} on definite mass nouns in object position
\item variables {\em w03a--w03d} on indefinite singular count nouns in object position
\item variables {\em w04a--w04c} on definite singular count nouns in object position
\item variables {\em w05a--w05c} on indefinite
plural count nouns in object position
\item variables {\em w06a--w06c} on definite plural count nouns in object position
\item variables {\em w06a--w06c} on definite plural count nouns in object position
\item variables {\em w07a--w07d} on nouns with (intrinsically) unique referents in object position
\item variables {\em w08a--w08d} on proper names in object position
\item variables {\em w09a--w09b} on order of article and proper names in object position
\item variables {\em w10a--w10c} on proper names modified by an adjective in object position
\item variables {\em w11a--w11b} on order of proper names and adjectives in object position
\item variables {\em w12a--w12f} on order of definite articles 
and nouns in object position
\item variables {\em w20a--w20e} on singular count nouns in vocative phrases
\item variables {\em w21a--w21e} on proper nouns in vocative phrases
\item variables {\em w22a--w22e} on plural nouns in vocative phrases. 
\end{itemize}
A specific description of each variable is given in the SSWL online site \cite{SSWL} or in the
updated Terraling database that SSWL migrated to, \cite{Terraling}.

\begin{figure} 
	\includegraphics[width = 5.5in]{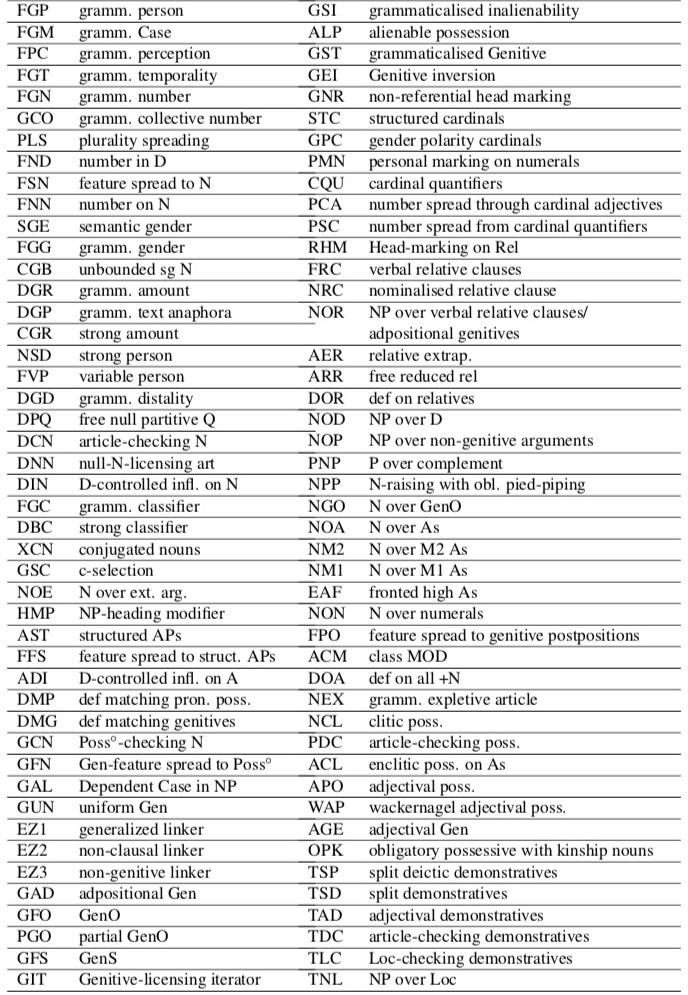}
	\caption{List of the LanGeLin syntactic parameters.
	\label{FigLanGeLin} }
\end{figure}

\smallskip

One of the main problems of the SSWL data is the fact that different languages
across the database are recorded with very different levels of completeness.
While for some languages $100\%$ of the SSWL binary variables are mapped,
other languages have  very small percentage of completeness. This problem occurs
not only between language families but also within each family. We will discuss later
how we deal with this problem of incomplete mapping. This problem does not
affect the LanGeLin data.

\smallskip

The data of the LanGeLin collaboration should be regarded a priori as an
independent set of data with respect to the features recorded in the SSWL data.
The variables in the LanGeLin can be regarded linguistically as genuine syntactic
parameters based on the Modularized Global Parameterization approach developed by 
Longobardi \cite{Longo3}, \cite{Longo2}. The LanGeLin data are not strictly binary variables,
since they also encode the possibility that some of the parameters may become undefined
as an effect of entailment relations from other parameters. Thus, they are encoded as
ternary variables with the two main binary states $\pm 1$ for a syntactic parameter being
expressed or not expressed in a given language and a third state $0$ to allow for the 
undefined case. The syntactic parameters recorded in the LanGeLin data are listed
in Figure~\ref{FigLanGeLin} (reproduced from \cite{Kazakov}).

\smallskip

\subsubsection{Languages and Language Families} 

The SSWL dataset covers a set of 253 languages. In our analysis here we focus only on
certain subsets of languages that belong to some of the main language families.
The two families that are best represented in the database are the Indo-European 
and the Niger-Congo, hence we will be primarily working with these languages. 
We will also consider two other families with fewer representatives in the SSWL 
database: the Afro-Asiatic and the Austronesian languages. 

\begin{itemize}
\item {\em Indo-European}: Afrikaans, Albanian, Ancient Greek, Armenian, Bellinzonese, Bengali,
Brazilian Portuguese, Breton, Bulgarian, Burgenland-Romani, Calabrian, Catalan, Cimbrian, 
Croatian, Cypriot Greek, Czech, Danish, Digor Ossetic, Dutch, Eastern Armenian, English, 
English (Middle), English (Singapore), European Portuguese, Faroese, Farsi, French, 
French (Ivorian), Frisian, Galician, German, Gothic, Greek, 
Greek (Calabria, Cappadocian, Homeric, Kydonies, Lesbos, Medieval, Pontic), Haitian Creole,
Hindi	, Hittite, Icelandic, Irish, Italian, Italian (Napoletano Antico, Old, Reggiano), Iron Ossetic, 
Jamaican Creole English, Kurdish (Sorani), Latin, Latin (Late), Lithuanian, Middle Dutch, M\`ocheno, 
Neapolitan, Nepali, Norwegian, Occitan, Odia (Oriya), Oevdalian,
Old English, Old French, Old Norse, Old Saxon, Panjabi,
Pashto, Polish, Portuguese, Romanian, Russian, Sanskrit, Scottish Gaelic, Saramaccan, Serbian,
Sicilian, Slovenian, Spanish, Swedish, Swiss German, Teramano, Tocharian A, Tocharian B, Ukrainian,
Vla\v{s}ki-\v{Z}ejanski-Istro-Romanian, Welsh, West Flemish, Western Armenian, Yiddish. 

\medskip

\item {\em Niger-Congo}: Agni Ind\'eni\'e, Akan-twi, Babanki, Bafut, Bambara, Bandial, 
Baoul\'e, Basaa, Baule-K\^{o}deh, Beng, Chichewa, Dagaare, Digo, Ewe, Ewondo, 
Farefari, Fe'efe'e, Ga, Ghom\'al\'a', Gu\'ebie, Gungbe (Porto-Novo), Gurene, Hanga,
Ibibio, Igala, Igbo, Ijo (Kaiama), Kenyang, KiLega, Kinande, Kindendeule, Kiswahili,
Kiyaka, Kom, Konni, Koyo, Kusaal, Lubukusu, Mada, Mankanya, Medumba, Naki,
Nawdm, Nda'nda', Ndut, Nkore-Kiga, Nupe, Nweh, Olukumi, Oluwanga, Oluwanga (Eji),
Shupamem, Tommo-So, Tuki (Tukombo), Twi, Vata, Wan, Wolof, Xhosa, Y\'emba,
Yoruba, Zulu.

\medskip

\item {\em Austronesian}: Acehnese, Atayal (Squliq), Bajau (West Coast), Fijian, 
Ilokano, Indonesian, Isbukun Bunun, Kayan, Malagasy, Maori, Marshallese, 
Niuean, Palue, Papuan Malay, Samoan, Sasak, Tagalog, Titan, Tongan, Tukang Besi, 
Zamboage\~{n}o Chabacano. 

\medskip

\item {\em Afro-Asiatic}: Amharic, Arabic (Gulf), Bole, Egyptian Arabic, Hausa,  Hebrew,
Hebrew (Biblical), Lebanese Arabic, Mbuko, Miya, Moroccan Arabic, Muyang, Senaya, 
Tigre, Wolane.

%\medskip
%\item {\em Sino-Tibean}: Burmese, Cantonese (Guangzhou), Hakka, Mandarin, 
%Meeteilon, Taiwanese Southern Min, Tiwa, Wuhu Chinese.

\end{itemize}

The list of languages included in the LanGeLin data is given by:
Kadiweu (Ka),    
Kuikuro (Ku),    
Ragusa (RGS),  
Mussomeli  (MuS),  
Aidone (AdS),  
Northern Calabrese (NCa), 
Southern Calabrese (SCa),  
Salentino (Sal), 
Campano (Cam),  
Italian (It),
Spanish (Sp), 
French (Fr),
Portuguese (Ptg),
Romanian (Rm),	
Latin (Lat),
Classical Greek (CIG),  
New Testament Greek (NTG),   
Salento Greek (SaG),  
Calabrian Greek A (CGA),   
Calabrian Greek B (CGB),   
Greek (Grk), 
Romeyka Pontic Greek (RPG),   
Cypriot Greek (CyG),  
Gothic (Got), 
Old English (OE),
English (E),	
German (D),
Danish (Da),
Icelandic (Ice),
Norwegian (Nor), 
Bulgarian (Blg),
Serb-Croatian (SC),
Slovenian (Slo),
Polish (Po), 
Russian (Rus),	
Irish (Ir), 
Welsh (Wel), 
Marathi (Ma), 	
Hindi  (Hi),
Farsi (Far), 
Pashto (Pas),
Mandarin (Man), 
Cantonese (Can),	
Inuktitut (Inu), 
Japanese (Jap),
Korean (Kor),	
Arabic (Ar) 
Hebrew	(Heb),
Hungarian (Hu), 
Khanty (Kh), 
Estonian (Est),
Udmurt (Ud), 
Yukaghir (Yu), 
Even (Ev), 
Evenki (Ek), 
Yakut (Ya), 
Turkish (Tur),	
Buryat (Bur), 
Central Basque (cB), 
Western Basque (wB), 
Wolof (Wo).

\subsubsection{The Altaic and Ural-Altaic hypothetical families}

We consider also the case of the Ural languages and the more
hypothetical Altaic grouping. This is interesting because of the 
contested Ural-Altaic family hypothesis. In particular, while earlier
attempts at identifying a hypothetical Altaic family had suggested
the inclusion of both Korean and Japanese into this putative
family, later studies had discarded the idea that these should be
included, while it retained the possibility of an Altaic family encompassing
languages like Turkish, Buryat, Yakut, Even, Evenki, Karachay, and Tatar.
The inclusion of the Turkic languages like Tuvan and Uyghur in a hypothetical
Altaic family is now also generally discarded. It was also hypothesized that
both the Uralic and the Altaic languages should fit into a large Ural-Altaic
family, although this hypothesis is also considered controversial, see \cite{Marca}, \cite{Sinor} 
for an overview.
The LanGeLin syntactic data have already been used recently to
study these families and the Ural-Altaic hypothesis, \cite{LongoCeo}. 
We discuss here what information one can extract about this historical linguistic
hypothesis from the persistent topology method. 

\smallskip

The SSWL database has few languages that can be used to
investigate the Ural family and the Altaic and Ural-Altaic hypothesis:
\begin{itemize}
\item {\em Uralic}: Estonian, Finnish, Hungarian, Udmurt.
\item {\em Altaic}:  Karachay, Tatar, Turkish, Tuvan, Uyghur.
\item {\em Previously hypothesized as Altaic}: Japanese, Korean, Okinawan.
\end{itemize}

\smallskip

The LanGeLin data have as list of languages covered
by the hypothetical  Ural-Altaic classification:

\begin{itemize}
\item {\em Uralic}: Estonian, Finnish, Hungarian, Udmurt, Yukaghir, Khanty.
\item {\em Altaic}:  Turkish, Buryat, Yakut, Even, Evenki.
\item {\em Previously hypothesized as Altaic}: Japanese, Korean.
\end{itemize}

\subsubsection{Comparative performance of datasets}

The comparative analysis carried out in \cite{OBM} shows that the SSWL and the
LanGeLin datasets behave very differently in terms of clustering properties. This
finding will be confirmed, by different methods, in the present paper. The results
of \cite{OSBM} show that both dataset perform reasonably well in terms of
phylogenetic reconstruction, provided some care is taken into dealing with
the lacunae of the SSWL data, but still with a tendency for the LanGeLin dataset 
to have better performance. Again we will confirm here, with a different method,
that the LanGeLin data behave better in terms of phylogenetic reconstruction,
although the trees we construct in this paper should not be regarded as
phylogenetic trees but as hierarchical clustering structures of syntactic features. 

\medskip

\section{Cluster analysis}\label{cluster_analysis}

%\begin{figure}
%	\subfloat[Indo-European]{\includegraphics[width = 5in]{filled_data_clusters_IE_SSWL_old.png}} \\
%	\subfloat[Austronesian]{\includegraphics[width = 5in]{filled_data_clusters_austro_SSWL_old.png}} 
%	 \caption{Number of clusters by radius for the SSWL data for the Indo-European and 
%	Austronesian languages, with percent variance of $60\%$ in the PCA. 
%	\label{clusters_SSWL_1} }
%\end{figure}

\begin{figure}
	\subfloat[Indo-European]{\includegraphics[width = 5in]{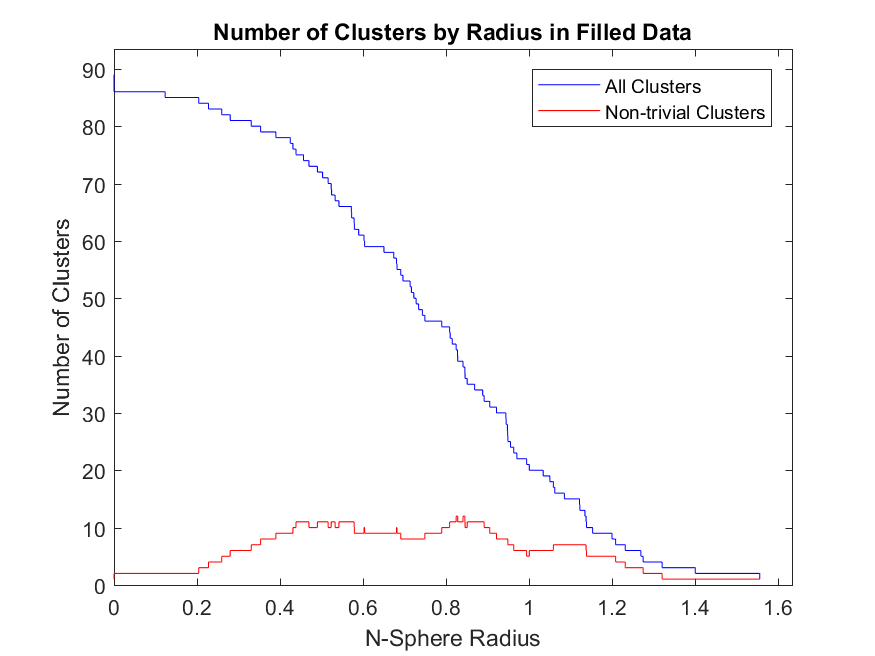}} \\
	\subfloat[Austronesian]{\includegraphics[width = 5in]{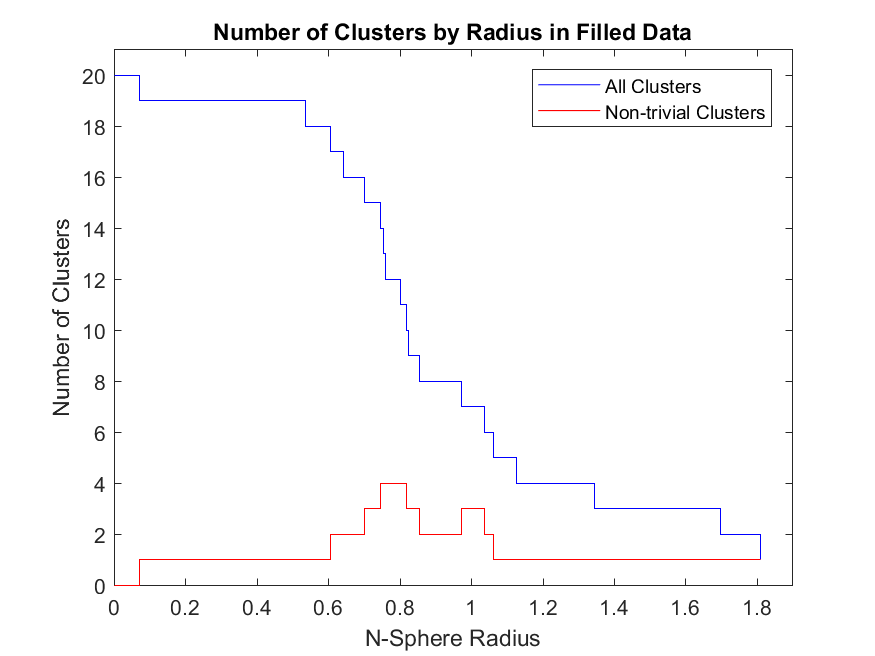}} 
	 \caption{Number of clusters by radius for the SSWL data for the Indo-European and 
	Austronesian languages, with PCA percent variance $60\%$. 
	\label{clusters_SSWL_1} }
\end{figure}

%\begin{figure}
%	\subfloat[Niger-Congo]{\includegraphics[width = 5in]{filled_data_clusters_AC_SSWL_old.png}} \\
%	\subfloat[Afro-Asiatic]{\includegraphics[width = 5in]{filled_data_clusters_AA_SSWL_old.png}}
%	\caption{Number of clusters by radius for the SSWL data set for the Niger-Congo and Afro-Asiatic languages, with percent variance of $60\%$ in the PCA. 
%	\label{clusters_SSWL_2} }
%\end{figure}

\begin{figure}
	\subfloat[Niger-Congo]{\includegraphics[width = 5in]{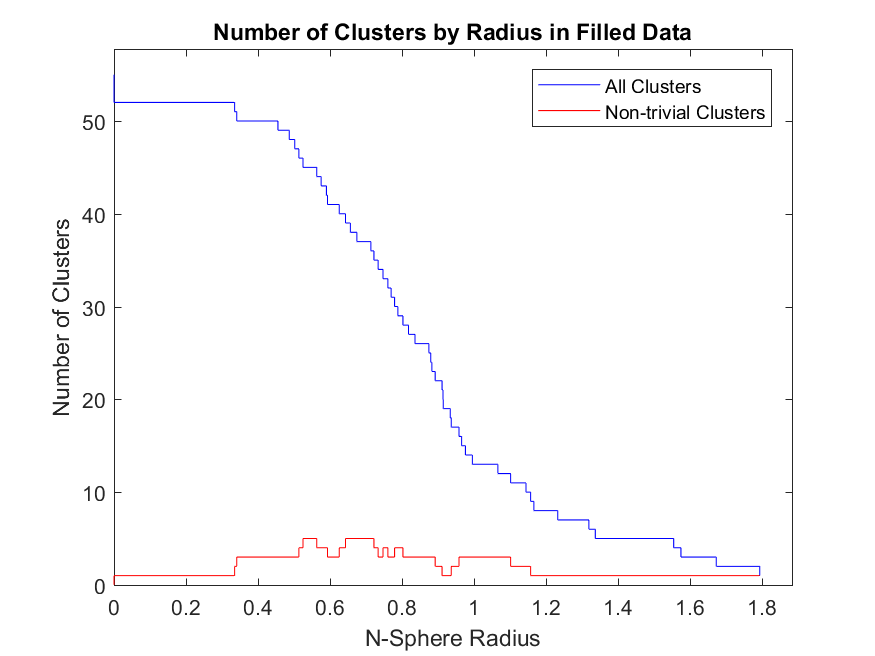}} \\
	\subfloat[Afro-Asiatic]{\includegraphics[width = 5in]{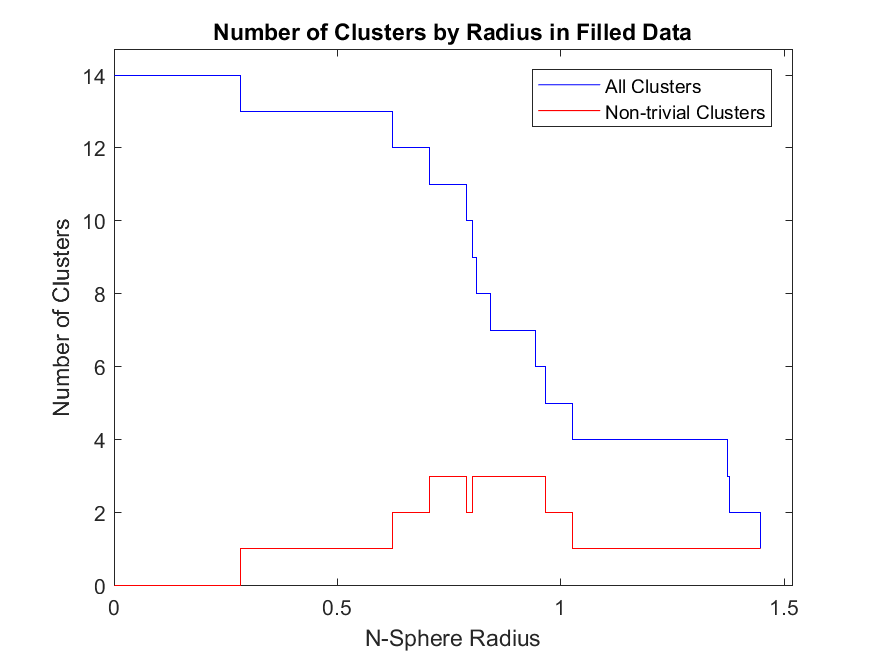}}
	\caption{Number of clusters by radius for the SSWL data set for the Niger-Congo and Afro-Asiatic languages, with PCA percent variance $60\%$. 
	\label{clusters_SSWL_2} }
\end{figure}

\begin{figure} [!htb]\center{\includegraphics[width=\textwidth] {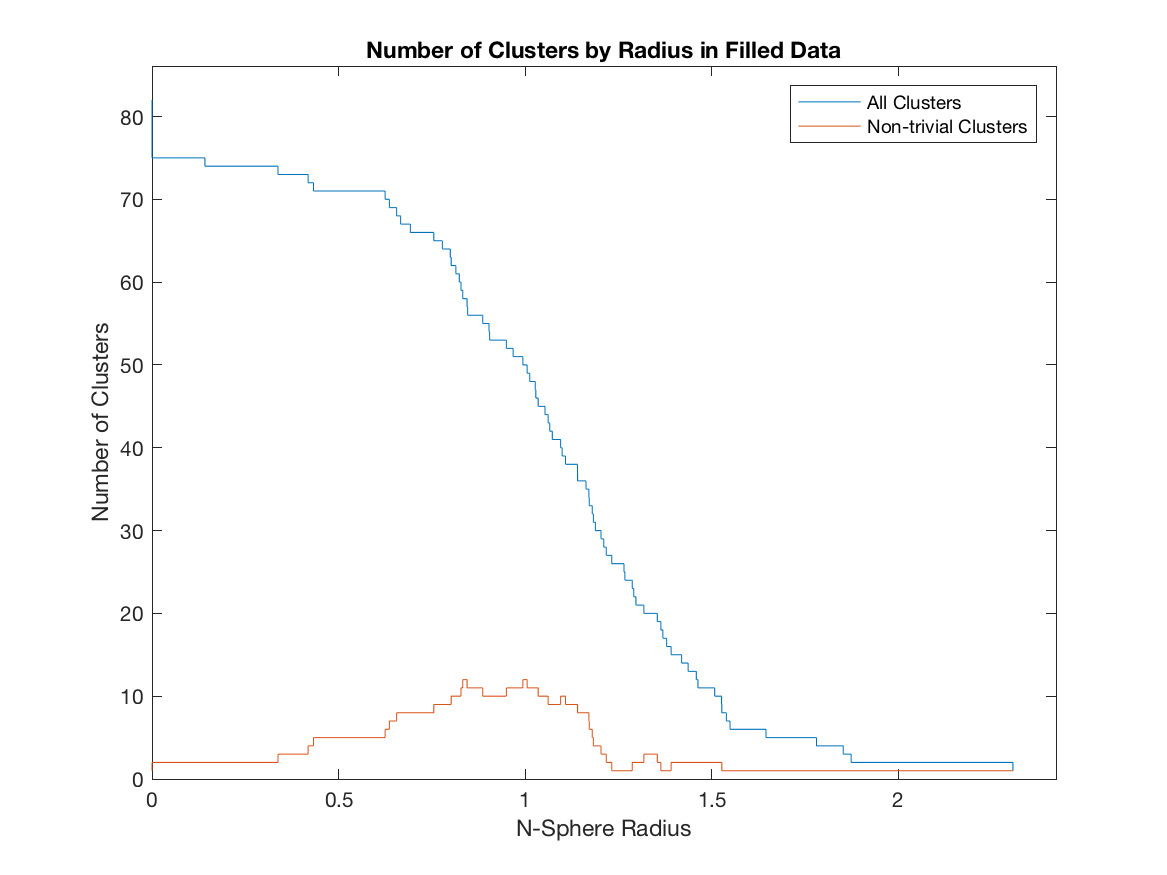}} \label{clusters_Longobardi}  \caption{The number of clusters by radius for the LanGeLin data set, with PCA variance $60\%$.\label{Fig_clusters_Longobardi}}
\end{figure}

The first thing to observe when looking at the number of clusters by radius (see Figure~\ref{clusters_SSWL_1}, Figure~\ref{clusters_SSWL_2} 
and Figure~\ref{Fig_clusters_Longobardi}) is that different language families (Indo-European, Austronesian, Niger-Congo, and Afro-Asiatic) exhibit 
different cluster structures. The Indo-European family consists of the largest number of non-trivial clusters (Figure~\ref{clusters_SSWL_1}). Other
language families (see Figures~\ref{clusters_SSWL_1} and \ref{clusters_SSWL_2}) barely have non-trivial clusters at any given radius.
This might suggest, as mentioned in \cite{Port}, that in the case of language families with a smaller amount of non-trivial clustering
the syntactic parameters are more centered and homogeneously distributed across the different languages compared to the Indo-European family. 

\smallskip

The Longobardi LanGeLin data set contains mainly Indo-European languages so we can compare the cluster
analysis of the LanGeLin data in Figure~\ref{Fig_clusters_Longobardi} to the clustering of the 
Indo--European language family in the SSWL database, Figure~\ref{clusters_SSWL_1}.

\smallskip
\subsection{Singletons and clusters}

Another interesting point is that in the SSWL there are singletons (i.e clusters containing only one data point) for every radius, and in every language family (except the Austronesian). In the Longobardi LanGeLin data, on the other hand, starting from a certain radius there are no more singletons. Linguistic interpretations for this phenomena may involve the presence in the SSWL database of languages that are farther away from each other, or the different
nature of the SSWL syntactic features.

\smallskip

In fact the different clustering structure of the SSWL and the LanGeLin data appears to reflect
the different nature of the syntactic variables that are recorded in the two databases. Indeed,
in the analysis carried out in \cite{OBM}, where the data are analyzed by syntactic
parameters rather than by languages, a similar phenomenon is observed, whereby
the SSWL data contain more singletons that last of a wider range of scales before being
incorporated into clusters, while in the LanGeLin data clusters form more rapidly.
Since the analysis of \cite{OBM} is by syntactic features across languages rather than
by languages, no difference is detected there between different language families, while
in the cluster analysis we performed here we see a clear difference between different
language families.

\smallskip

One can raise the question of whether a part of this effect may also be due to the inherent
problems of the SSWL data, namely the fact that languages are not uniformly mapped.
However, this would more likely create an opposite effect, where languages with a large
set of incomplete binary variables may appear closer than they really are because
of a large overlapping set of lacunae. Similarly homeoplasy effects due to languages that
are historically far away from each other being closer at the level of syntactic structures
would also tend to produce an opposite effect with more rather than less clustering.

\smallskip
\subsection{Clustering and the role of Greek-Italian Microvariations}

In \cite{Gua16}, an in depth study of linguistic diversity is carried out for a range of
Romance and Greek dialects, analyzed at the syntactic level using the LanGeLin data
of syntactic parameter. This type of analysis of ``microvariations" applies to languages
that are either genealogically very closely related or belong to distinct genealogical groups 
but are in close proximity and interaction within a limited geographic region. This is indeed
the case of a range of dialects between the Romance and Greek language families in
the Southern region of Italy. The languages in the LanGeLin database that are involved
in this microvariations structure are the Romance dialects 
Ragusa (RGS), Mussomeli  (MuS),  Aidone (AdS), Northern Calabrese (NCa), 
Southern Calabrese (SCa), Salentino (Sal), Campano (Cam), and in the Greek family 
the dialects Salento Greek (SaG), Calabrian Greek A (CGA), and Calabrian Greek B (CGB).

\smallskip

We observe here, by looking at the position of the Greek languages within the
two databases, that the very different clustering structure they exhibit with
respect of the rest of the Indo-European languages may reflect the fact that the
LanGeLin data include all the information about these microvariations in the
region between the Romance and Greek languages, while the SSWL does not.

\smallskip

More precisely, a more in depth look into the clusterification of the two data sets, 
via the tree that keeps track of how singletons merge into non-trivial clusters as
the scale parameter increases, reveals an interesting difference with respect to
the role of the Greek language family. 
In the SSWL dataset, the farthest/last singletons that are added are Cappadocian 
Greek, Ancient Greek and Greek. Then, other Greek languages occur, Medieval 
Greek and Homeric Greek, along with other languages such as Kurdish, Latin, 
and late Latin. Only then we get the first non-singleton clusters. In other words, in
the SSWL data for the Indo-European languages, the Greek languages are 
added later as farthest from the rest of the languages in the data set and 
far from each other. 
The only Greek languages that are not added later are highly incomplete languages, 
hence we can ascribe their faster clustering to a spurious effect due to lacunae. 
All of the Greek languages mentioned above are highly completed with more than 
$85\%$ parameters filled in most of them, and not less than $65 \%$ parameters filled 
in the rest, so their singleton behavior can be considered reliable and not 
affected by incompleteness.

\smallskip

In the Longobardi data set, on the other hand, the Greek languages have 
neighbors quite early on and are not isolated for long. This in part reflects 
the presence in the LanGeLin database of the set of Southern Italian
dialects corresponding to the Greek--Italian microvariation, with 
the dialects Salento Greek, Calabrian Greek A, and Calabrian Greek B belonging to the Greek family. 
As we will see when
we discuss the persistent components tree for the LanGeLin data, we also find a proximity
between Bulgarian and some of the Greek languages, see Section~\ref{LanGeLinH0Sec}.

\bigskip

\begin{figure}
	\subfloat[filter value 116]{\includegraphics[width = 6.5in]{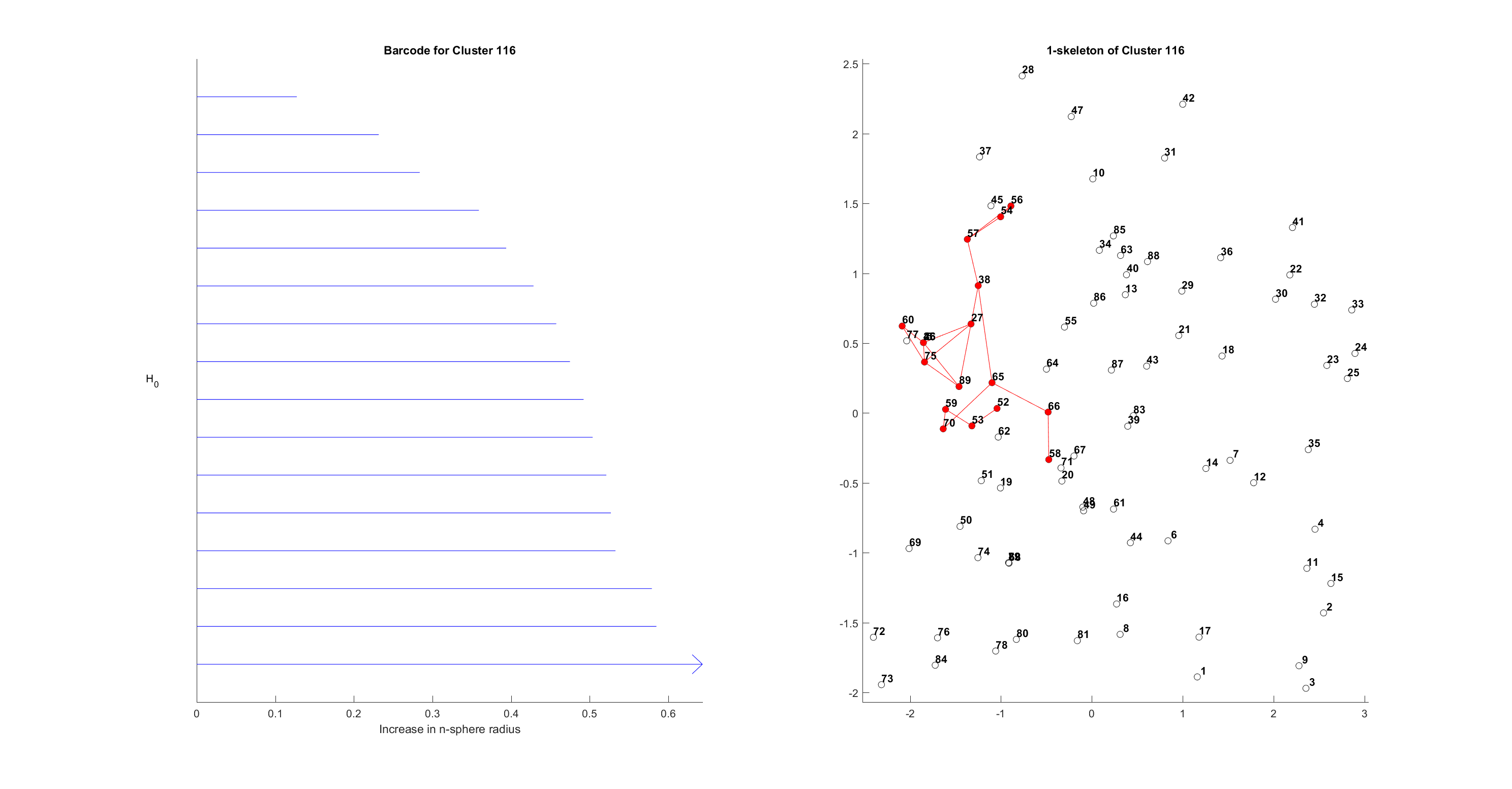}} \\
	\subfloat[filter value 120]{\includegraphics[width = 6.5in]{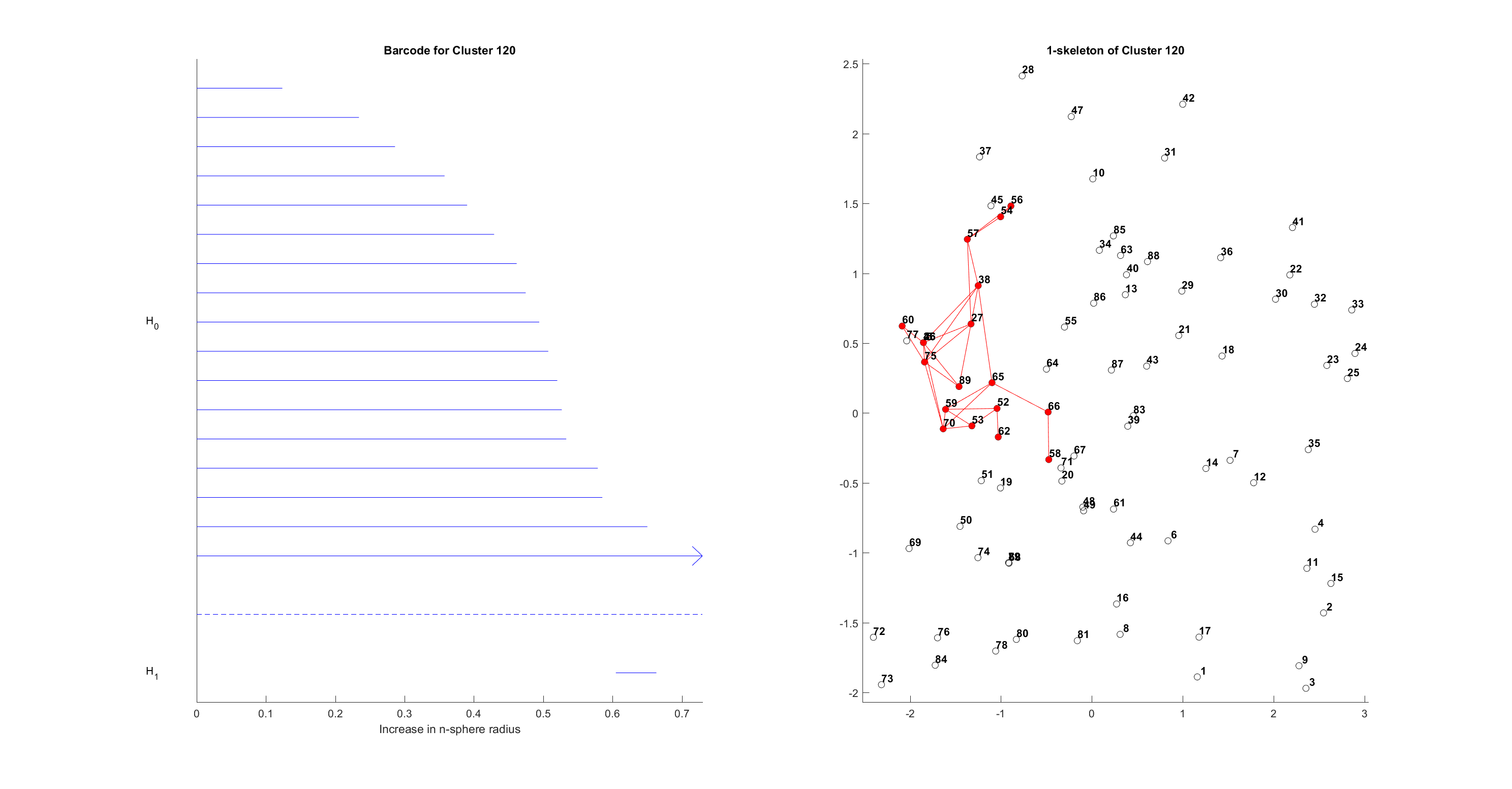}}\\

	\caption{Barcode graphs and their corresponding $1$-skeleta at filter values cluster $116$ and cluster $120$, SSWL data.
	\label{clusters_SSWL} }
\end{figure}

%\begin{figure}
%	\subfloat[filter value 140]{\includegraphics[width = 6.5in]{figure_cluster_140.png}} \\
%	\subfloat[filter value 141]{\includegraphics[width = 6.5in]{figure_cluster_141.png}}\\

%	\caption{Barcode graphs and their corresponding $1$-skeleta at filter value $140$ and $141$.
%	\label{clusters_SSWL} }
%\end{figure}

\section{Principal Components and Persistent Topology Barcodes}

For the sake of computation time and in order to investigate the relations between the parameters, we introduce PCA as a method of dimensional reduction on the data. PCA rewrites the data in an ordered way that lets us reduce its dimension based on variance. We observe that running the analysis at different levels of PCA variance 
preserved (e.g.~$60\%$ and $80\%$) does affect the structure of the clusters (hence the structure of the resulting persistent components trees). However it does not do so in a major way. This means that typically the
varying the PCA variance level can cause some displacements of branches within smaller subfamilies, 
but does not affect the major subdivisions into main clusters. We will illustrate in detail
an example where one can see an effect caused by varying the level of PCA variance preserved, 
when we discuss the case of the Germanic languages in Section~\ref{SSWLGermSec}. 
Since some form of PCA is often used in the phylogenetic reconstruction based on syntactic parameters, 
it is useful to keep in mind that there can be effects due to the level of PCA variance that can slightly 
alter the form of the resulting trees. 

\smallskip

\begin{figure}
	\subfloat[$2D$ scatter]{\includegraphics[width = 5in]{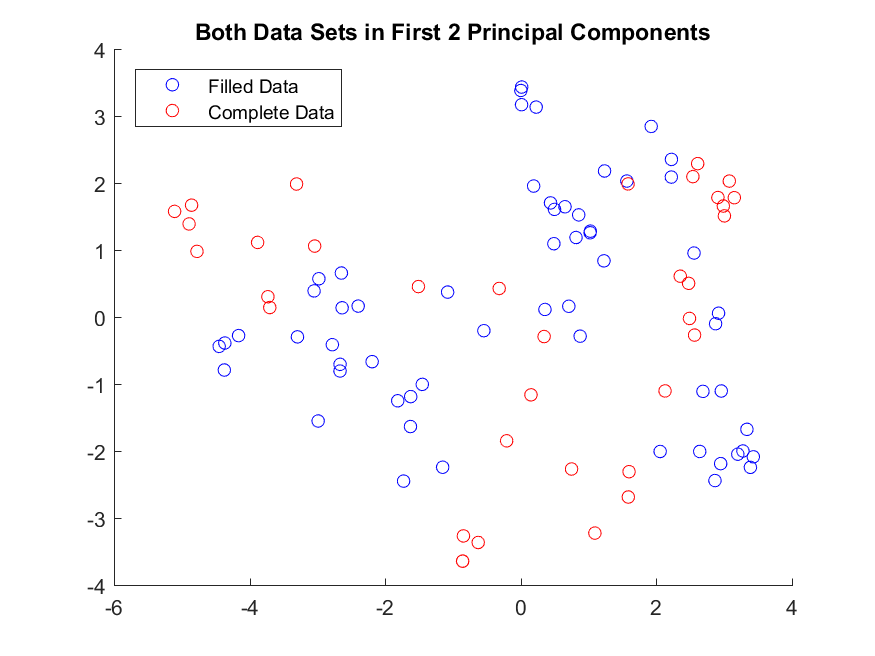}} \\
	\subfloat[$3D$ scatter]{\includegraphics[width = 5in]{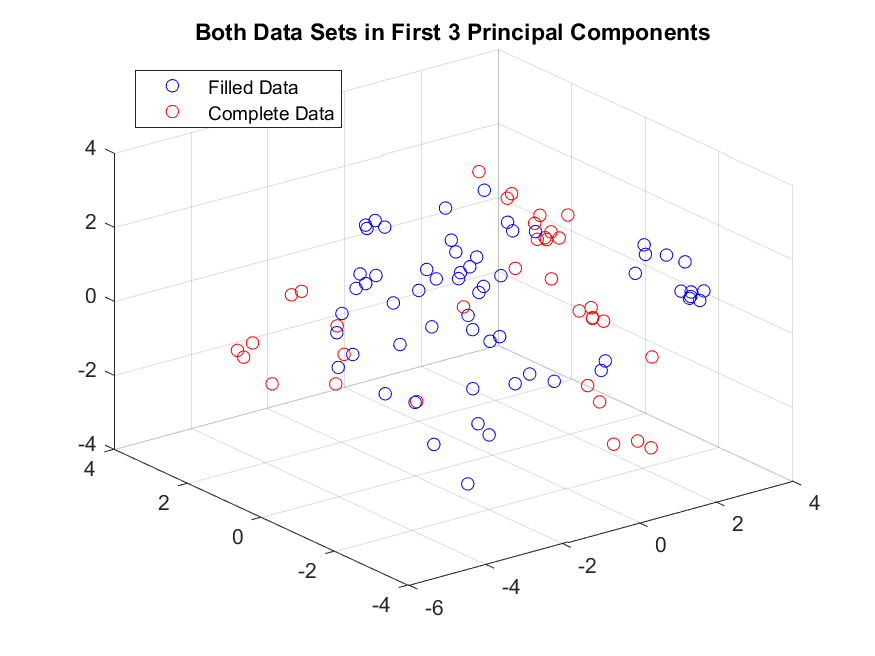}}
	
	\caption{$2D$ and $3D $ scatters of the data in the first $2$ and $3$ principal components.
	%in the LanGeLin data set, both filled and completed. 
	\label{scatters_Longobardi} }
\end{figure}

Our data are originally recorded as a discrete data set but after PCA it is given by continuous variables and the distance between the points is given by the Euclidean distance in the ambient space. Based on these data of distances, the \textit{Perseus} program \cite{Perseus} builds a simplicial complex at each radius and computes the persistent homology. The outputs are given as barcode graphs and $1$-skeleta as illustrated in Figure~\ref{clusters_SSWL}, $2D$ and $3D$ scatters of the data in the first $2$ and $3$ principal components,
as shown in Figure~\ref{scatters_Longobardi}, and the number of clusters in each radius, as in Figure~\ref{clusters_SSWL} and Figure~\ref{Fig_clusters_Longobardi}. 
A more thorough explanation of the this part of the code can be found in \cite{Port}.  

\smallskip

The use of a dimensional reduction method like PCA prior to the persistent homology
computation performed by Perseus is necessary in order to make the problem computationally
tractable, as Perseus would not be able to handle the persistent homology computation over
the entire high-dimensional space. 

\smallskip

A comment that will be useful later, when we analyze the persistent components
trees of various subfamilies within larger families, is the construction of the radius and steps in radius increase in our algorithm. We start, for a given set of data, by
computing a critical radius $r_c$ which is the minimal length such that by drawing balls of radius $r_c$ around each point in the set one obtains a single connected
component. Then a scale $\epsilon$ is fixed to be a chosen fraction of that radius, such as $\epsilon=r_c/100$, and the algorithm then increases the radius in steps of
$\epsilon, 2\epsilon, \ldots, N \epsilon\ldots$. It is possible that, if the initial critical value is computed over the entire dataset, when restricting to subfamilies of very closely
related languages, the chosen step size $\epsilon$ will not be large enough to resolve all the points in the subfamily, missing some of the tree structure. It is then necessary
to adapt the radius $r_c$ and the size steps $\epsilon$ to the individual subfamilies to resolve all the languages in the subfamilies. 

\smallskip

\subsection{Principal components and weighted syntactic structures}

The PCA method, as mentioned above, replaces the original set of
binary variables, seen as vectors with binary coordinates inside an
ambient real Euclidean space, with a new set of real variables given 
by the components of the original vectors along a new basis given
by the principal components. 

\smallskip

In other words, we can describe the principal components as a 
weighted sum of the original coordinate axes that provide the new
set of coordinate axes that maximize variance so as to replace
a set of possibly correlated variables by a set of values of linearly 
uncorrelated variables.

\smallskip

In our setting the original coordinate axes are identified with the
different binary syntactic variables (syntactic parameters), while
the principal components are then weighted sums of these original
variables. Thus, the process of computing principal components
can be viewed as an assignment, based on the data of the set
of languages considers, of a real valued weight to each syntactic
parameter/variable. The process reduces the dimensionality of
the data by replacing the full set of original variables by the first
few principal components, see Figure~\ref{scatters_Longobardi}.

\smallskip

Since the principal components select a set of new linearly
independent variables, this weight assignment can be
interpreted heuristically as a measure that weights more
heavily those syntactic parameters/variables that are more
likely to be genuinely independent variables while weighting
less those that are more dependent on the others. 
Thus, linguistically we can think of
this operation as replacing the original choice of the syntactic
parameters by a new weighted system of the same variables
which express the coordinates of the new principal components. 

\smallskip

It was argued in \cite{OSBM} and \cite{SiTaMa} that for the
purpose of considering dynamical models of language change,
either in the Markov models on trees used in the phylogenetic
algebraic geometry method in \cite{OSBM} or in the spin
glass models considered in \cite{SiTaMa} one should not
consider it satisfactory to assume that the syntactic parameters
are independent identically distributed random variables, because
of the existence of unspecified relations between them. Thus, it
would be more natural to weight the syntactic parameters by an
indicator of dependence/independence. One natural choice
of how to do that is by using as weights the coefficients of
the principal components that maximizes variance.

\smallskip

We observe that the principal components look quite different
in the case of the SSWL data and in the case of the LanGeLin data. 
The different structure indicates fewer {\em linear} relations
among the SSWL variables than among the LanGeLin variables.
Notice though that the absence of linear relations does not indicate
an absence of relations of other kinds, see for instance the results
of \cite{ParkMa} for a strong indication of other relations among the SSWL variables. 

\smallskip

In the next section we discuss more in detail the question of the dimensionality
of the data of syntactic structures, as a way to better detect the presence of
relations between the syntactic variables, beyond what can be seen in terms
of principal components.

%\Note{... Figure to be added here ...}

\section{Dimensionality analysis} \label{DimSec}

One of the main problems when analyzing syntactic data is the fact that the parameters are not at all independent variables. Investigating the relations between the parameters and finding a good set of ``universal coordinates" 
for the space of syntactic parameters, or in other words, finding ``the geometry of syntax" can be regarded as 
one of the main open questions in Chomski's Principles and Parameters model of syntax. Therefore, we would 
like to estimate the dimension of the data in order to get a better sense of the dependencies and the relations between the parameters. 

\begin{figure}
	\includegraphics[width = 1.8in]{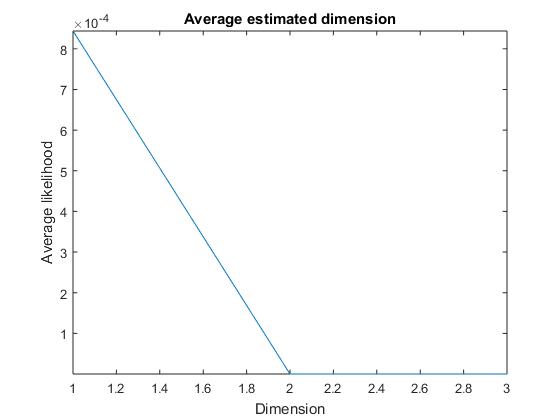}
	\includegraphics[width = 1.8in]{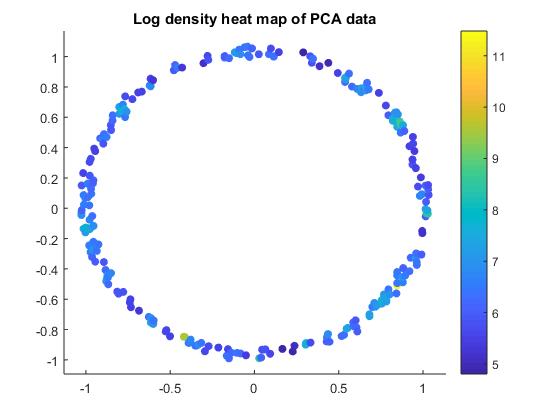} 
	\includegraphics[width = 1.8in]{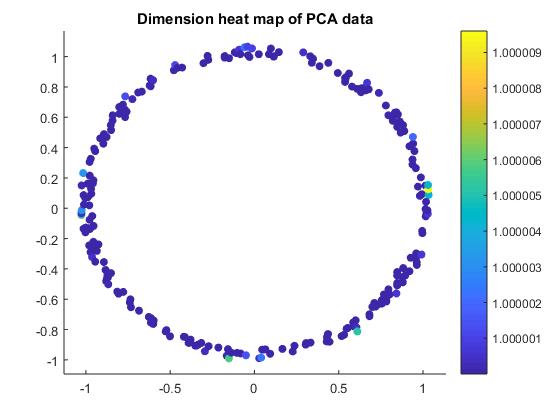} \\
	
	\includegraphics[width = 1.8in]{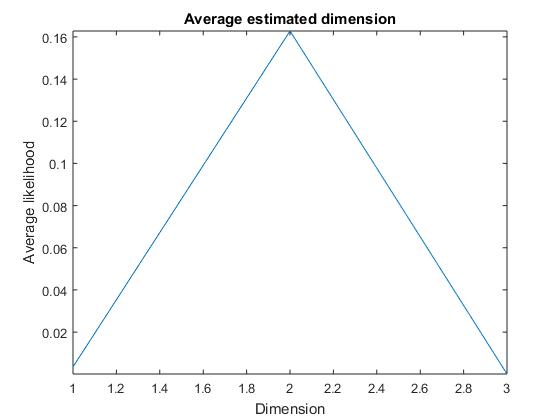}
	\includegraphics[width = 1.8in]{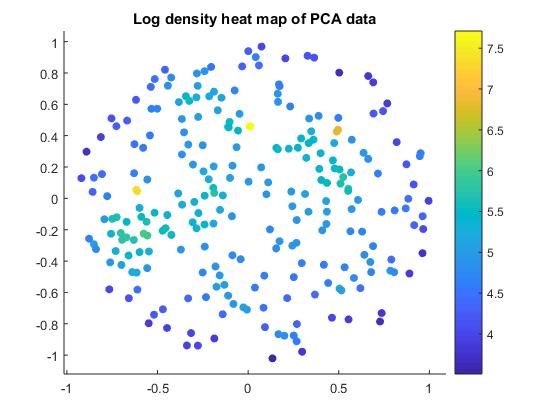} 
	\includegraphics[width = 1.8in]{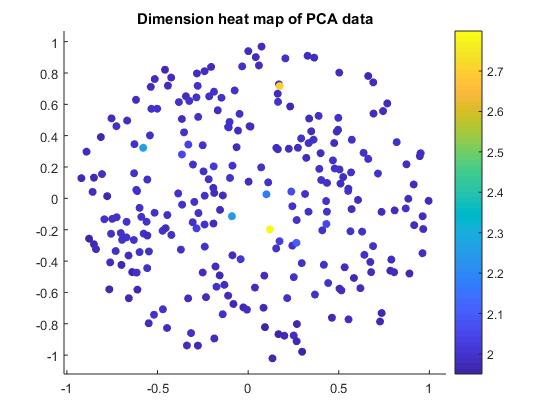} \\
	
	\includegraphics[width = 1.8in]{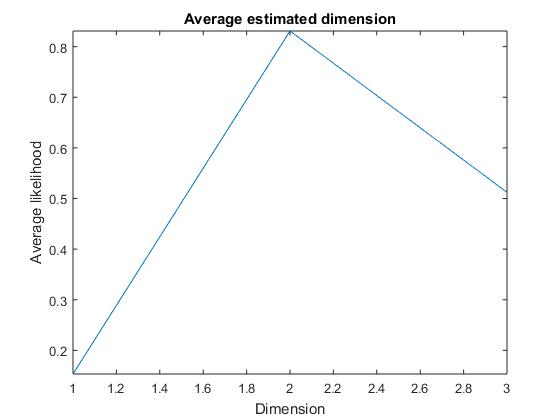}
	\includegraphics[width = 1.8in]{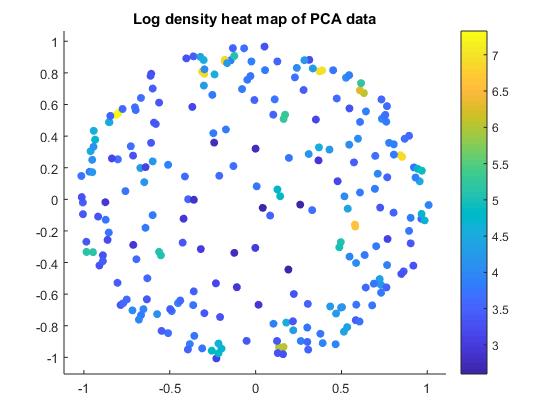} 
	\includegraphics[width = 1.8in]{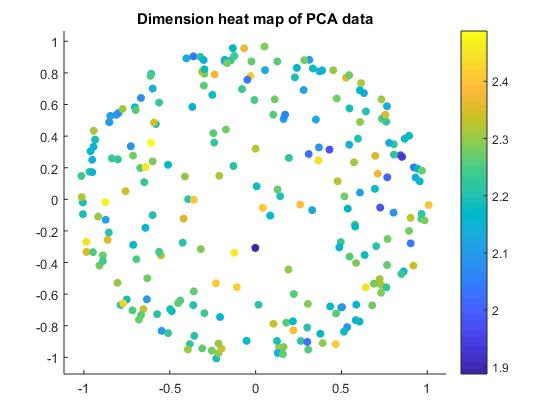} \\
	
	\includegraphics[width = 1.8in]{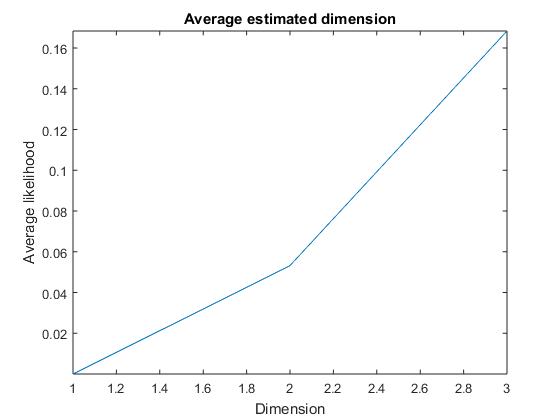}
	\includegraphics[width = 1.8in]{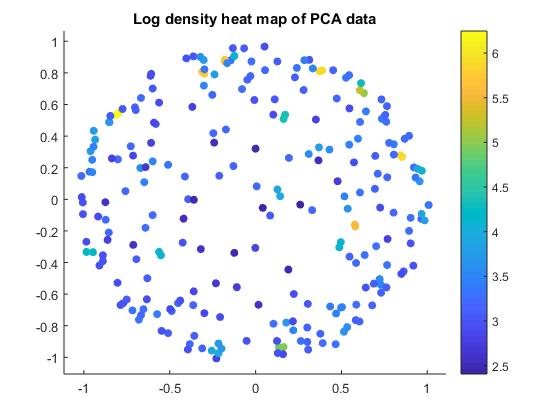} 
	\includegraphics[width = 1.8in]{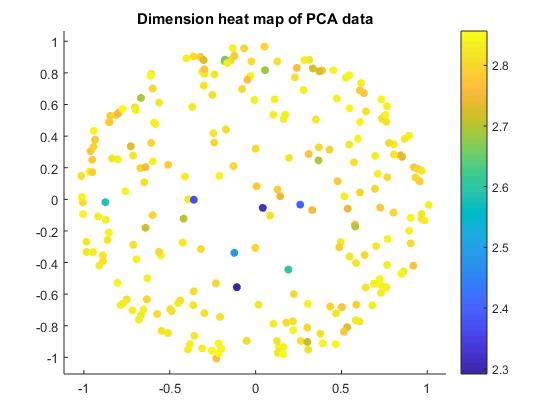} \\
	
	\caption{Dimensionality for simple geometries: $1$-sphere with $\alpha=1/3$, $2$-ball  with $\alpha=1/3$, and $2$-sphere
	with $\alpha=1/10$ and $\alpha=1/3$ in $\R^3$.
	\label{SpheresFig} }
\end{figure}

\subsection{Description of the algorithm}
\begin{itemize}
	\item Let $D \subseteq {\R}^d$ be our data set. 
	We choose $p \in \{1, \dots, |D|\}$ and a point ${\vec{x}_1}^{(p)}$.
	We then sort $D$ into a vector $\{{\vec{x}_i}^{(p)}\}_{i=1}^{|D|}$ where the entries are ordered by distance from the selected point, i.e.~where the distances monotonically increase, $$d({{\vec{x}_i}^{(p)}}, {{\vec{x}_k}^{(p)}}) \leq d({{\vec{x}_j}^{(p)}}, {{\vec{x}_k}^{(p)}}) \ \ \text{ for } \,  i \leq j .$$
	
	\smallskip
	
	\item Choose $s \in \{1, \dots, |D| \}$ to be the number of nearest neighbors and define $$X^{(p,s)} = \frac{1}{d({\vec{x_{s+1}}}^p, {\vec{x_1}}^p )}  
	{\begin{bmatrix}
		{\vec{x}_2}^{(p)} - {\vec{x}_1}^{(p)}\\
		{\vec{x}_3}^{(p)} - {\vec{x}_1}^{(p)} \\
		\vdots \\
		{\vec{x}_{s+1}}^{(p)} - {\vec{x}_1}^{(p)}
		\end{bmatrix}} \, .$$
	
	\smallskip
	
\item 	The data is spread out and we shift the selected points and their $s$-nearest neighbors so that 
	they will fit into the $d$-dimensional unit ball centered at the selected point. 
	
	\smallskip
	
\item Let $W^{(p,s)} \in M_{s \times s (\R)}$ be the \textit{weight matrix} with the following properties:
 \begin{enumerate}
 \item $W^{(p,s)}$ is diagonal,   \medskip
 \item $\det(W^{(p,s)}) =1$,   \medskip
 \item $W^{(p,s)}_{i,i} \sim e ^ {\frac{{-d}({{\vec{x}_{i+1}}^{(p)} - {\vec{x}_1}^{(p)}})^2}{\alpha}}$. 
 \end{enumerate}
 
 \medskip
 
 \item The range where $\alpha$ is small corresponds to the local behavior and the range where $\alpha$ is large to the global behavior.
 
 \smallskip

\item Following a common best fit algorithm we now define a \textit{ weighted covariant matrix}: 
$$C^{(p,s)} = \frac{1}{s} ({\vec{x}^{(p,s)}})^T W^{(p,s)} {\vec{x}^{(p,s)}}, $$

\smallskip

\item We compute its eigenvalues ${\lambda}^{(p,s)}_1  \ge  \dots \ge {\lambda}^{(p,s)}_d$ and corresponding eigenvectors $\{ {\vec{v}_{\lambda_1}}^{(p,s)}, \dots, {\vec{v}_{\lambda_d}}^{(p,s)}\} $ in the same ordering. The output is the \textit{eigenbasis matrix} $V^{(p,s)} \in M_{d \times d}(\R)$ where the $i$-th column corresponds to ${\vec{v}_{\lambda_i}}^{(p,s)}$. 

\medskip
  
\item We choose $f \in \{1, \dots, s\}$, the \textit{dimension of fit}. Let $P^{(f)} \in M_{d \times d}(\R)$ such that $P^{(f)}$ is diagonal and $$P^{(f)}_{ii}= \begin{cases}
0, \text{ if }  i \leq f  \\ 
1, \text{ if }   i > f
\end{cases} \, .$$

\medskip

\item The $(p,s,f)$-error is given by: 
$$W^{(p,s)} X^{(p,s)} ({V^{(p,s)}}^T)^{-1} P^{(f)} {V^{(p,s)}}^T \, ,$$
where the magnitude of the $i$-th row of $X^{(p,s)} ({V^{(p,s)}}^T)^{-1} P^{(f)} {V^{(p,s)}}^T$ is the orthonormal error of the $i$-th nearest neighbor. 

\medskip

\item Compute all the $(p,s,f)$-errors on the selected data and also on balls and spheres of $dim \leq d$ for comparison. 

\medskip

\item Run paired $T$-test between the selected points and balls/spheres database to get maximum likelihood $T$-value of $T$, see Figure~\ref{dim_est}.
%($d$ degrees of freedom then likelihood $2tcdf(-|T|,d)$)
\end{itemize}

\medskip

\subsection{Density estimation} For our data set $D \subset \R^d$ we define $R^{(p)} \in M_{{(|D|-1)} \times 1}$ as $$R^{(p)}_i = \frac{1}{\text{Vol}(B^d) \cdot d \cdot {d(\vec{x}^{(p)}_{i+1}, (\vec{x}^{(p)}_1)}^d }\, . $$ Computing then the quantity $$ \frac{\sum_i W^{(p, |D|-1)}_{i,i} R^{(p)}_i}{\sum_i W^{(p, |D|-1)}_{i,i}} $$ for every $p$ gives us a heat density map,
see Figure~\ref{dim_est2}.

\smallskip
\subsection{The test case of spheres and balls}

When running the algorithms described above for simple geometries like the circle, the disk, and the $2$-sphere, we
see more clearly the role of the $\alpha$ parameter and the information contained in the heat-map. In these cases
the dimension is clearly identified, although for larger values of $\alpha$ the dimension counting can appear altered towards
the larger dimension of the ambient space rather than the actual dimension of the submanifold. 
In the heat-map, while random points on the sphere have a more homogeneously distribution of 
densities, in a ball or disk those points located near the boundary tend to show a higher density 
than points located deeper into the bulk, see Figure~\ref{SpheresFig}.

\smallskip
\subsection{Estimated dimension of the space of syntactic structures} 

\begin{figure}
	\includegraphics[width = 3.5in]{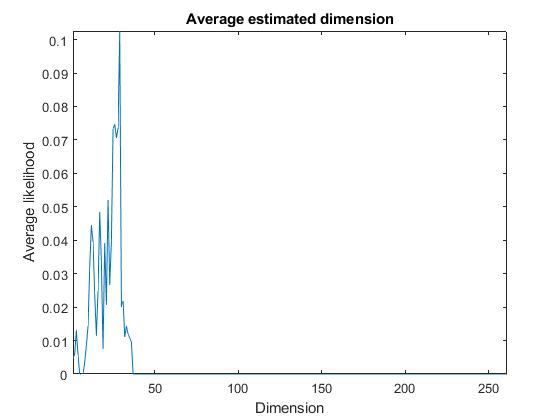} 
	\includegraphics[width = 3.5in]{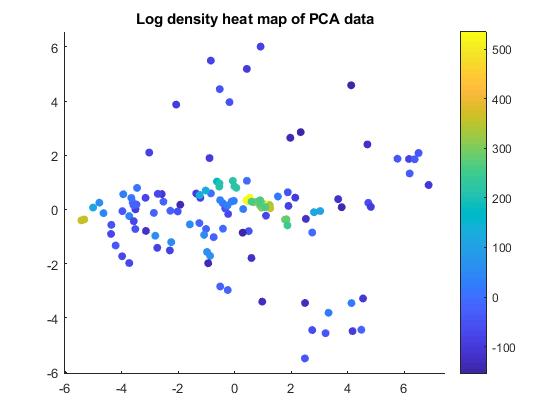} 
	\includegraphics[width = 3.5in]{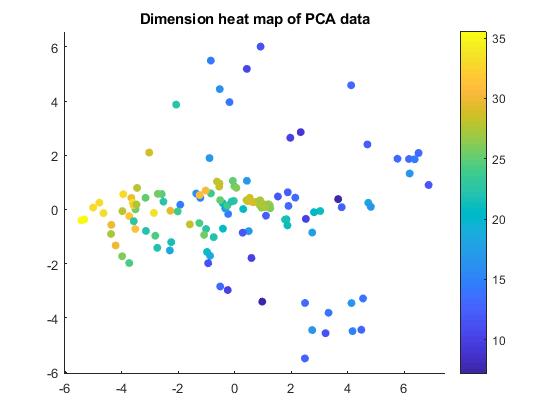}
	\caption{Dimensionality of the space of syntactic variables for the SSWL data at $\alpha=1$.
	\label{filterSSWLDimFig} }
\end{figure}

\begin{figure}
	\includegraphics[width = 5in]{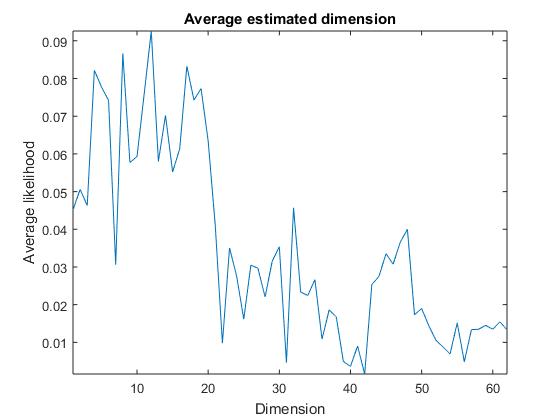} \\
	\includegraphics[width = 5in]{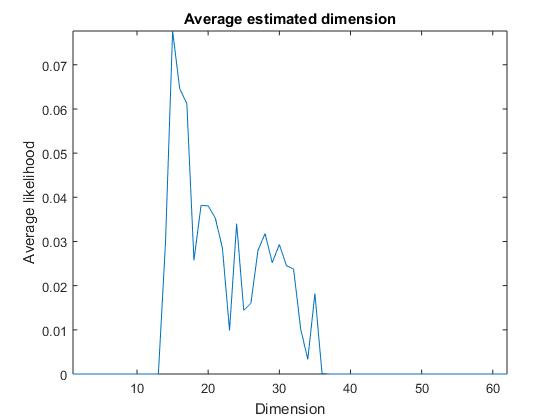}
	
	\caption{Dimension analysis of the LanGeLin data for $\alpha=0.1$ and $\alpha = \frac{1}{3}$ when averaging over the points.
	\label{dim_est} }
\end{figure}

\begin{figure}
	\includegraphics[width = 5in]{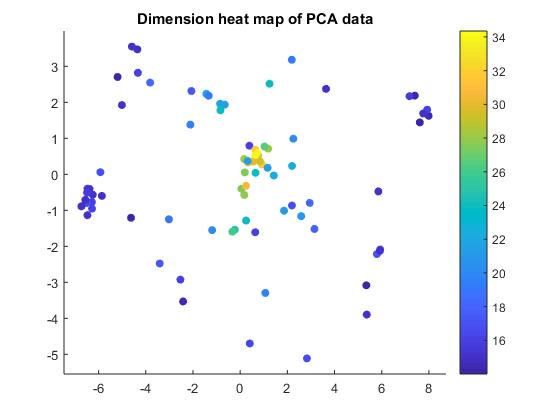} \\
	\includegraphics[width = 5in]{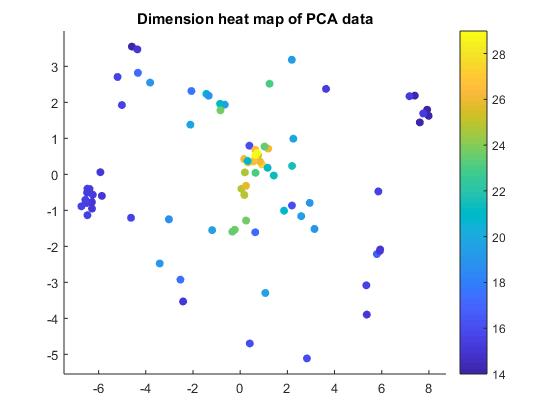}
	
	\caption{Dimension analysis of the LanGeLin data for $\alpha = 0.1$ and $\alpha = \frac{1}{3}$ when averaging 
	over the dimensions.
	\label{dim_est2} }
\end{figure}

In this dimensional analysis we have been considering the transposed data matrix, where each
data point corresponds to a syntactic parameter/syntactic variable and the vector entries correspond
to a language with the value of the parameter for that language recorded. This means that we
are looking at how syntactic features are distributed across languages and we would like to
identify the dimension of a submanifold of the full ambient space on which the data points lie.

\smallskip

After running the code implementing the algorithm described above on both the LanGeLin data set
and the SSWL data set for various $\alpha$ values, we observe the following outcomes.

\smallskip

For the LanGeLin data set at $\alpha = \frac{1}{3}$ (see Figure~\ref{dim_est} and Figure~\ref{dim_est2})
we obtain that the estimated dimension of the data is $d \sim 15$, whereas for $\alpha = 0.1$ it is too noisy to approximate, see Figure~\ref{dim_est}.  
For the SSWL data, on the other hand, we get that for $\alpha = 1$ the estimated dimension of the space of syntactic variables peaks
at around $d\sim 30$, see Figure~\ref{filterSSWLDimFig}.

\smallskip
\subsection{Dimension estimates of syntactic structures within language families}

\begin{figure}
	\includegraphics[width = 3.5in]{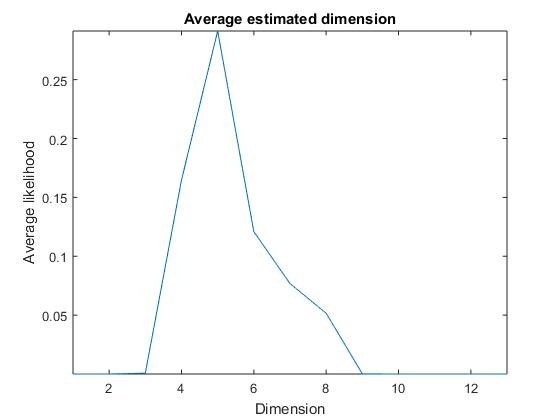} 
	\includegraphics[width = 3.5in]{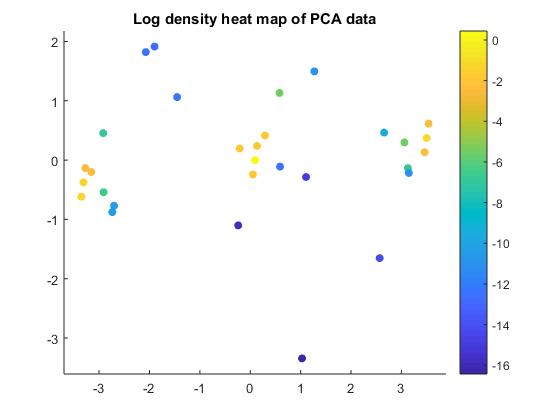} 
	\includegraphics[width = 3.5in]{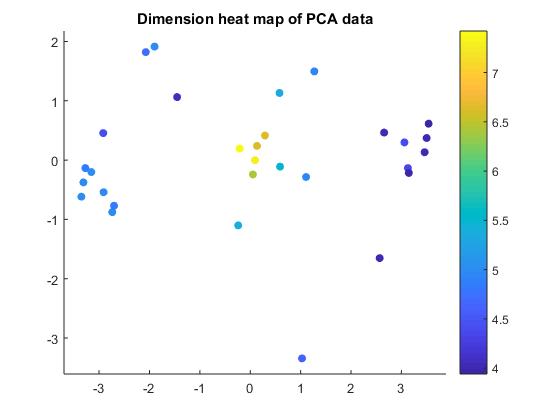}
	\caption{Dimensionality of the space of syntactic parameters for the Romance language family (LanGeLin data, $\alpha=1/3$).
	\label{RomanceDimFig} }
\end{figure}

\begin{figure}
	\includegraphics[width = 3.5in]{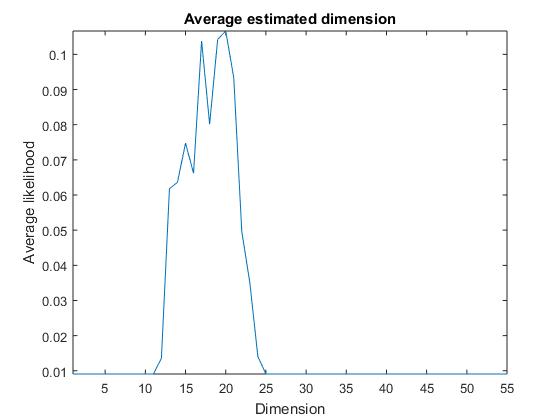} 
	\includegraphics[width = 3.5in]{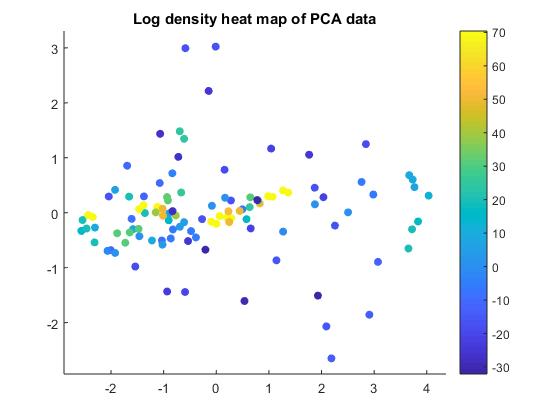} 
	\includegraphics[width = 3.5in]{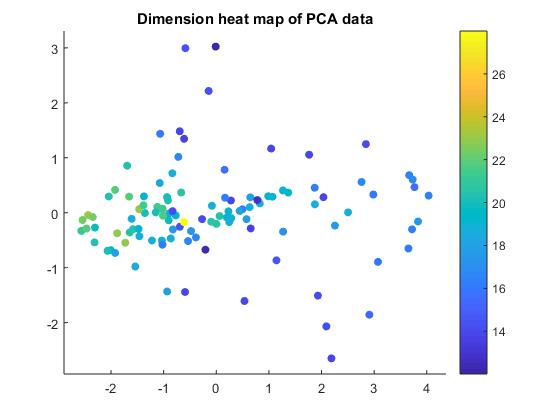}
	\caption{Dimensionality of the space of syntactic variables for the Niger-Congo language family (SSWL data, $\alpha=1/3$).
	\label{NigerCongoDimFig} }
\end{figure}

%\begin{figure}
%	\includegraphics[width = 3.5in]{atlantic-congo_alpha033333avedim.jpg} 
%	\includegraphics[width = 3.5in]{atlantic-congo_alpha033333heatden.jpg} 
%	\includegraphics[width = 3.5in]{atlantic-congo_alpha033333heatdim.jpg}
%	\caption{Dimensionality of the space of syntactic variables for the Niger-Congo language family (SSWL data, $\alpha=1/3$).
%	\label{NigerCongoDimFig} }
%\end{figure}

\begin{figure}
	\includegraphics[width = 3.5in]{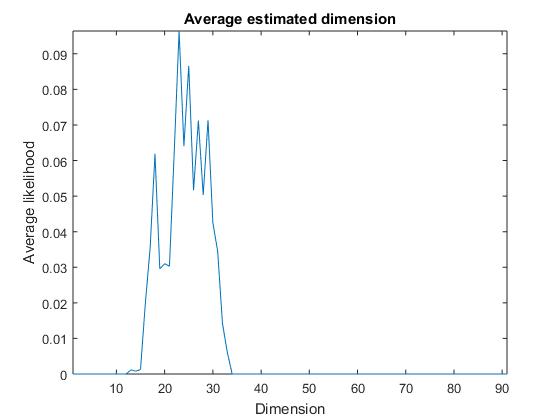} 
	\includegraphics[width = 3.5in]{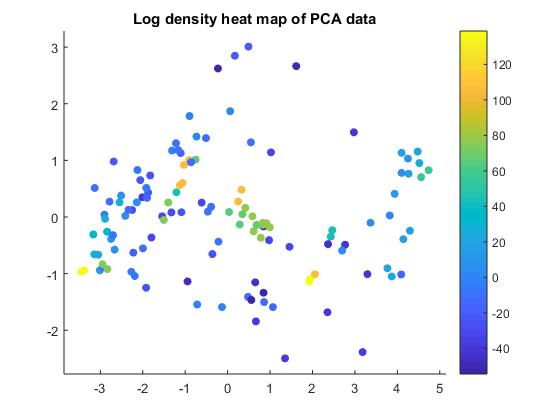} 
	\includegraphics[width = 3.5in]{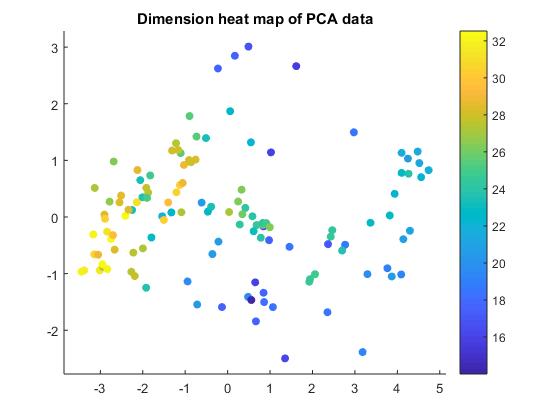}
	\caption{Dimensionality of the space of syntactic variables for the Indo-European language family (SSWL data, $\alpha=1/3$).
	\label{IEDimFig} }
\end{figure}

%\begin{figure}
%	\includegraphics[width = 3.5in]{indoeuropean_alpha033333avedim.jpg} 
%	\includegraphics[width = 3.5in]{indoeuropean_alpha033333heatden.jpg} 
%	\includegraphics[width = 3.5in]{indoeuropean_alpha033333heatdim.jpg}
%	\caption{Dimensionality of the space of syntactic variables for the Indo-European language family (SSWL data, $\alpha=1/3$).
%	\label{IEDimFig} }
%\end{figure}

One can also consider the question of whether there are relations between syntactic parameters that
only hold within certain language families rather than being universal across languages.
If this is the case, then we expect to find different dimension estimates for the space of
syntactic structures when only the coordinates of a given syntactic parameter that correspond
to languages in a given family are retained. 

\smallskip

We consider the subset of the LanGeLin data for the Romance languages and we
look at the dimensionality of the space of syntactic parameters restricted to the set of Romance
languages, see Figure~\ref{RomanceDimFig}.
We find a peak at a much lower dimension, $d=5$, than for the full data set across all languages, which has a peak around $d=15$. 
This implies the presence of additional relations between the syntactic parameters that are language specific
rather than universal across language families. 

\smallskip

We see a similar phenomenon when we consider the SSWL data and we restrict the syntactic variables
to certain language families. For example we find that the dimension of the space of syntactic variables
for the Niger--Congo languages peaks at a lower value, around $20$, than the dimension for the space of Indo-European
languages, where the dimension estimate peaks around $23$ (see Figure~\ref{NigerCongoDimFig} and Figure~\ref{IEDimFig}), 
suggesting that there are slightly more language-family-specific relations between the syntactic variables
for the Niger--Congo languages than for the Indo-European languages.

\smallskip
\subsection{Density and dimension estimates of language families}

\begin{figure}
	\includegraphics[width = 1.8in]{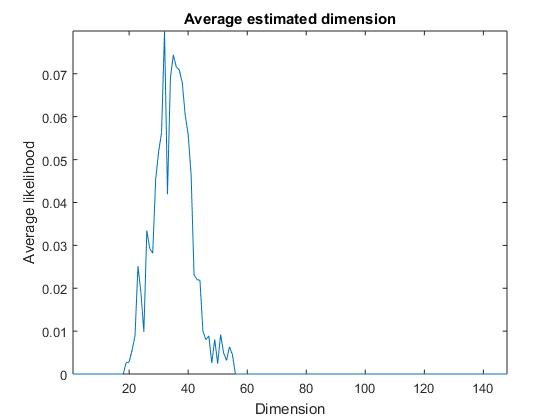} 
	\includegraphics[width = 1.8in]{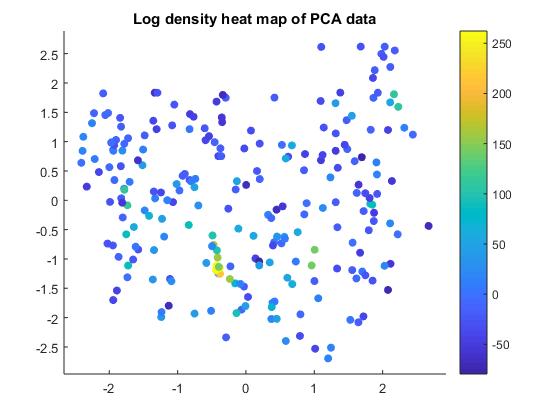} 
	\includegraphics[width = 1.8in]{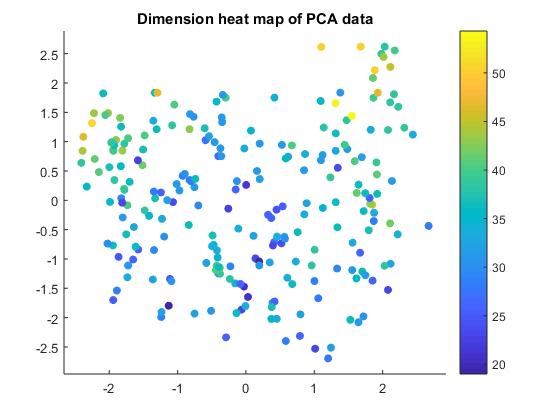} \\

	\includegraphics[width = 1.8in]{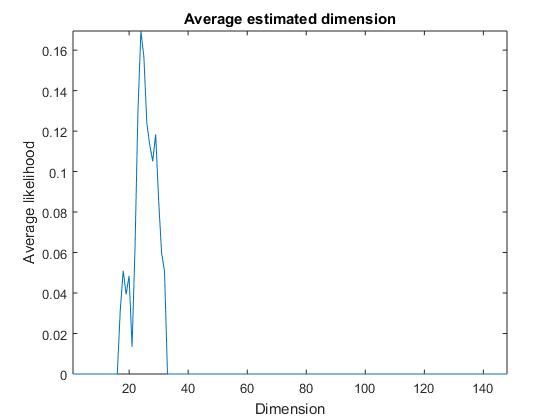} 
	\includegraphics[width = 1.8in]{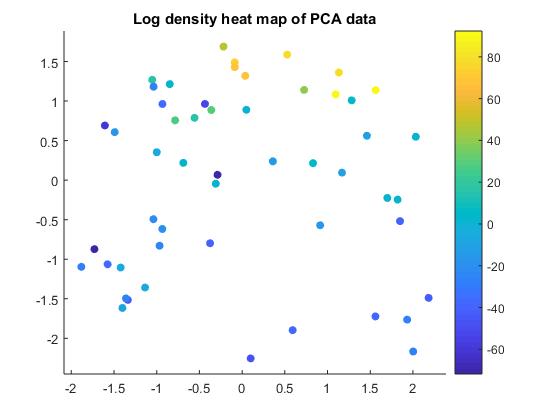} 
	\includegraphics[width = 1.8in]{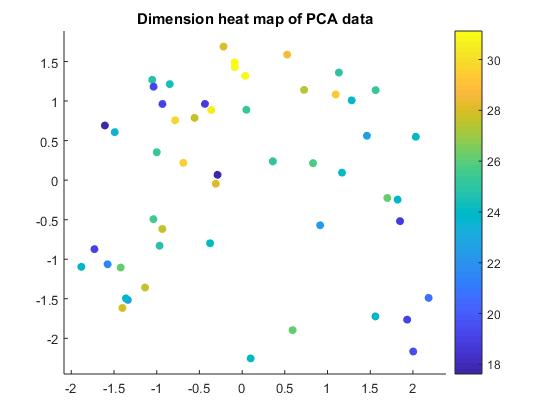} \\

	\includegraphics[width = 1.8in]{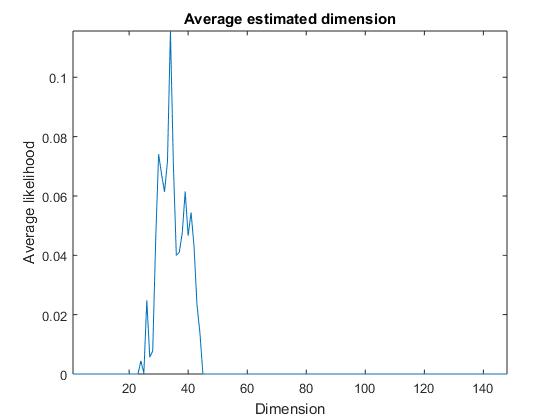} 
	\includegraphics[width = 1.8in]{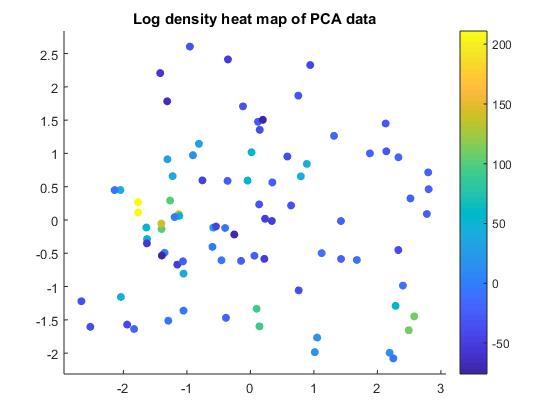} 
	\includegraphics[width = 1.8in]{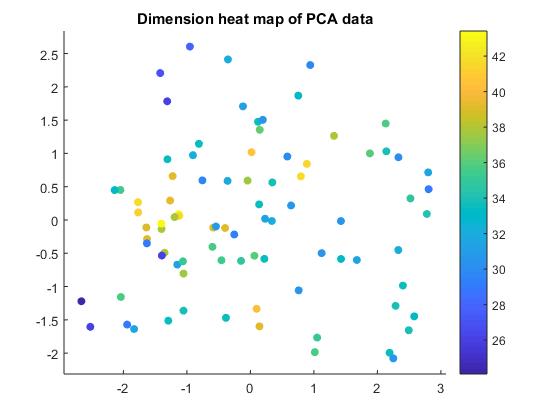} \\

	\includegraphics[width = 1.8in]{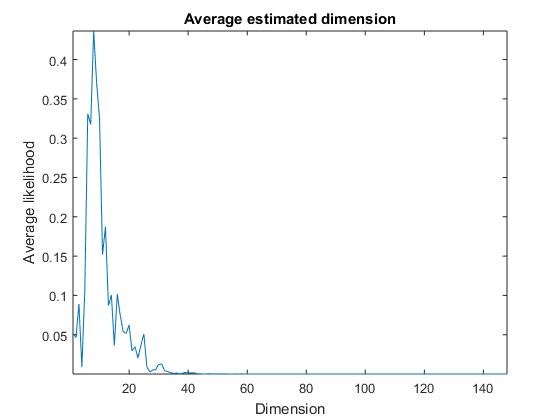} 
	\includegraphics[width = 1.8in]{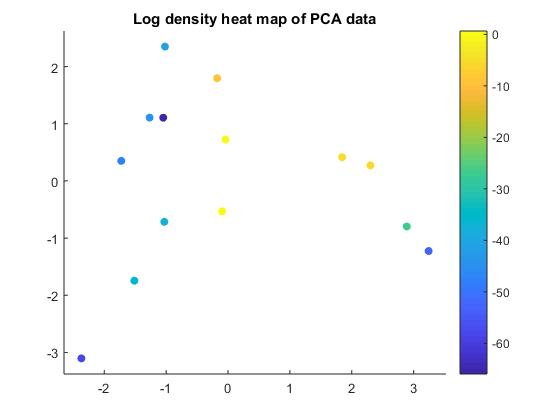} 
	\includegraphics[width = 1.8in]{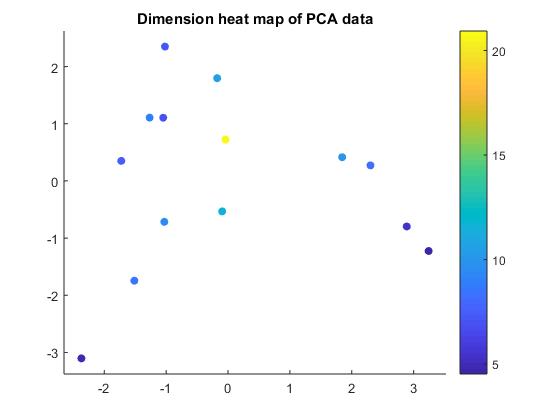} \\

	\includegraphics[width = 1.8in]{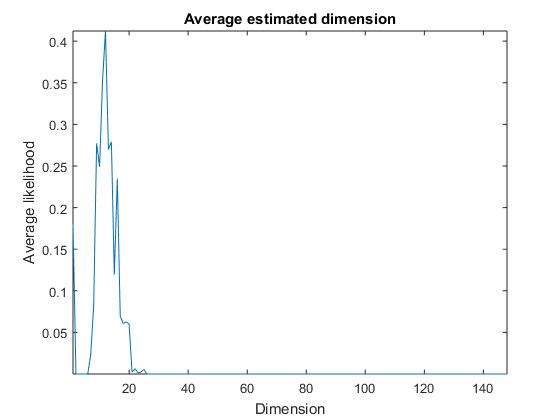} 
	\includegraphics[width = 1.8in]{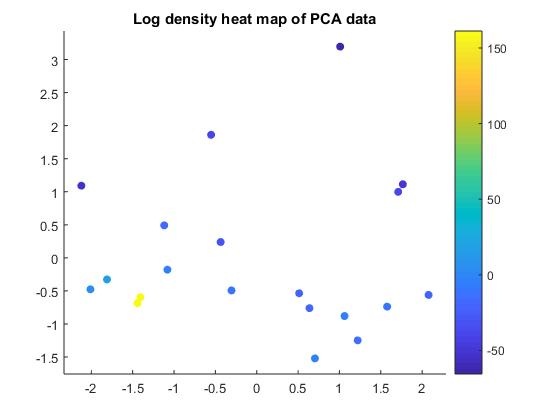} 
	\includegraphics[width = 1.8in]{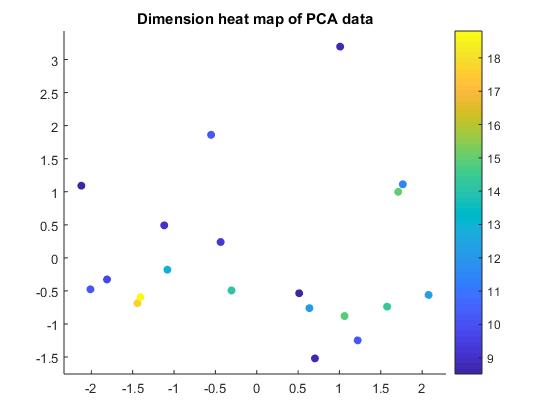} \\

	\caption{Dimensionality of the space of languages for the full database, the Niger--Congo family,
	the Indo-European family, the Afroasiatic family, and the Austronesian family  (SSWL data, $\alpha=1/3$).
	\label{LangDimFig}}
\end{figure}

In the data matrix where data points are languages and coordinates 
are syntactic parameters/variables, one can also run the estimates
of density and dimension discussed above. In this case, rather than
describing relation between syntactic variables that are either universal
across language families or family-specific, the estimates give us
information about language relatedness through information about
how spread out (in the sense of density and dimension) languages
within a given family are then regarded in terms of their syntactic structures.

\smallskip

Thus, we consider the data points of certain given language families and
we run the same dimensionality analysis discussed above, see Figure~\ref{LangDimFig}.
When looking at the dimension estimates for the different language sub-families 
at $\alpha = \frac{1}{3}$ we see that we usually get clear peaks for the dimension estimation 
(versus larger $\alpha$ values, or larger data sets).  Regarding the density maps, the estimations 
are less sensitive to the alpha value than the dimension estimates. 

\smallskip

Over the entire set of languages of the SSWL database we find a dimension estimate around $38$.
The Niger--Congo family has dimension peak around $23$, while the 
Indo-European language families has dimension estimate that is closest to the overall estimate, also around $38$.
The Afro-Asiatic and Austronesian language families have estimated dimensions respectively around
$8$ and $12$. 

\smallskip

\begin{figure}
	\includegraphics[width = 1.8in]{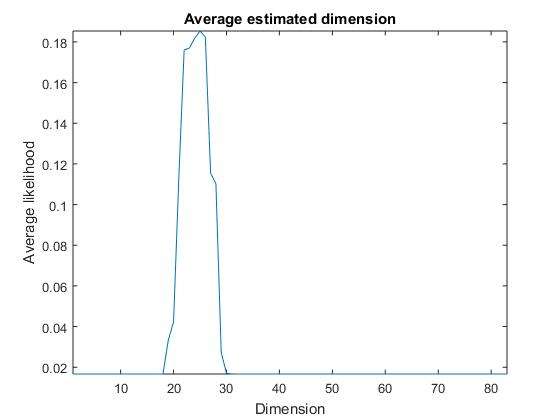} 
	\includegraphics[width = 1.8in]{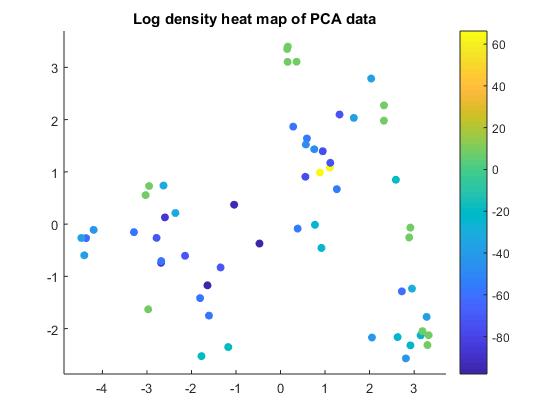} 
	\includegraphics[width = 1.8in]{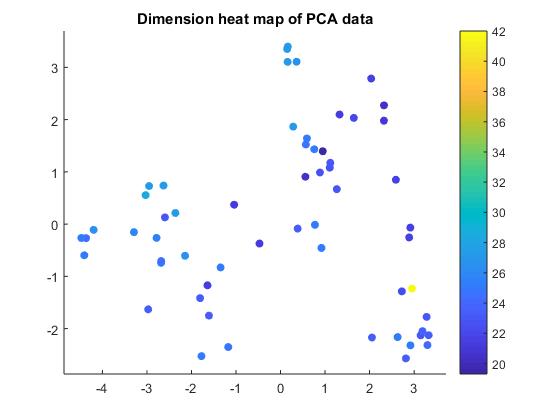} %\\
	
\caption{Dimensionality of the space of languages for the full database  (LanGeLin data, $\alpha=1/3$).
	\label{LangDimFig2} }
\end{figure}

For the LanGeLin data, the dimension estimate for the space of languages is illustrated in Figure~\ref{LangDimFig2}.
The estimated dimension for the full set of languages in the database peaks around $25$.

\smallskip
\subsection{The Altaic and Ural-Altaic hypothesis} \label{UrAltDim}

\begin{figure}
	\includegraphics[width = 1.8in]{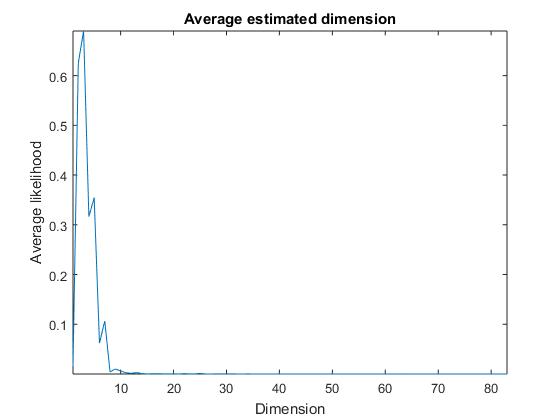} 
	\includegraphics[width = 1.8in]{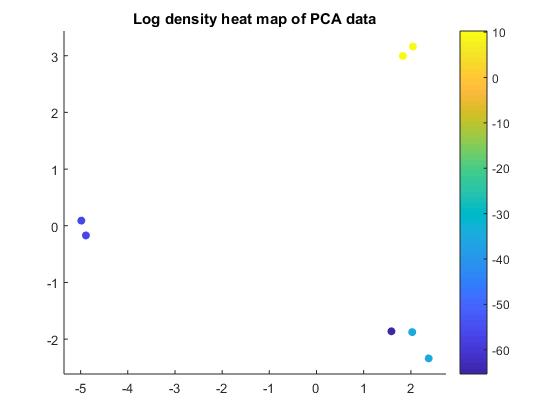} 
	\includegraphics[width = 1.8in]{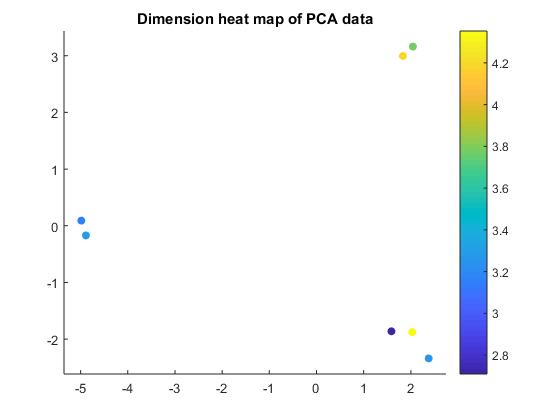} \\
	
	\includegraphics[width = 1.8in]{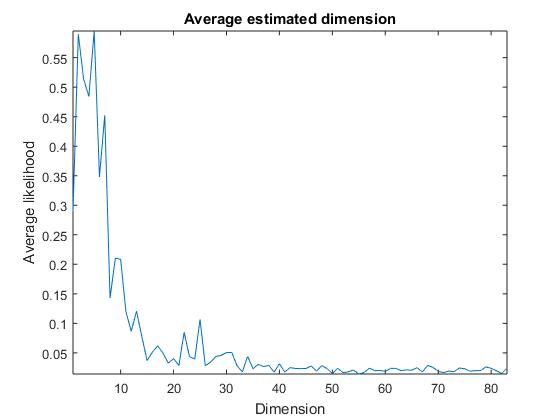} 
	\includegraphics[width = 1.8in]{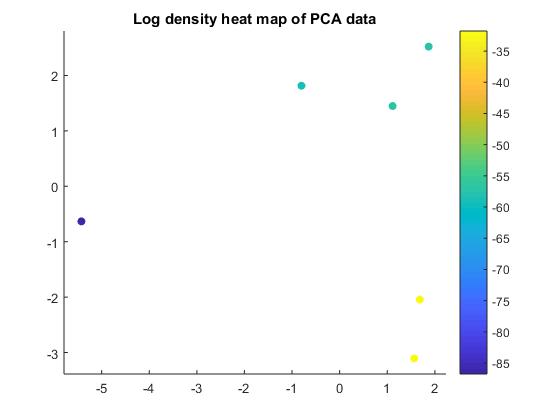} 
	\includegraphics[width = 1.8in]{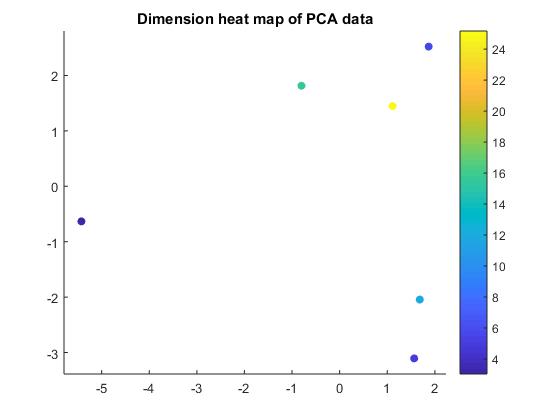} \\

         \includegraphics[width = 1.8in]{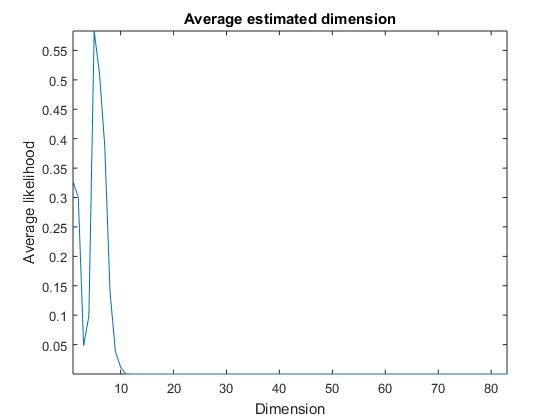} 
	\includegraphics[width = 1.8in]{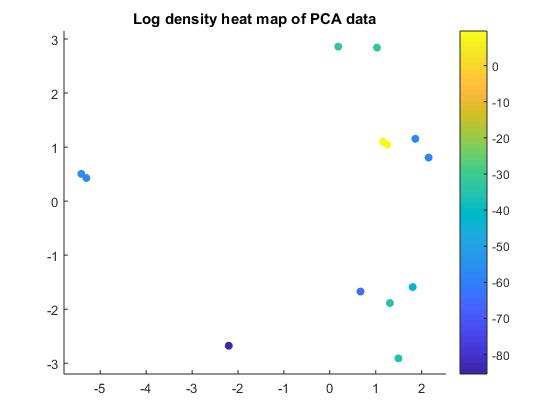} 
	\includegraphics[width = 1.8in]{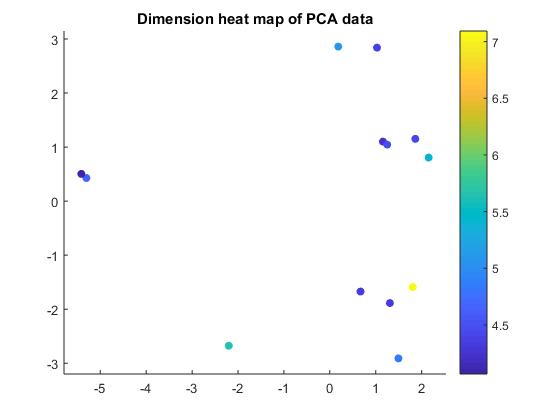} \\

\caption{Dimensionality of the space of languages for the Altaic, Uralic, and Ural-Altaic languages  
(LanGeLin data, $\alpha=1/3$).
	\label{LangDimUrAltFig} }
\end{figure}

We discuss here the Altaic and Ural-Altaic hypothesis in terms of the dimensional analysis for these
language families. This perspective, although carried out with a different mathematical method, is similar 
to the point of view discussed in \cite{LongoCeo} in their analysis of the Ural-Altaic hypothesis through
the LanGeLin data. We use the same LanGeLin data here and we compute the dimension estimates
separately for the Uralic and the Altaic languages and for the two sets together. The dimension distribution
does not seem to rule out an Altaic hypothesis. In fact we find a clearer single peak for the estimated
dimension for the Altaic languages than for the Uralic ones. However, when the Uralic and the Altaic families
are joined together, the dimension estimates shows two clear separate peaks. This may be interpreted as
a cautionary indicator about the possibility of a single Ural-Altaic family hypothesis, see Figure~\ref{LangDimUrAltFig}.

\medskip
\section{Relations and the topology of the space of syntactic parameters} 

In this section we return to the question of investigating relations between syntactic variables and identifying the
topology and geometry of the space of syntactic parameters, previously discussed in \cite{Kazakov}, \cite{OBM}, \cite{ParkMa}, \cite{ShuMar}. We have already discussed in Section~\ref{DimSec} how the dimensionality analysis identifies a space of dimension around $d=12$ inside the space of the LanGeLin syntactic parameters, and a space of dimension around $d=30$ inside the $116$-dimensional space of the SSWL syntactic variables, that represent the dimension of the 
submanifold on which the data of syntactic features are distributed. Thus, we know from these estimates that one should expect a high degree of dependence between the syntactic variables. As we have seen with dimensional analysis, there are additional family-specific relations that further lower the dimension when the data are 
restricted to specific language
families. Here we seek to identify the relations between syntactic variables more explicitly by computing the persistent homology of the data set (organized with syntactic features as data points)
and the tree of the persistent connected components.  

\smallskip

When we consider the data points as syntactic features with coordinates giving the values of
a feature over a set of languages and we compute the persistent homology of these data set,
the clusters given by the $H_0$ identify subsets of syntactic features that are in close proximity to each other, in the sense that these syntactic features tend to be similarly expressed over subgroups of languages. 
It is clear that proximity between syntactic features detects some kind of relations between them (for example, if one is expressed in a language, so does the other). However, not all types of relations between syntactic parameters will be of this form: while persistent $H_0$ detects parameters that tend to align, it does not detect anti-alignments, nor it detects other forms of entailment such as those explicitly identified in the LanGeLin data. In this section we examine the proximity between the parameters using the phylogenetic tree built based on the $H_0$ computations and we compare briefly the resulting relations that we identify in this way to previous work aimed at identifying relations between syntactic features by different methods. In particular, we compare our results to the heat kernel analysis of \cite{OBM} 
and to the analysis of the relations between the LanGeLin syntactic parameters via machine learning algorithms of \cite{Kazakov}.

\smallskip
\subsection{Our tree construction algorithm} \label{treealgoSec}

The main steps in reconstructing the persistent components trees are Principal Component Analysis (PCA), computation of the clusters and an inclusion-based tree reconstruction. 
Here is a step-by-step explanation of the Matlab script $\text{TreeFromDataGUI.m}$ we wrote and used in this analysis. 

\begin{itemize}
	\item First the data is loaded and the files are labeled. Then, empty spaces in the data are filled with the midpoint of the data (empty spaces in the SSWL are filled with $\frac{1}{2}$ and in the Longobardi with $0$ value).
	We also discuss trees obtained by filtering out languages in the SSWL that are too incomplete, see below.
	
	\smallskip
	
	\item The second step is the PCA. It finds a PCA basis and takes up to the percent variance we choose (for our use it is either $60 \%$ or $80 \%$). 
	
	\smallskip
	
	\item Computes the Euclidean distance pairwise and finds the critical radius in which the simplicial complex becomes completely connected (i.e when we have one cluster). As noted previously, after the process of  PCA our data becomes continuous, and therefore the metric we use is the Euclidean metric (rather than Hamming distance which is used in a discrete data set). 
	
	\smallskip
	
	\item Computes all the clusters in small incremental radii and assembles the persistent components tree based on inclusion. If $C_r$ is the set of all clusters at radius $r$ and $C = \sqcup _r C_r$: in $C$, cluster $C_i$ is a child of cluster $C_j$ if $C_i \subseteq C_j$ and there is no cluster $C_k$ such that $C_i \subseteq C_k \subseteq C_j$.  
\end{itemize} 

\medskip

In this section we will present a detailed analysis of persistent components trees of two different data sets, the SSWL data set and the LanGeLin data set, where 
we view the data points (in transpose form) as being the individual syntactic features and their coordinates being the values by languages. We use the resulting
trees to identify structures of relations between the syntactic variables. 
The trees are constructed using the computational method described above, based on PCA and persistent homology computations, specifically the group $H_0$ 
that determines the persistent connected components of the data. We use the number on the vertex to refer to the same cluster number and also to the subtree that it forms. 

\smallskip
\subsection{Comparison with the heat kernel method}\label{HeatCompareSec}

In \cite{OBM} the Belkin–Niyogi heat kernel dimensional reduction method \cite{BeNi}
is used in order to analyze the relations between the syntactic parameters. This method gives 
more information than one gets by only using PCA, since it also gives connectivity relations 
(and not only proximity). Parameters that exhibit high connectivity to other 
parameters (with a higher valence in the resulting graph) are interpreted 
as more ``dependent" and vice versa.  The analysis of \cite{OBM} is performed on both
the LanGeLin and the SSWL data sets.

\begin{figure} [!htb]'
%\begin{center}
\includegraphics[width = 7.5in, angle=90]{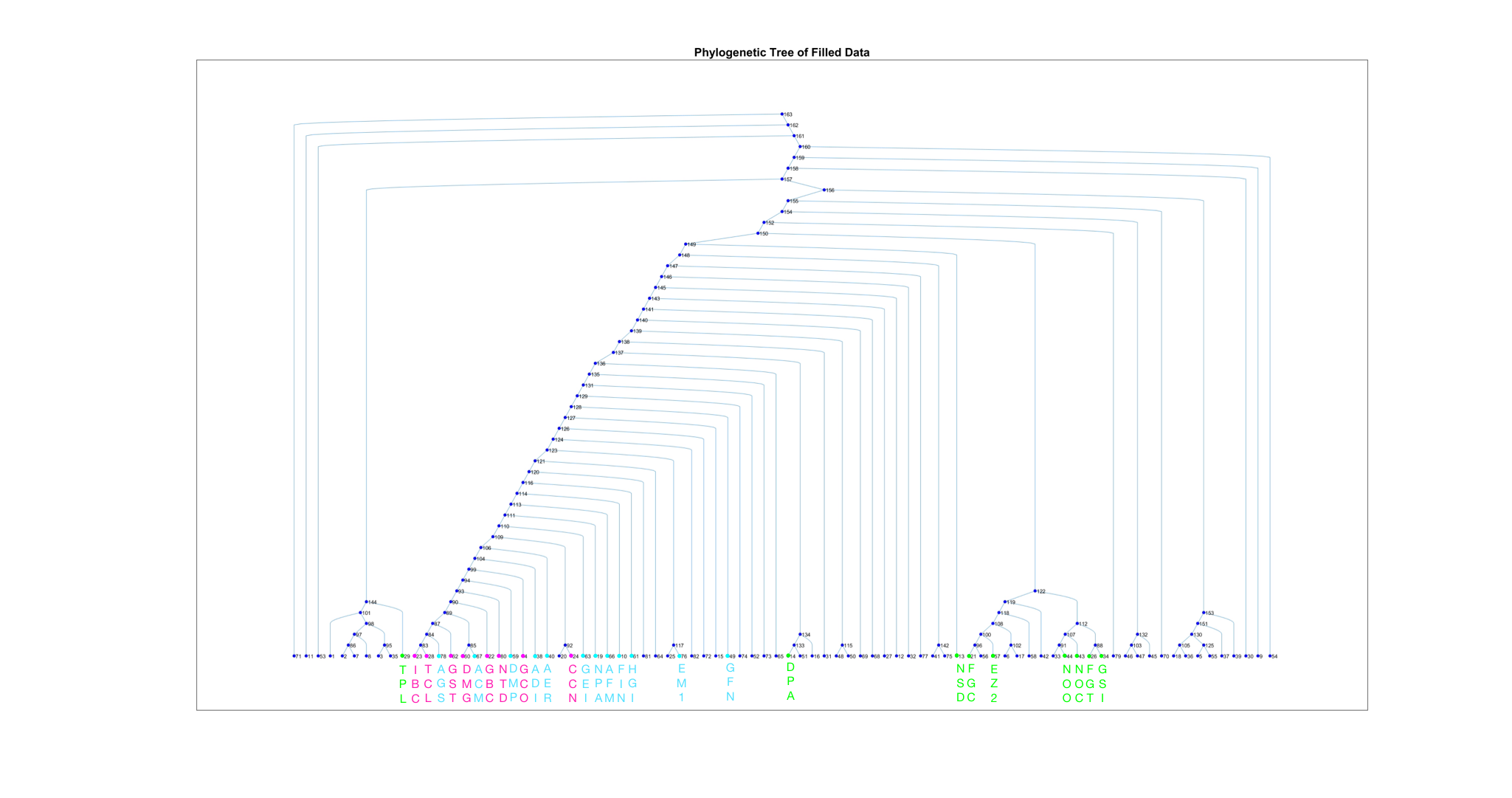} \caption{Persistent connected components tree for the syntactic 
parameters of the Longobardi data set and comparison with heat kernel clusters. \label{Longo_parameters_tree} }
%\end{center}
\end{figure}
\begin{figure} [ht]
 	\includegraphics[width = 7.53in, angle=90]{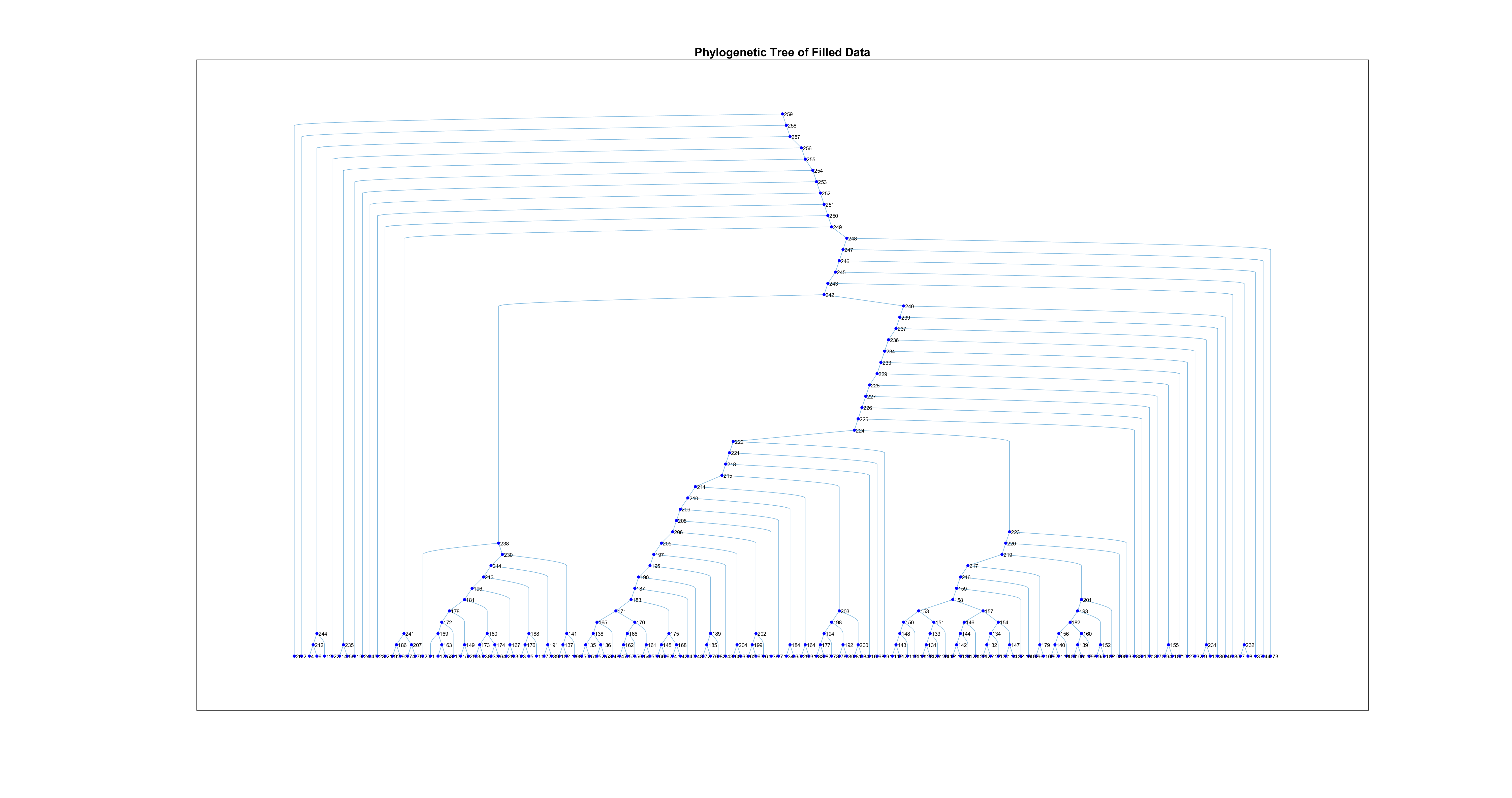}
\caption{Persistent connected components tree for the syntactic variables of the SSWL data set.
		\label{SSWL_trans_tree}}
 \end{figure}

\begin{figure} [ht]		
 	\includegraphics[width = 7.5in, angle=90]{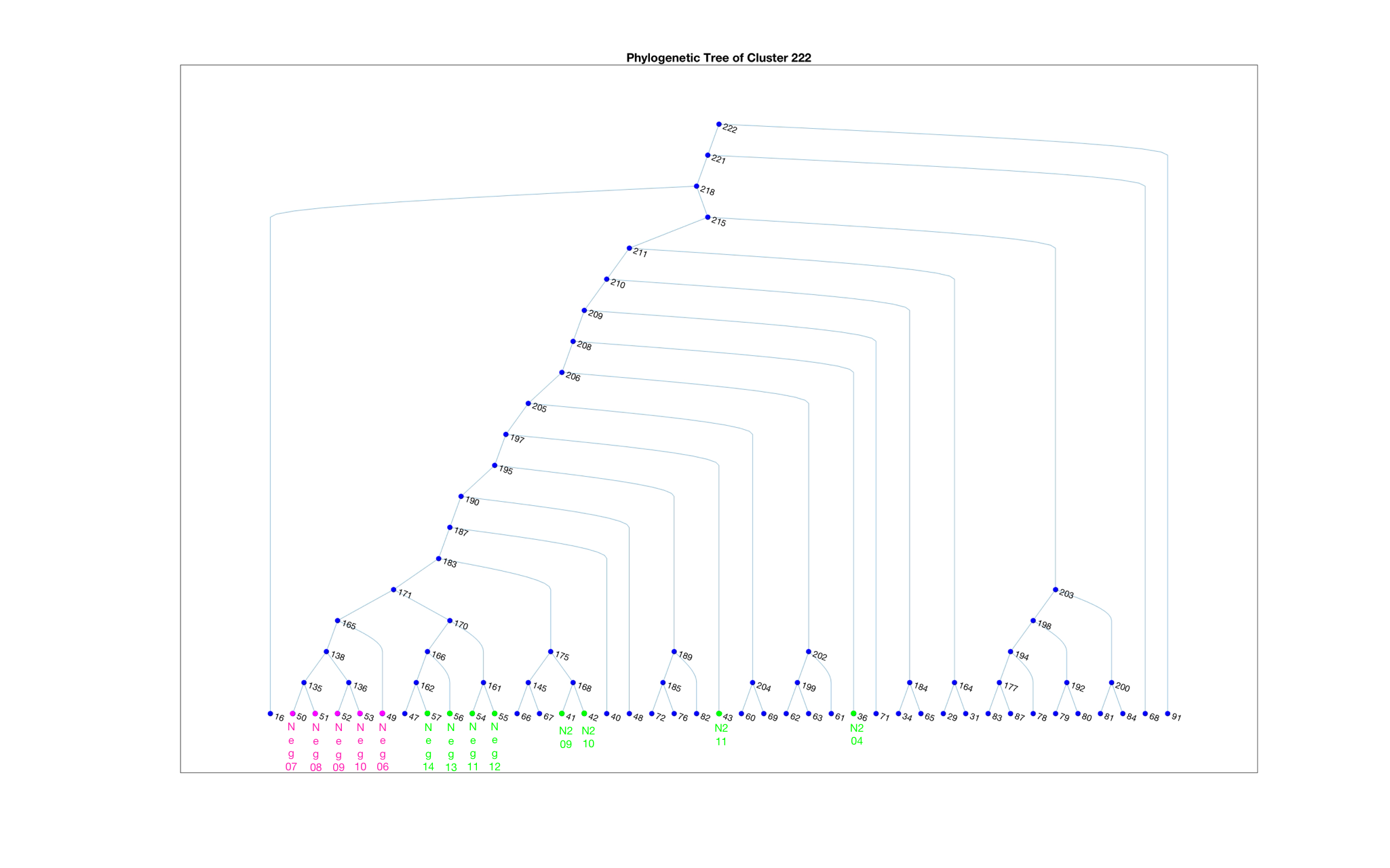}
 \caption{Persistent connected components tree for the syntactic variables of the SSWL data set, cluster N.~$222$, and heat kernel clusters.
		\label{SSWL_trans_tree2}}
 \end{figure}

\smallskip

In our persistent topology analysis the LanGeLin data set has one main cluster which contains most of the points (N.~$144$), another small cluster (N.~$156$) and some singletons. Cluster $144$ contains two subtrees (N.~$122$ and N.~$153$) and apart from that the other points are joined as singletons or $2$-leaf trees, see Figure~\ref{Longo_parameters_tree}.

\smallskip

In the heat kernel analysis of \cite{OBM}, the Longobardi data set appears to have two main clusters 
(for scale size $\epsilon = 15$). A big one composed by two sub-clusters consisting of  
the parameters EZ$2$, NOO, NOC, FGC, TPL, FGT, NSD, DPQ, GSI, HMP (the pink-colored points in Figure~\ref{Longo_parameters_tree}) and AFM, ACM, AGS, GCO, FIN, GEI, HGI, NPA, GFN, DMP and AER (the blue-colored points in Figure~\ref{Longo_parameters_tree}). Moreover, a smaller cluster contains DMG, GCO, GST, BAT, CCN, GBC, IBC, NTD,  and TCL (the green-colored points in Figure~\ref{Longo_parameters_tree}). 
We can see that, except for a few points, the two main clusters that are visible in the heat kernel analysis are also well represented in our persistent components tree, although the bigger cluster is not seen in the tree as containing the two sub-clusters as separate subtrees, but rather mixing them. 

\smallskip

To avoid problems due to the incompleteness of the SSWL database and the presence of poorly mapped languages,
we filter the SSWL data by restricting the set of languages to only those that are at least $50\%$ mapped. 
The persistent connected components analysis shows a very different clustering structure in this case, see
Figure~\ref{SSWL_trans_tree}. It contains $3$ main subtrees, corresponding to cluster N.~$222$, N.~$238$ and N.~$203$ (see Figure~\ref{SSWL_trans_tree}) 
but it also shows a lot more substructures, as opposed to only one main subtree and mostly 
singletons in the LanGeLin data set. 

\smallskip

When comparing to the clustering of SSWL syntactic features obtained in \cite{OBM} using the heat kernel method, we see that, except for two of the syntactic variables involved (which appear in the tree as clusters N.~$206$ and N.~$203$), all the features involved in the relations obtained in \cite{OBM} are contained in the subtree corresponding to cluster N.~$222$ (colored points in Figure~\ref{SSWL_trans_tree2}). Using the heat kernel method there are two main clusters that are formed; at $\epsilon = 15$ with Neg$06$--Neg$10$, and at $\epsilon = 20$ with Neg$06$--Neg$10$ and new points that are added, which are N$209$--N$211$, N$203$, N$204$, N$206$, Neg$11$--Neg$14$. In our persistent connected components tree the points from the 
$\epsilon = 15$ case are clustered together (pink-colored points in Figure~\ref{SSWL_trans_tree2})
and the points from the second cluster ($\epsilon = 22$) are also mostly grouped together (green-colored 
points in Figure~\ref{SSWL_trans_tree2}). Thus, it seems that the relations obtained via the tree of
persistent connected components capture the clustering information obtained via the heat kernel method.
The tree structure contains additional information that was not seen by the heat kernel analysis.

\begin{figure}
	\subfloat[SSWL data]{\includegraphics[width = 5in]{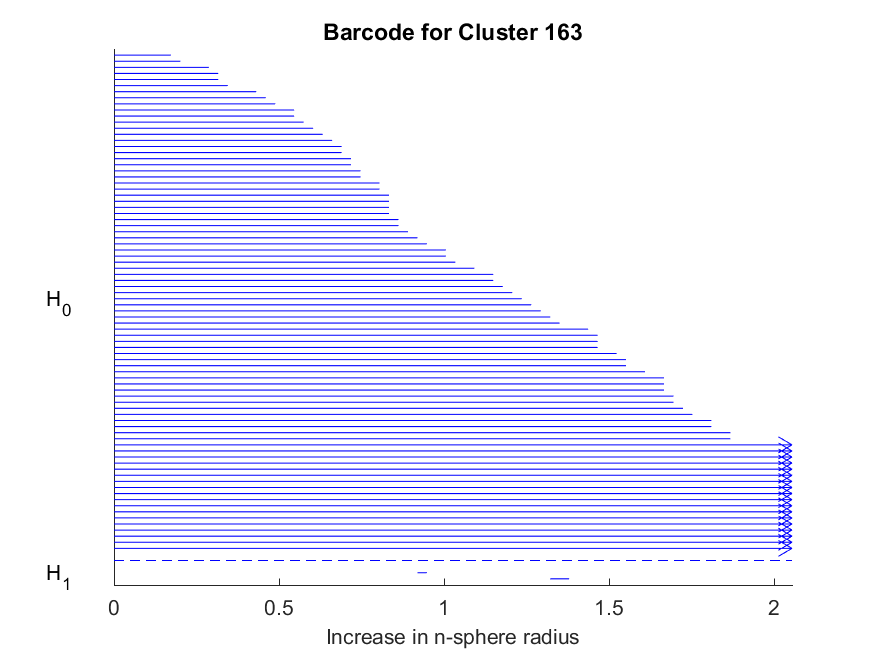}} \\
	\subfloat[LanGeLin data]{\includegraphics[width = 5in]{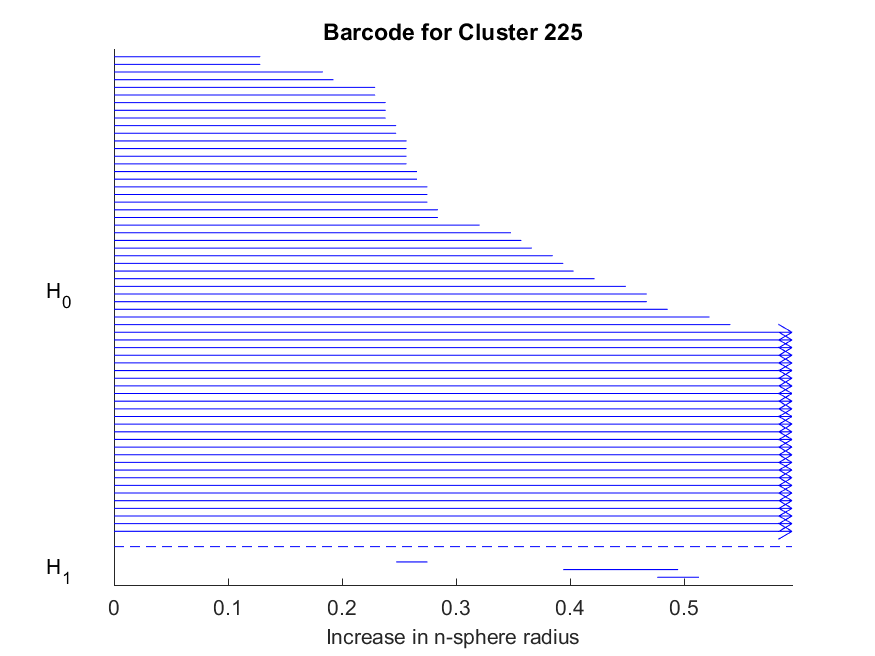}} 
	 \caption{Truncated barcode diagrams for the data of syntactic features (syntactic parameters) for the SSWL and the LanGeLin data showing persistent $H_1$.
	\label{H1paramsFig} }
\end{figure}

\smallskip
\subsection{Comparison with machine learning methods}

In \cite{Kazakov} machine learning methods are used to construct a dependency graph for the syntactic parameters of the LanGeLin data, in which each parameter is represented by a vertex and relations between the parameters are represented by directed edges in the following way: if a parameter $x$ can be predicted by knowing all values of parameter $y$ then there is an edge from $y$ pointing to $x$. 

\smallskip

The relation investigated in \cite{Kazakov} can be of a different nature from the kind of alignment
relations that the persistent $H_0$ (or to some extent the heat kernel method) detect. Indeed the
graph of relations obtained in \cite{Kazakov} also involves entailment relations where the value of
one parameter may force another parameter to either anti-align or to become undefined. This type
of relations would not be visible through the persistent connected components. 

\smallskip

Moreover, another difference is the fact that the method of \cite{Kazakov} identifies relations that
hold ``exactly" over the entire set of languages in the database, while the relations identified by the
topological connected components are proximity relations that hold ``approximately" over a 
sufficiently large subset of languages. In other words it is the difference between saying that the
data points lie exactly on a submanifold defined by the relations or that the relations define a
submanifold that interpolates the data with some precision (the data points are close to the
submanifold but not necessarily on it). 

\smallskip

Indeed we see for example that pairs of parameters in the graph of relations of \cite{Kazakov} 
that have an arrow between them in both directions, such as
\{GCN, GFN\}, \{FGN, GCO\} and \{DMP, DMG\}
correspond, respectively, to nodes number $\{48,49\}$, $\{3,4\}$ and $\{59, 60\}$ in our
tree of persistent connected components, but these nodes are not placed 
in close proximity to each other in the tree.

\smallskip
\subsection{First homology of the space of syntactic variables} 

The computation of the persistent topology barcode diagrams for the SSWL data is computed after filtering the
data as indicated above. The barcode diagrams show the presence of non-trivial persistent generators of the $H_1$
homology group, though no higher dimensional persistent homology $H_k$ for $k\geq 2$ is detected.

\smallskip

Figure~\ref{H1paramsFig} shows the barcode diagrams (truncated at a certain scale) for the SSWL syntactic variables
data (restricted to the Indo-European languages) and for the LanGeLin data (all languages). 
It appears that the space of the SSWL variables exhibits more persistent $H_1$-generators
than the space of the LanGeLin syntactic parameters. We will leave to future work to carry out a more in depth investigation
of these $H_1$-generators and the linguistic significance of the relations between syntactic variables that they represent.
We will also be discussing elsewhere the relations between syntactic parameters, detected through the persistent $H_0$
and $H_1$ for relations that are specific to certain language families. The presence of such family-specific syntactic relations is indicated by our dimensional analysis.

\medskip
\section{Language relatedness trees from persistent components}\label{H0Sec}

We now consider again our data sets so that data points are languages and their 
coordinates are given by the values of syntactic variables/parameters for a given
language. 

\smallskip

One of the main goals of the present investigation is to test whether the persistent components (persistent $H_0$)
in the topological data analysis provide a valid independent method for the computational reconstruction of
phylogenetic trees of language families. We will show here that the trees obtained from the persistent connected
components are closely related to phylogenetic trees, in the sense that they largely provide an accurate splitting into
subfamilies, but they also have significant differences. In particular, while in the usual phylogenetic trees of language
families one should interpret the inner nodes of the trees as ancient languages that are ancestors to the modern
languages, in the case of trees arising from the persistent connected components the inner nodes indicate the
hierarchical structures of clusters, which is related to the subdivision into subfamilies, but does not represent an
actual branching of an evolutionary process. In particular, the inner nodes cannot in general be identified with individual 
ancestors in the form of ancient languages or proto-languages, although we will see some specific examples where 
such an interpretation applies.      In general, the inner vertices of our trees only represent a hierarchical 
structure of clustering of syntactic features. 

\smallskip

Given the family of Vietoris--Rips complexes associated to the data at varying scales, one can
compute the corresponding zeroth homology $H_0$. This counts the number of connected 
components of the Vietoris--Rips complex at that scale. Clearly, for sufficiently small scales,
each data point will be a separate connected component, hence there is simply a different
generator of the $H_0$ for each data point and no higher dimensional homology. 
At the other end of the interval of scales, for 
sufficiently large scales, all the points in the data set will be contained in a single ball of
diameter the given scale around any of the points. In this large scale range the Vietoris--Rips 
complex consists of a single simplex of dimension equal to the total number of data points.
This is a connected and contractible space hence its $H_0$ has a single generator and all
the higher homology vanishes. In between these two extremes, the different singletons 
of the low scale picture begin to merge, in a certain order at certain values of the scale
parameter, where the Vietoris--Rips complex correspondingly jumps by adding $1$-simplices
and possibly higher dimensional simplices as well. One can compute a tree that takes care
of the order in which points are joined together by simplices in the Vietoris--Rips 
complex as the scale increases. The question is then whether the tree obtained in this
way is closely related to the phylogenetic tree of historical language change within
specific language families. 

\smallskip

Since vertical relations between the languages are not reflected in the persistent components tree,
which accounts for the hierarchical merging of clusters only, how to interpret the location of the old 
languages is a question by itself. Old languages appear as leaves of the tree rather than as root vertices
of certain subtrees, so when comparing the persistent components trees with phylogenetic trees we should 
imagine them as lying above the other languages in their subtrees of the previous sub-clusters they merge with.

\smallskip

\subsection{Various computational methods for phylogenetic trees} \label{subsec_phylo_methods}

The reconstruction of phylogenetic trees of language families is an integral part of historical linguists research since its beginning. In recent years, due to a vast research in the field of computational biology, more and more methods for phylogenetic tree reconstruction have been introduced. It is a point of interest for linguists, and other scientists alike, whether these methods can yield good results when applied to linguistic data, such as linguistic characters (lexical, morphological and phonological) and syntactic parameters. In this paper we are interested only in the latter type of data. The three main methods for tree reconstruction that have been used in this setting are distance based methods (such as neighborhood joining and UPGMA) and character based methods (such as maximum parsimony and maximum likelihood), as well as phylogenetic algebraic geometry.

\begin{itemize}

	\item Distance based methods: a phylogenetic tree is reconstructed based on the distances between the data points. 
	Two very well known distance based methods are the neighbor joining method and UPGMA (Unweighted Pair Group Method with Arithmetic Mean). 
	
	\item Character based methods: a phylogenetic tree is reconstructed using a character table which describes the data. The goal is to reconstruct a tree where similar character values occur near each other. The most used methods here are Maximum Parsimony and Maximum Likelihood. 
		
	\item Phylogenetic Algebraic Geometry: a phylogenetic algebraic variety is associated to a hidden Markov
	model on a tree, with the data providing the probability distribution at the leaves (the visible nodes). The
	phylogenetic invariants (generators of the ideal defining the phylogenetic variety) are used to evaluate
	how well the data fit the model.
\end{itemize}

Phylogenetic algebraic geometry is the study of algebraic varieties derived from phylogenetic trees (which are usually associated with an evolution of biological sequences), see \cite{PaStu}. In algebraic phylogenetics, a tree $T$ is associated with a geometric object $(V_T,x_{T,\mathcal{P}})$ where $V_T$ is an algebraic variety determined by the topology of the tree $T$ and $\mathcal{P}$ is a boundary distribution on the leaves of $T$, which is a polynomial function of the model parameters. The point $x_{T,\mathcal{P}} \in V_T$ lies on the sublocus $V_T(\R_{+}) \subset V_T(\R)$. In some cases, when $V_T$ is a classical well studied algebraic variety, it is possible to get an interesting information about underlying structures of $T$ from the sub-variety $x_{T,\mathcal{P}}$ lies on. For example, it can provide a connection between the distribution $\mathcal{P}$ and the splittings of the tree into sub trees. In the process, trees are generated using some algorithms (such as DNA parsimony algorithms), and using algebraic phylogenetic methods, phylogenetic invariants are calculated. Using these invariants and estimates of the
Euclidean distance in the ambient affine space between the point $\mathcal{P}$ and the phylogenetic variety $V_T$, a best candidate tree is chosen. As shown in \cite{OSBM}, trees that were reconstructed via this method from SSWL and LanGeLin data were correctly corresponding to the actual phylogenetic trees known to historical linguists. The authors also argued that this method is better at generating phylogenetic trees from syntactic parameters than other distance based methods. 

\smallskip

There is some controversy over the use of computational methods for the reconstruction of language phylogenetic trees, especially for the Indo-European family,
see \cite{PerLe}. In \cite{ShuMar2} the authors discuss the advantages 
of the algebro-geometric approach used in \cite{OSBM} over other distance 
based methods. It is shown that the Algebro-Geometric approach yields better results, especially when additional information, such is the subdivision into 
subfamilies and the position of the old languages is taken into account. 
We discuss here our construction of persistent connected components trees and their comparison with known phylogenetic trees.

\smallskip

It is important to stress here the fact observed in \cite{OSBM} that historical phylogenetic trees can be
correctly reconstructed from the (filtered) SSWL data and the LanGeLin data, since this implies that the
discrepancies we will outline below between the trees we construct using persistent connected
components and the phylogenetic trees are not due to some inadequacy of the data, but show a genuine
intrinsic difference between the information encoded in the hierarchical structure of the clustering of
the connected components in the persistent $H_0$ and the trees of historical language development.

\smallskip
\subsection{Persistent components trees and language relatedness}

In this section we present a detailed analysis of persistent components trees for both the SSWL and the LanGeLin data sets, where data points are
languages, with their coordinates given by the values of their binary syntactic variables (SSWL) or syntactic parameters (LanGeLin). 
This procedure produces trees that reflect the merging of the persistent components as the scale parameter of the Vietoris-Rips complex increases,
and provide a measure of language relatedness seen at the syntactic level. We discuss to what extent the trees obtained in this way reflect historical
phylogenetic trees of language families. We apply our same tree construction algorithm described in Section~\ref{treealgoSec}.

\smallskip

Other methods of computational phylogenetics were used to produce phylogenetic trees of language families based on syntactic data \cite{Longo3}, \cite{LongoCeo}, \cite{Longo1}, \cite{LongoGua2}, and also \cite{OSBM}, 
as well as on a combination of syntax and other linguistic data, \cite{evolang11}, \cite{RWT}. What is new in
the analysis we present here is not the use of syntactic data but the method for tree reconstruction, which
is based on persistent connected components rather than on other computational methods of tree reconstruction.
As mentioned above the resulting trees should not be read exactly as phylogenetic trees, although we will see
that they do contain phylogenetic information. 

\smallskip

In Section~\ref{subsec_phylo_methods} we analyze in more detail the different methods for phylogenetic tree reconstruction in comparison to ours, whereas in the rest of this section we focus on the results obtained 
with our method.   

\smallskip
\subsection{Persistent components trees of Indo-European languages} 

We use separately the two different data sets, the SSWL and the LanGeLin syntactic data, 
to build our persistent components trees. Although the SSWL contains more languages, the languages are not 
uniformly mapped in the present version of the database. The LanGeLin data set on the other hand 
contains fewer languages but nearly all of the languages are fully mapped (in ternary rather than binary variables),
and with variables that more accurately correspond to the linguistic notion of ``syntactic parameter". 
Indeed, as was also observed in \cite{OSBM} in tree reconstructions obtained via the phylogenetic 
algebraic geometry method, the trees generated using the LanGeLin data appear to reflect more 
accurately in their topology and clustering what is known from historical linguistics 
about phylogenetic trees of subfamilies of the Indo-European families.   

\smallskip

It was shown in \cite{ShuMar2} that if one attempts a phylogenetic tree reconstruction using
the entire SSWL database simultaneously with the lacunae in the data simply filled by a $0.5$ 
value one obtains nonsensical answers regarding language proximity in the tree, with languages
misplaced within their family, or placed in a completely different family altogether, and with ancient 
languages especially likely to be misplaced. Although the computational method for phylogenetic 
reconstruction used in \cite{ShuMar2} to illustrate this point was simply the standard parsimony method
built into the PHYLIP phylogenetic software, 
the same problem is likely to affect the trees obtained via our topological persistent components as well.

\smallskip

Indeed, we will first compute the outcome of applying our method to the entire Indo-European
family of SSWL data, retaining all languages, without filtering them by completeness. We will
then analyze where the tree constructed in this way fails to represent faithfully the historical
linguistic information. We will then proceed to analyze smaller subfamilies by filtering the
languages by completeness and proceed to a more accurate tree reconstruction which we
will again compare to known historical linguistic information. We will show that, although this
filtering of the SSWL data improves the correlation between the persistent components trees
and the phylogenetic trees, they still show significant differences. 

\smallskip
\subsubsection{Indo-European tree from the full unfiltered set of SSWL data}

First let us view the tree generated from the SSWL data. When we use the full data, including the poorly mapped languages, we obtain a tree that
has one main big subtree (cluster N.~ $133$) and the other subtrees represent either small clusters (with maximum of $4$ points) or singletons 
that are joined later on. This tree splits into two main subtrees. One is cluster N.~$129$ which in turn splits into cluster N.~$125$ that 
contains most of the Romance languages and cluster N.~$124$ which contains most of the Germanic languages. The other is cluster N.~$132$ which 
contains a mix of many different language subfamilies such as some of the Greek languages, Altaic languages, Balto-Slavic languages, some Indo-Iranian languages, and more. 
Other than that, in a close proximity to this subtree (N.~ $130$) we have the Armenian languages and Ossetic languages (Indo-Iranian) grouped together in subtree $134$. 
This may support the hypothesis of Proto-Armenian situated between Proto-Greek and Proto-Indo-Iranian. As we know that although Armenian languages 
form a branch by themselves they share common features with Indo-Iranian and Greek languages.  Most of the other languages are added as singletons (this was discussed in Section~\ref{cluster_analysis}). 

\smallskip

However, as we discuss below in the examples of specific language subfamilies of the Indo-European languages, the use of the full unfiltered data including poorly mapped languages and incomplete variables causes a lot 
of incorrect placements of languages both within and across subfamilies, as already observed with other tree construction methods in \cite{ShuMar2}.
Thus, we restrict the SSWL data by first keeping only languages that are at least a certain percentage complete and then for those languages we keep only the syntactic variables that are fully mapped, as done in \cite{OSBM}.

\smallskip
\subsubsection{Persistent components tree from the filtered SSWL data}	\label{SSWLfilterH0Sec}

\begin{figure} [!htb]  \center{\includegraphics[width = 7.6in, angle=90]{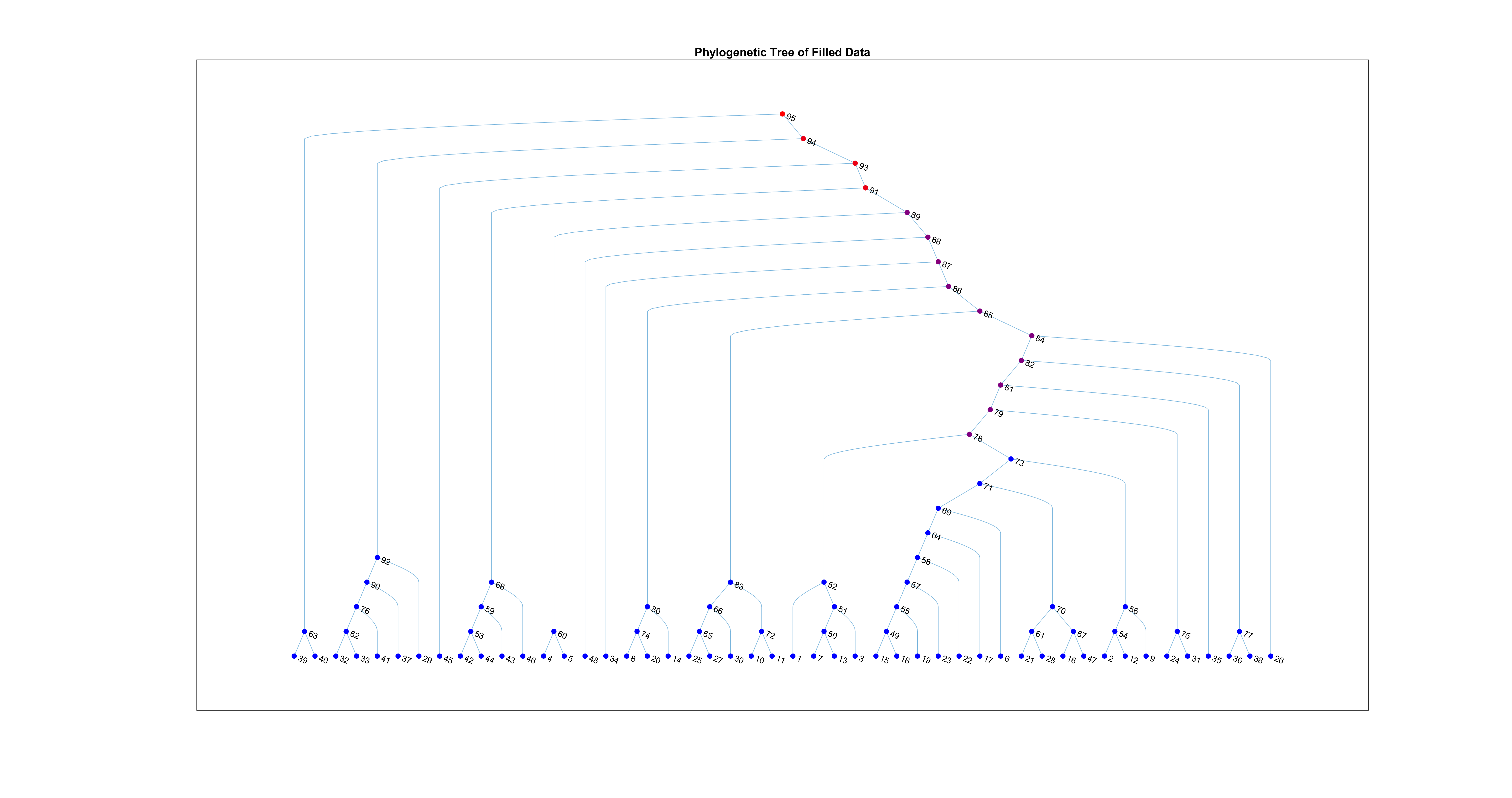}}
\caption{Persistent components tree for the filtered SSWL data. \label{SSWLtreefilterFig}}
\end{figure}

\begin{figure}[ht]
\includegraphics[width =5.87in]{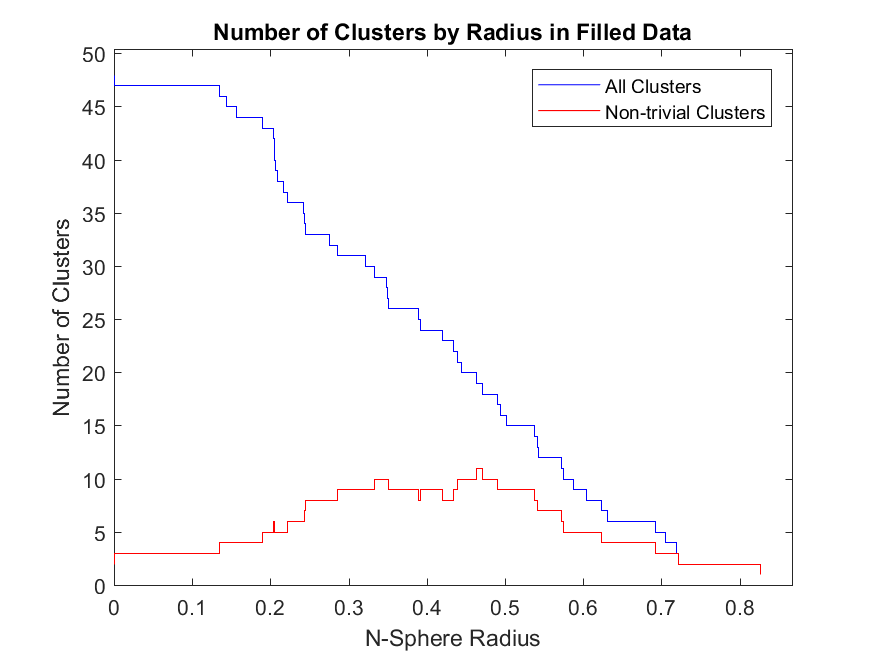}\\
\includegraphics[width =5.9in]{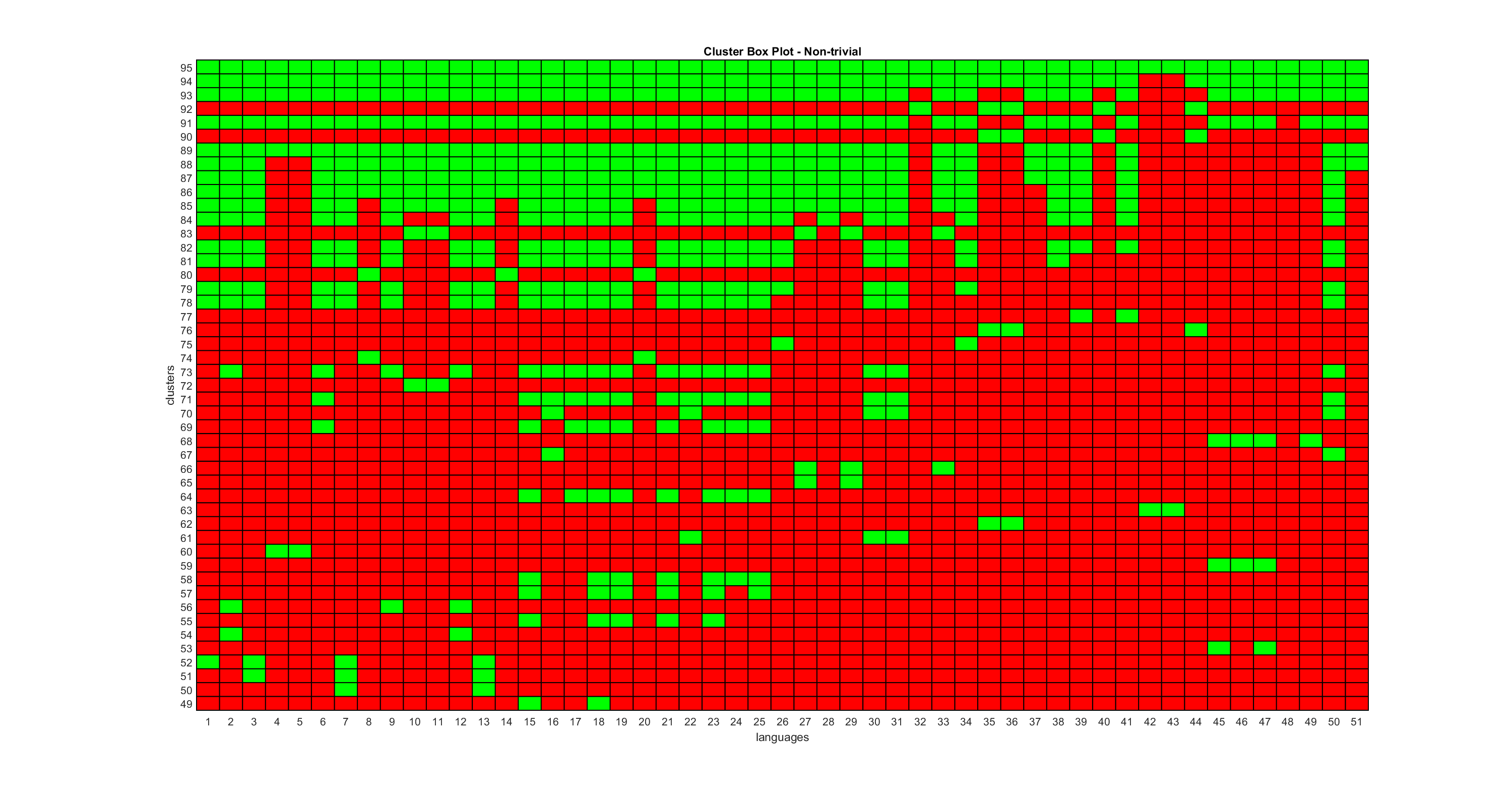}
\caption{Clusters and clustering structure in the tree of 
the filtered SSWL data. \label{IEtreefilterFig2}}
\end{figure}

We consider the subset of languages of the SSWL database that are at least $50\%$ 
complete and we restrict their coordinates to the subset of syntactic variables that are 
completely mapped for all the languages in this set. We then compute the persistent
components tree with the method discussed above. The resulting tree is illustrated in 
Figure~\ref{SSWLtreefilterFig}. We focus on the Indo-European languages together with
the Uralic and Altaic languages (we include also Korean and Japanese although no
longer regarded as plausibly Altaic). 

\smallskip

In the cluster structure of the tree of Figure~\ref{SSWLtreefilterFig} we see (from left to right)
a cluster N.~63 with Korean and Japanese which is a reasonable clustering. 
Cluster N.~92  contains the Armenian languages (Eastern Armenian and Western Armenian joined
together as sub-cluster N.~62), successively joined by Turkish, Hungarian, and Cappadocian Greek.
This clearly does not correspond to a historical phylogenetic clustering since it mixes some Indo-European
and some Uralic and Altaic languages. Cluster N.45 is a singleton with Sorani Kurdish, which only joins other
components very high up in the tree (cluster N.~93). Cluster N.~68 is again a cluster that has good
correlation with historical phylogenetic trees: it contains the Indo-Iranic Indo-European languages
Hindi, Panjabi, Pashto, and Nepali. Cluster N.~60 only contains English and Singaporean English and is
also added very late to the other components (cluster N.~89): in particular, even after filtering in order
to correct for the incompleteness of the data, when all language families are considered together,
English is not correctly grouped near the other Germanic languages. Clusters N.~48 and N.~34,
two singletons, respectively consisting of Haitian and Czech, which also join the tree at higher levels: again
these languages are misplaced out of grouping within the appropriate subfamily. Cluster N.~80 contains
Gothic, Icelandic, and Late Latin. This is an indication of the tendency of the SSWL data, when analyzed
without prior subdivision into subfamilies, to misplace the location of the ancient languages. This fact
was also already observed by different methods in \cite{ShuMar2}. Cluster n.~83 again groups together
mostly ancient languages:  Latin, Ancient Greek, and Homeric Greek (sub-cluster N.~66) and Old English
and Old Saxon (sub-cluster N.~72). It is not surprising that, without additional indication about
subfamilies, the syntactic similarities between the ancient Indo-European languages is detected as
a closer clustering than the respective similarities with their modern descendants. Cluster N.~52 contains
some Germanic languages: Afrikaans, Dutch, German, and West Flemish, all of which belong to the
West-Germanic split of the Germanic languages. So again we see here a subtree that correlates well with
historical phylogenetic trees (except for the misplacement of English outside of this tree as 
we mentioned above). This cluster N.~52 is part of a larger cluster (N.~78) which contains
another main subtree (cluster N.~73) which in turn splits into three main sub-clusters, N.~69, N.~70,
N.~56. Of these, cluster N.~56 contains other Germanic languages all belonging to the North-Germanic
split: Danish, Norwegian, and Swedish. Two other North-Germanic languages in the database, Icelandic and
Faroese are misplaced: Icelandic, as we have seen, is grouped with Gothic and Late Latin, and Faroese
is in the same larger cluster N.~78 that contains the Germanic languages, but placed in closer proximity to
the sub-cluster (N.~69) that contains the Romance languages Portuguese, Brazilian Portuguese, 
Catalan, Italian, Napoletano Antico, Sicilian, Spanish, and French. The remaining sub-cluster in the larger
cluster N.~78 is cluster N.~70 that contains Northern Calabrian, Romanian, 
Cypriot Greek, Greek, and Albanian. This sub-cluster again does not correspond to a
historical phylogenetic tree since it contains a mixture of Hellenic and Romance languages. 
Cluster N.~75 has Galician and Medieval Greek, which again do not have close historical relatedness.
Cluster N.~35 has Polish that appears as a singleton, while Russian appears in the nearby two-point
cluster N.~77 joined with the Uralic language Finnish. After filtering the data for completeness no other
Slavic language is left. The last singleton cluster, N.~26 contains Old French.

\smallskip

The clustering structure of the tree of the persistent connected components 
clearly does not reflect the branching into main families (there is a mixing of
Indo-European and Uralic and Altaic languages for instance) nor the
splitting into subfamilies of the Indo-European language family, even though it 
retains some information about the subdivision into subfamilies. As we discuss 
more in detail below when we focus on  smaller subfamilies of the Indo-European
family, this discrepancy between persistent components trees and historical
phylogenetic trees is not caused by a problem with the SSWL data. Indeed we
will see that the discrepancy persists when we focus on smaller groups of
languages for which it was already established in \cite{OSBM} that the SSWL
data do reconstruct the correct historical phylogenetic tree, when the phylogenetic
algebraic geometry method is used for the inference of the tree structure.
Thus, the discrepancy with the persistent components tree should be ascribed
more intrinsically to a different type of information about syntactic proximity
carried by the clustering of the persistent $H_0$. We will show later that some
amount of misplacement at smaller scales (within subfamilies but nor across
larger family subdivisions) can also be caused by the variance level in the
PCA analysis: this may also account for some of the misplacements within 
the smaller subfamilies.

\smallskip
\subsubsection{The Indo-European persistent components tree of the LanGeLin data} \label{LanGeLinH0Sec}

\begin{figure} [ht]  \includegraphics[width = 7.62in, angle=90]{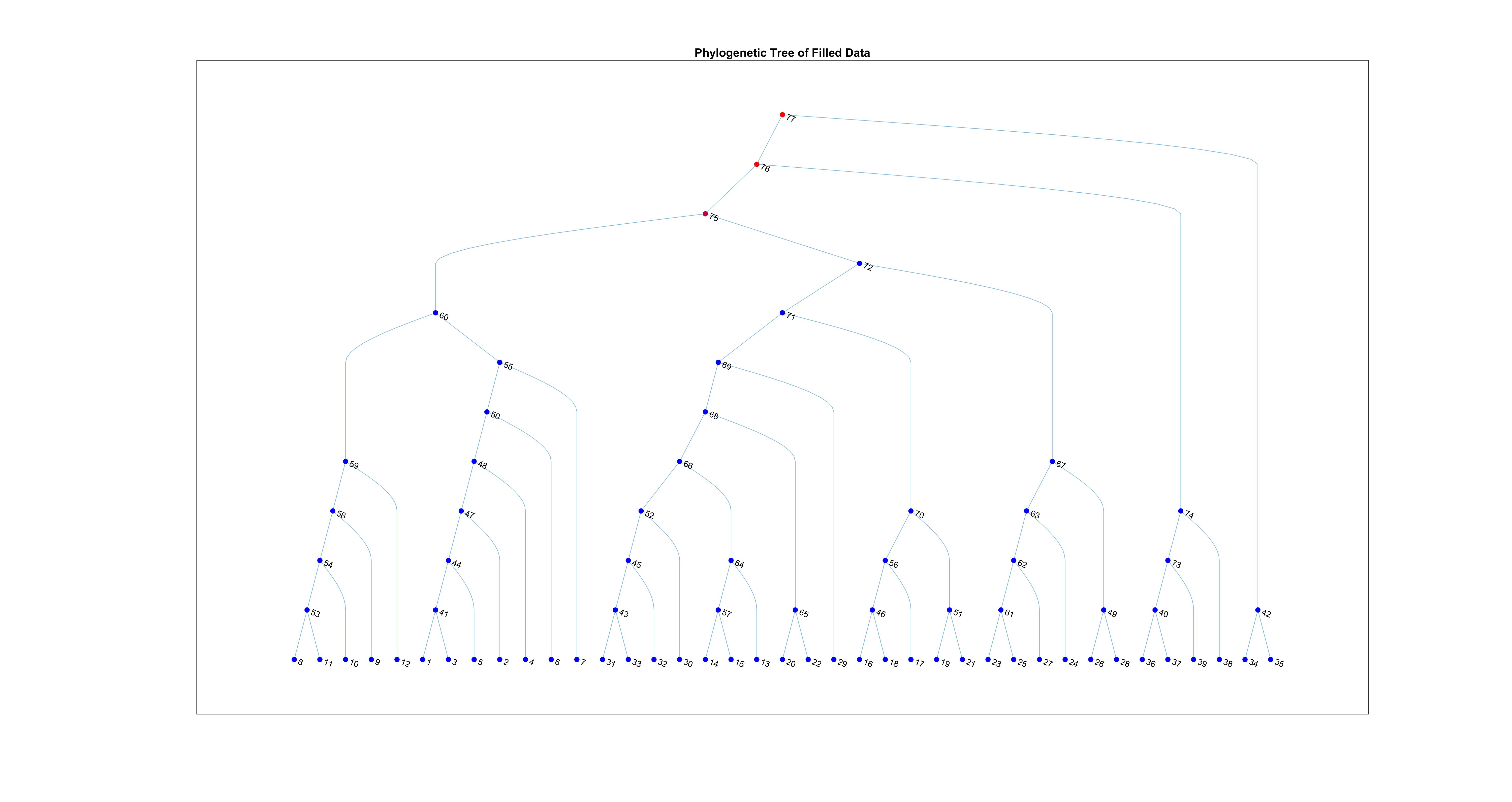}
\caption{Persistent components tree from LanGeLin data, PCA $60\%$.\label{Longobardi_fulltree}}
\end{figure}

\begin{figure} [ht]  \includegraphics[width = 5.59in]{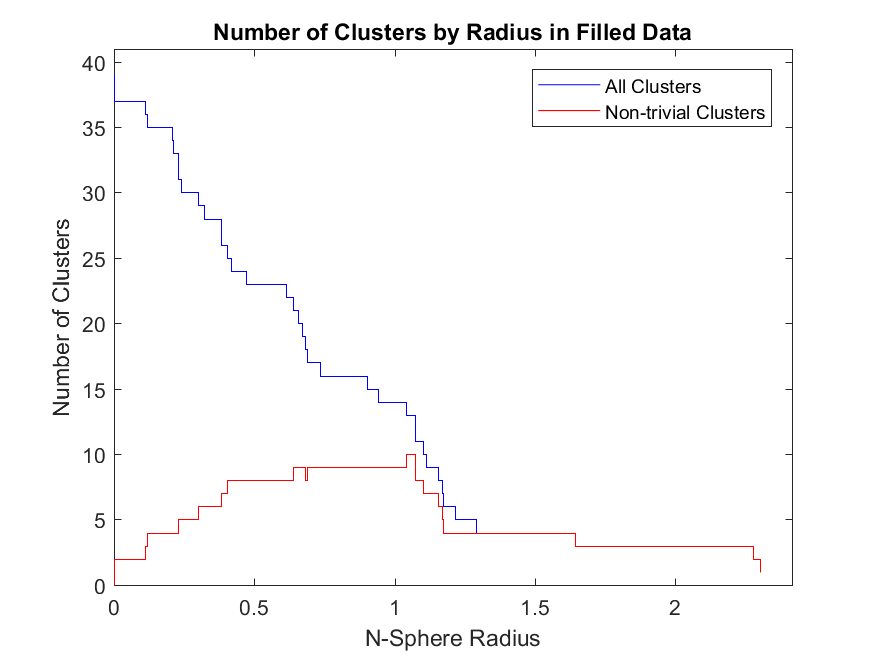}\\
\includegraphics[width = 6.2in]{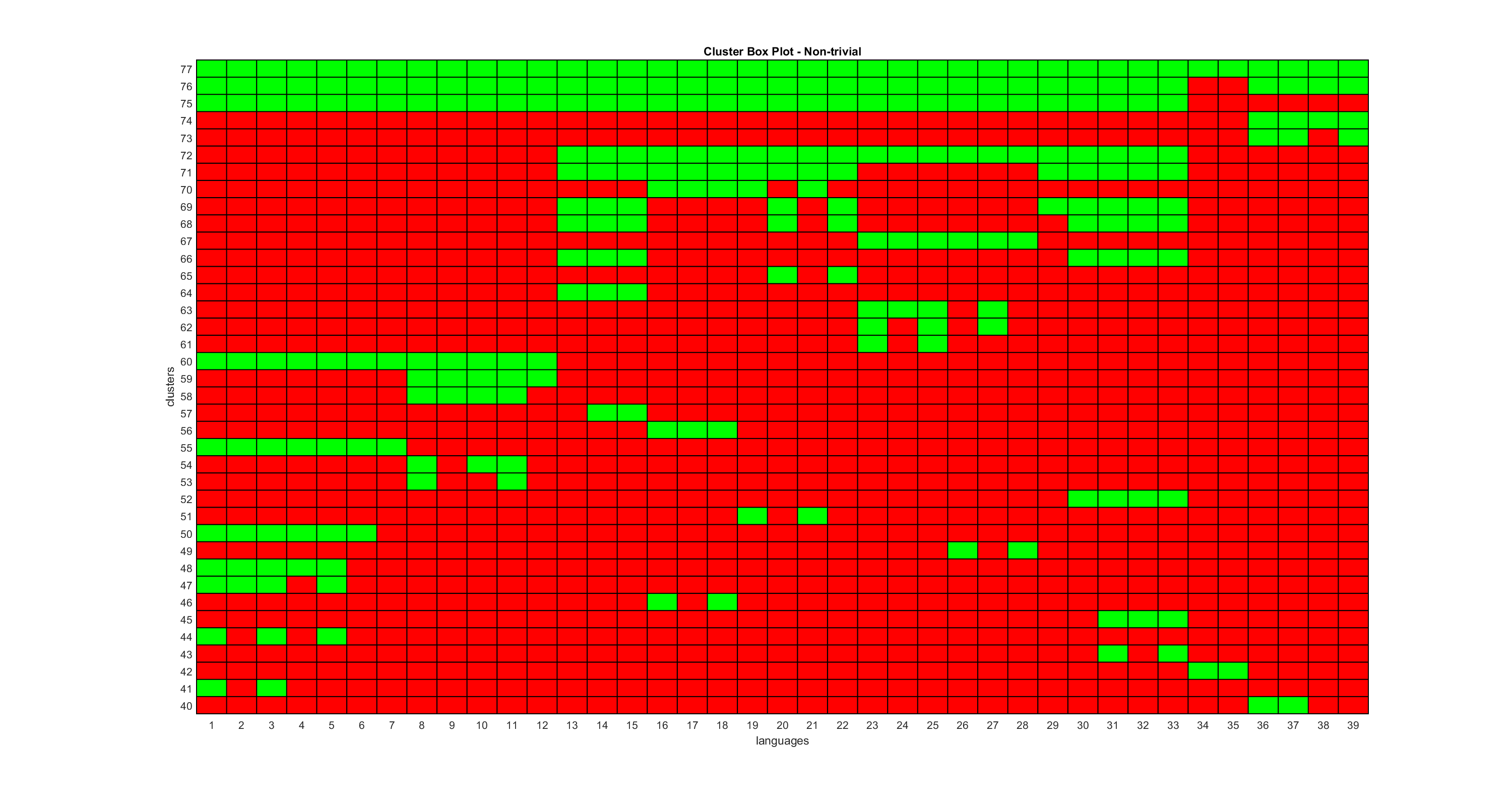}
\caption{Clusters and clustering structure in the persistent components tree for the LanGeLin data, PCA $60\%$.\label{Longobardi_fulltree2}}
\end{figure}

\medskip

Our topological reconstruction of a persistent components tree for the LanGeLin data (see Figure~\ref{Longobardi_fulltree})
exhibits a stronger correlation between the clustering of the persistent connected components
and the tree topology of the historical phylogenetic trees 
than the case of the tree of the SSWL data analyzed above.
As discussed before (Section \ref{cluster_analysis}) the LanGeLin persistent components tree almost does not have any singletons (especially at PCA level of $60 \%$). 

\smallskip

The structure of the subtrees (from left to right) gives the following clustering of languages.
We have a large cluster (N.~60) that contains two main subclusters, N.~59 and N.~55. 
Cluster N.~59 contains the main modern Romance languages: Italian,
Spanish, French, Portuguese, and Romanian, while the nearby cluster N.~55 contains the
Romance Southern Italian dialects: Ragusa, Mussomeli,
Aidone, Southern Calabrese, Salentino, Northern Calabrese, and 
Campano. There is then another very large cluster, N.~72, which contains several
sub-cluster structures which include Hellenic, Germanic, and Slavic languages.
In particular, cluster N.~52 contains all the Slavic languages except Bulgarian:
Serb-Croatian, Slovenian, Polish, and Russian. Cluster N.~64 groups together the
ancient languages Latin, Classical Greek, and New Testament Greek. Cluster N.~65
has Romeyka Pontic Greek grouped together with Gothic and adjacent to the
previous cluster with ancient Greek. We will return to discuss the association of
Gothic and the Hellenic languages in relation to $H_1$-structures later. Cluster N.~29
contains Bulgarian as a singleton that is added to the previous clusters (N.~52 Slavic
languages, N.~64 and N.~65 ancient languages). As we discuss more in detail below,
this occurrence of a mixture of Hellenic and Slavic languages together with Gothic
seems to be related both to the $H_1$-structure that we observed in \cite{Port} in the
SSWL data and to an $H_1$-structure in the LanGeLin data that we will discuss more
in detail in Section~\ref{H1sec}. This large cluster N.~69 is joined with cluster N.~70 to
form the larger cluster N.~71. Cluster N.~70 contains the Hellenic languages including the 
Greek Southern Italian dialects of the Greek-Italian microvariations of \cite{Gua16}:
Salento Greek, Calabrian Greek A, Calabrian Greek B, grouped together with
Modern Greek and Cypriot Greek. The large cluster N.~71 is then joined with 
cluster N.~67 that contains the Germanic languages Old English,
English, Dutch, Danish, Icelandic, and Norwegian. As in the SSWL case, Icelandic
is incorrectly placed: here it occurs together with the West-Germanic languages
instead of the North-Germanic. The whole cluster structure N.~72 (Hellenic, Germanic, and
Slavic languages) is joined to the previous cluster N.~60 of the Romance languages
into cluster N.~75. This is then joined by the smaller cluster N.~74, which contains
the Indo-Iranian languages Marathi, Hindi, Farsi, and Pashto, followed by a joining 
with the small cluster N.~42 with Irish and Welsh. 

\smallskip

As opposed to the SSWL case,  with the LanGeLin data all of the Germanic languages are grouped together within the same subtree, with Norwegian and Danish as a subtree (these correctly belong to the North-Germanic branch) and the rest (Old English, English, German, Icelandic and Goth) 
in a subtree together with the Balto-Slavic languages.
Regarding the position of the old languages within the tree, we find Gothic located closer 
to the Balto-Slavic subtree than to the West-Germanic subtree and Icelandic erroneously located 
together with the West-Germanic languages.  Despite the structure involving Gothic that
we will discuss later, the old languages Gothic and Old English are correctly located in the 
same clustering structure involving the sub-cluster of the West Germanic languages.  

\smallskip

However, the nearby grouping of the Germanic and the Balto-Slavic languages in the persistent component tree of the LanGeLin data
differs from the usual phylogenetic tree of the Indo-European languages (including the one constructed from the LanGeLin data
in \cite{Longo1}) which have an Indo-Iranian and Balto-Slavic split followed at a lower level in the tree by a Germanic split. The
main branches in the tree of \cite{Longo1} derived from an earlier version of the LanGeLin data take the form 
\begin{center}
\Tree [ Celtic [ [ Indo-Iranian Balto-Slavic ] [ Germanic [ Hellenic Romance ]]] ]   
\end{center}
The persistent components tree for the full LanGeLin data presents a different branching structure between the Indo-European
subfamilies, of the form
\begin{center}
\Tree [ Celtic [  Indo-Iranian [ [ Hellenic [ Germanic Balto-Slavic ] ] Romance ] ] ] 
\end{center}
This subdivision differs also from other proposed phylogenetic trees of the main Indo-European language families. 
It is more likely that this signals the different meaning between the clustering information contained in the persistent
components and the historical relatedness through phylogenetic trees rather than indicating a different possible
phylogenetic proposal for the Indo-European family tree. 

\smallskip

When computing the larger tree (also at PCA $60\%$) for all the languages in the
LanGeLin data, we also find a larger cluster that contains the Indo-Iranian languages 
as a sub-cluster, but now we find
the non-Indo-European Uralic language Yukaghir grouped within the same
sub-cluster with the Indo-Iranian languages. In proximity of this sub-cluster we
find another sub-cluster with other Uralic languages, Hungarian, Finnish, Estonian, and  Khanty.
These Uralic languages are in turn placed next to the putative Altaic languages Yakut, Turkish, 
Buryat, Even, and Evenki. The remaining Uralic language Udmurt (Ud) appear together with the 
Altaic languages, see the more detailed discussion in Section~\ref{UrAltSec}. 
These sub-clusters are joined into a large cluster that also contains 
a singleton given by the Inuit language Inuktitut.
Japanese (Jap) and Korean (Kor) are placed in proximity of each other but not in proximity of 
the Altaic languages, in agreement with the current understanding that they should not be 
considered part of a hypothetical Altaic or Ural-Altaic family. 

\smallskip
\subsection{Germanic language family}\label{GermSec}

First let us consider the Germanic family persistent components tree. 
It is known from historical linguistics that the Germanic language family consists of 
two main branches: the North Germanic branch and the west Germanic branch. 
We discuss the trees obtained by computation of the persistent connected components
for the SSWL data and for the LanGeLin data and we compare them with the available 
information from historical linguistics. 
This is a good example where we see that the persistent components tree can
differ from the phylogenetic tree in significant ways. It is also an example that
illustrates that there can be significant effects caused by changing the level of
variance of the PCA as we discuss below.

\begin{figure} [h]  \includegraphics[width = 7.5in, angle=90]{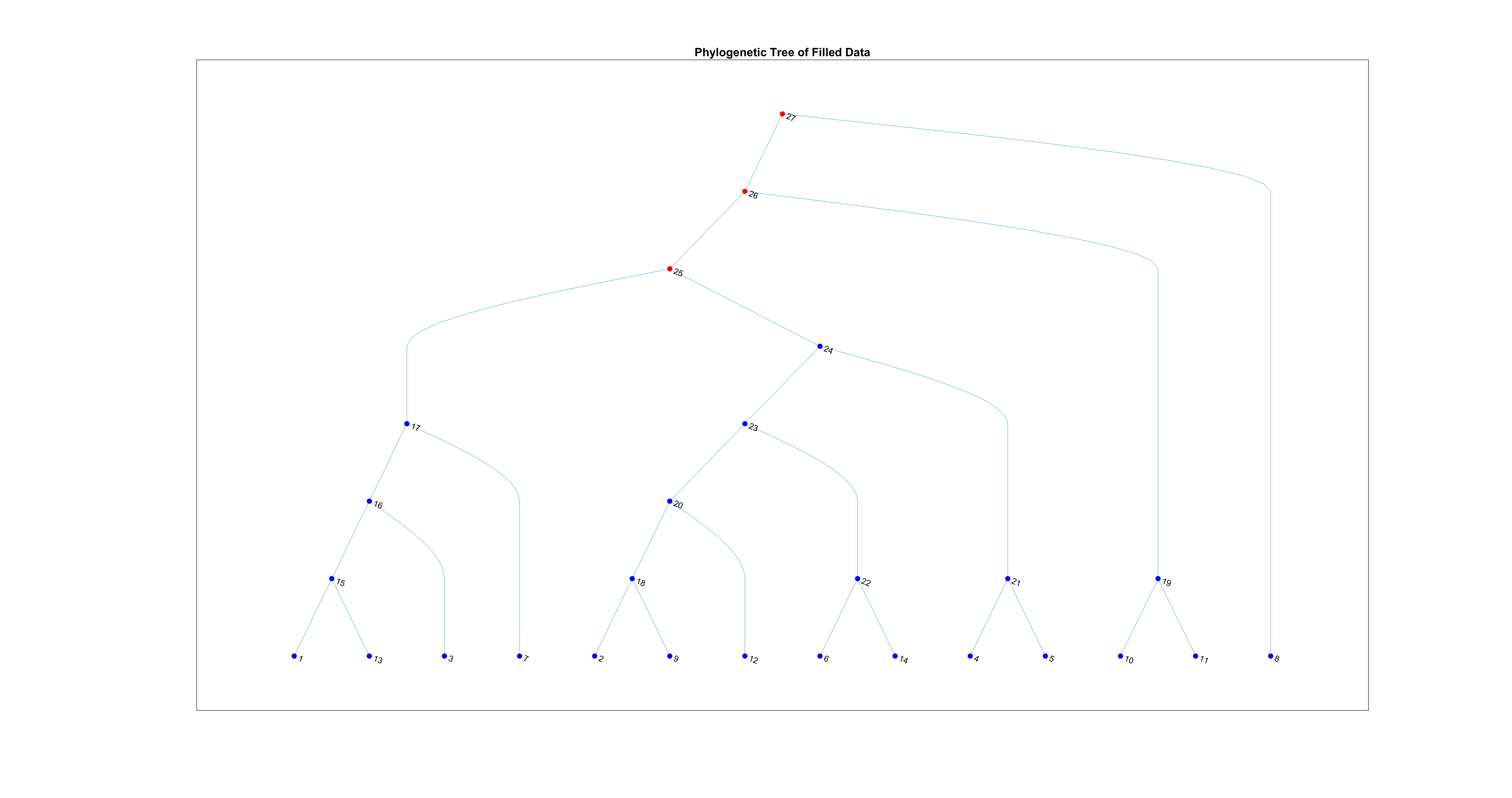}
\caption{Persistent components tree of Germanic languages from filtered SSWL data, PCA $60\%$.\label{SSWLGermanicFiltertree}}
\end{figure}

\begin{figure} [h]  \includegraphics[width = 5.59in]{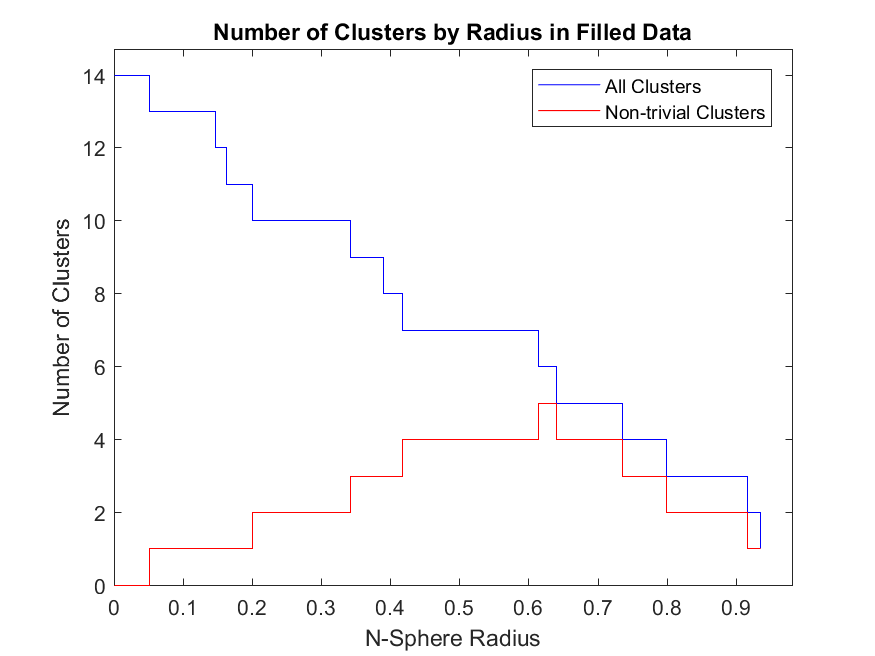}\\
\includegraphics[width = 6.25in]{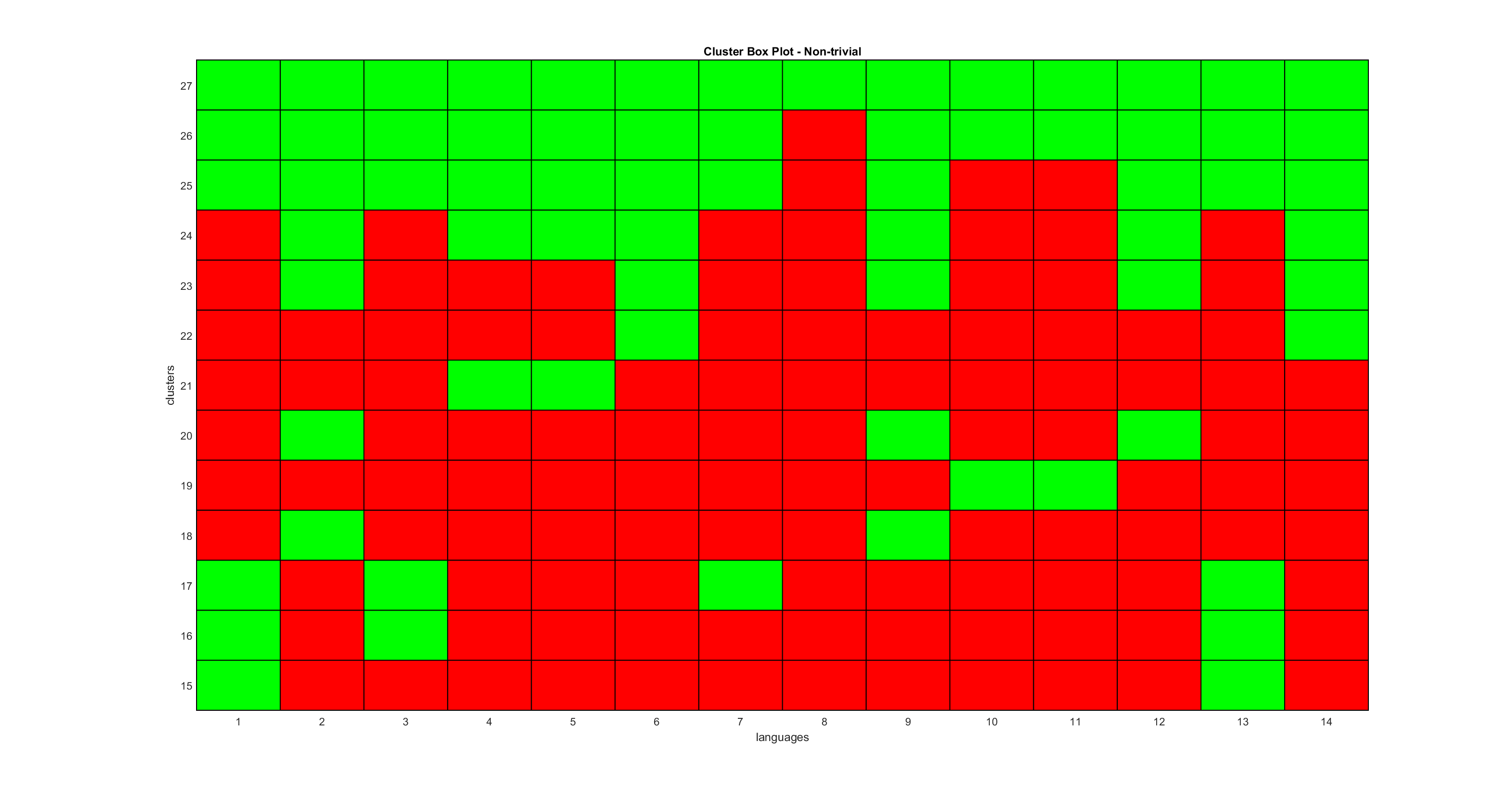}
\caption{Clusters and clustering structure in the persistent components tree for the Germanic languages with filtered SSWL data, PCA $60\%$.\label{SSWLGermanicFiltertree2}}
\end{figure}

\smallskip
\subsubsection{Germanic language family in the SSWL data}\label{SSWLGermSec}
The Germanic languages recorded in the SSWL database include:
Afrikaans, Cimbrian, Danish, Dutch, English, Middle English, Old English,
Faroese, Frisian, German, Gothic, Icelandic, Middle Dutch, 
M\`ocheno, Norwegian, Oevdalian, Old Norse, Old Saxon, Swedish, 
Swiss German, West Flemish, and Yiddish.

\smallskip

Looking at the full tree for the unfiltered SSWL data we see that that most of the Germanic language are grouped together in tree N.~$124$, but there are notable misplacements to other family trees, for example the case of English which is incorrectly located within the Romance languages subtree (tree N.~$125$). 
The tree N.~$124$ splits into two subtrees; The first is tree N.~$119$ containing Afrikaans, Faroese, Frisian, Icelandic, Norwegian, Swiss German and West Flemish and the second one, tree N.~$114$, containing Dutch, German, and Swedish.  Swedish is misplaced in the same subtree with Dutch and German. While the latter two belong to the same West Germanic branch, Swedish is a North Germanic language.
A further look into the second subtree containing the rest of the Germanic languages reveals that it splits into two branches, one consisting of West Frisian and Swiss German and the other containing a sub-branch of Faroese, Icelandic, Norwegian, and another sub-branch of Afrikaans and West Flemish. 
The positions of some of these languages do indeed correspond to the known phylogenetic tree, while other positions such as that of Swedish
are clearly wrong, and some languages are entirely missing from the subtree, like English which is misplaced in a completely different family tree.

\smallskip

This further confirms that, although most of the German subfamily is grouped together, the position of some languages 
within this Germanic tree is incorrect with respect to its known position. This reflects indeed the same type
of problems discussed in \cite{ShuMar2} regarding the reconstruction of phylogenetic trees using the entire
unfiltered SSWL data.

\smallskip

To circumvent the problem of the incompleteness of the SSWL data, we  proceed as before and we 
filter the set of Germanic languages by retaining only those that are at least 
$50\%$ complete  and then for this set we retain only those parameters that 
are completely mapped for all the languages in the set. This reduces the 
dimensionality of the ambient space. We then further select smaller sets of
Germanic languages and use only syntactic variables that are fully mapped for
all of these languages. The remaining languages in the Germanic family after
this filtering are Afrikaans, Danish, Dutch,
English and Singaporean English, Faroese,
German, Gothic, Norwegian
Old English, Old Saxon, Swedish, 
West Flemish, and Icelandic. 

\smallskip

After retaining only the languages and fully mapped syntactic variables selected in this way, 
the persistent components tree that we obtain from the SSWL data for the Germanic 
languages (at PCA $60\%$) is of the form illustrated in Figure~\ref{SSWLGermanicFiltertree}.

\smallskip

The clustering structure (from left to right in Figure~\ref{SSWLGermanicFiltertree}, see also Figure~\ref{SSWLGermanicFiltertree2}) has a cluster N.~17 containing Afrikaans, 
Dutch, German, and West Flemish which is joined to a cluster N.~24 which is in turn
a merge of two subclusters: N.~23 which contains the North-Germanic languages
Danish, Faroese, Norwegian, Swedish, Icelandic and cluster N.~21 which contains
English and Singaporean English.  Except for the fact that the English sub-cluster
joins the North-Germanic sub-cluster before nerging with the West-Germanic sub-cluster,
this larger cluster does show the expected West-Germanic/North-Germanic split.
The ancient languages are correctly joined higher up in the tree,  with one
sub-cluster N.19 with Old English and Old Saxon and a singleton cluster with Gothic.
In this case, the filtering and restriction to the Germanic subfamily improve the 
correlation of the persistent components tree to the historical phylogenetic tree.

\smallskip
\subsubsection{The Germanic family in the LanGeLin data}\label{LanGeLinGermSec}

\begin{figure} [h]  \includegraphics[width = 7.5in, angle=90]{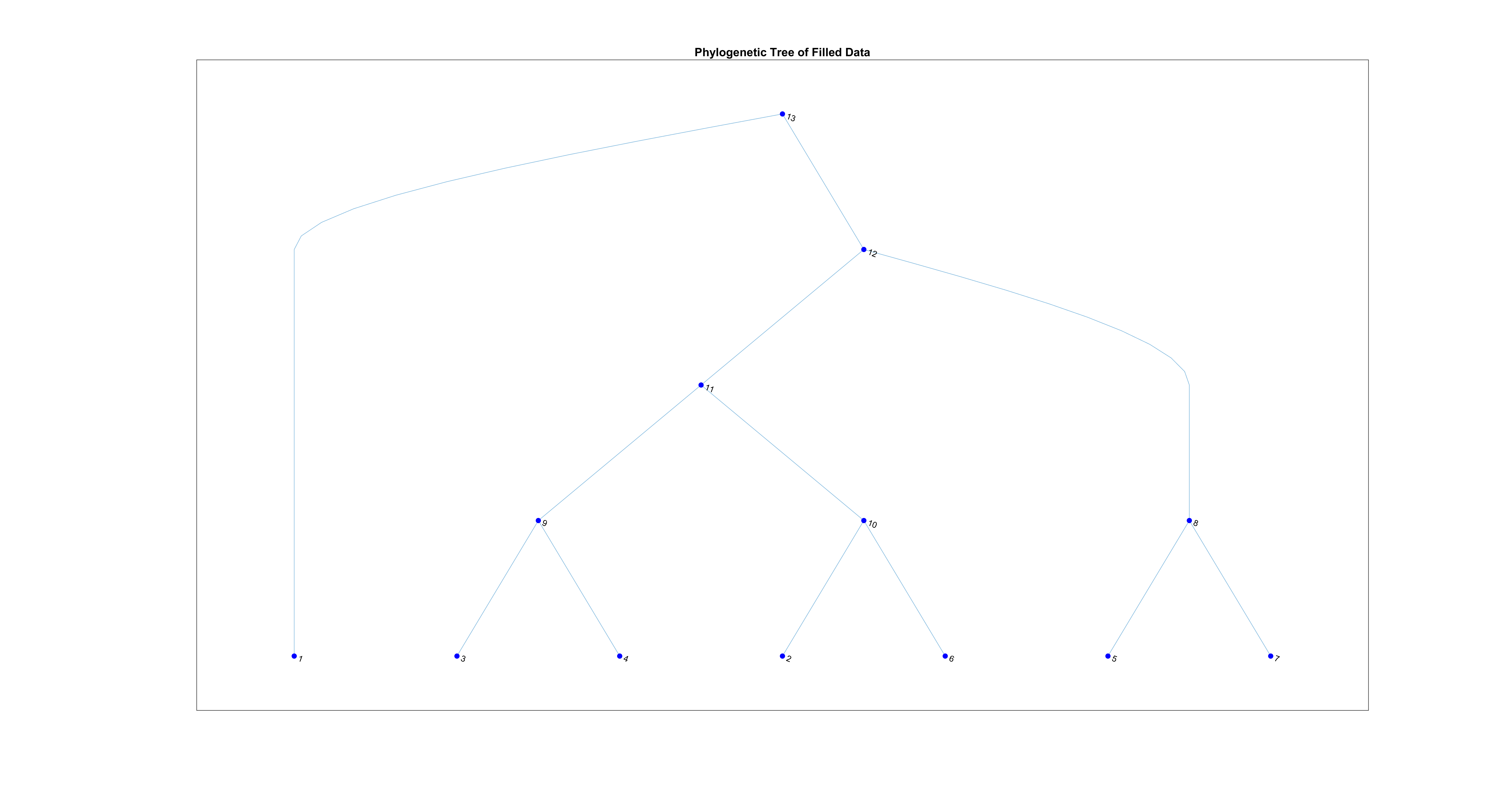}
\caption{Persistent components tree of Germanic languages from LanGeLin data, PCA $60\%$.\label{LanGeLinGermanicFiltertree}}
\end{figure}

\begin{figure} [h]  \includegraphics[width = 5.59in]{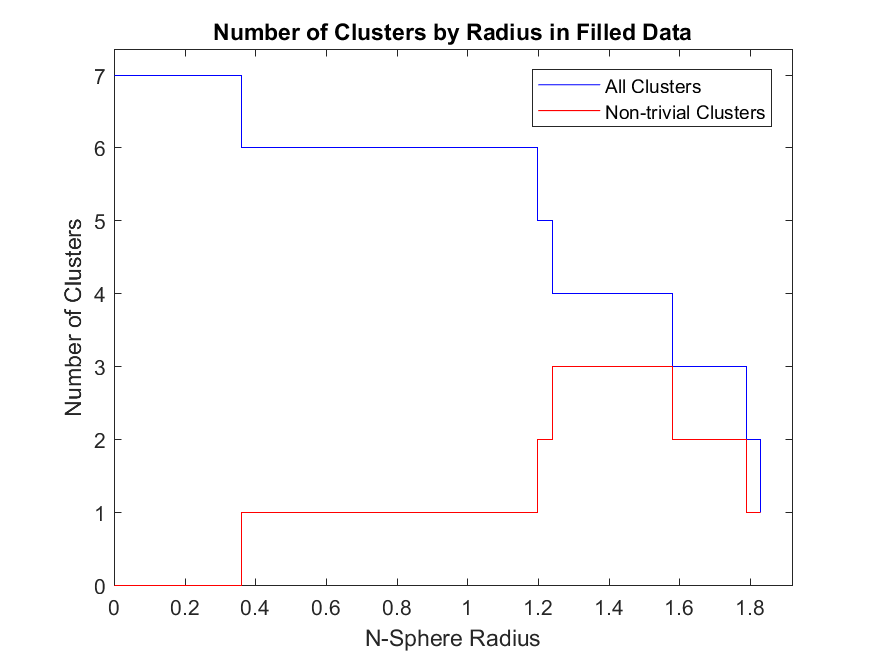}\\
\includegraphics[width = 6.25in]{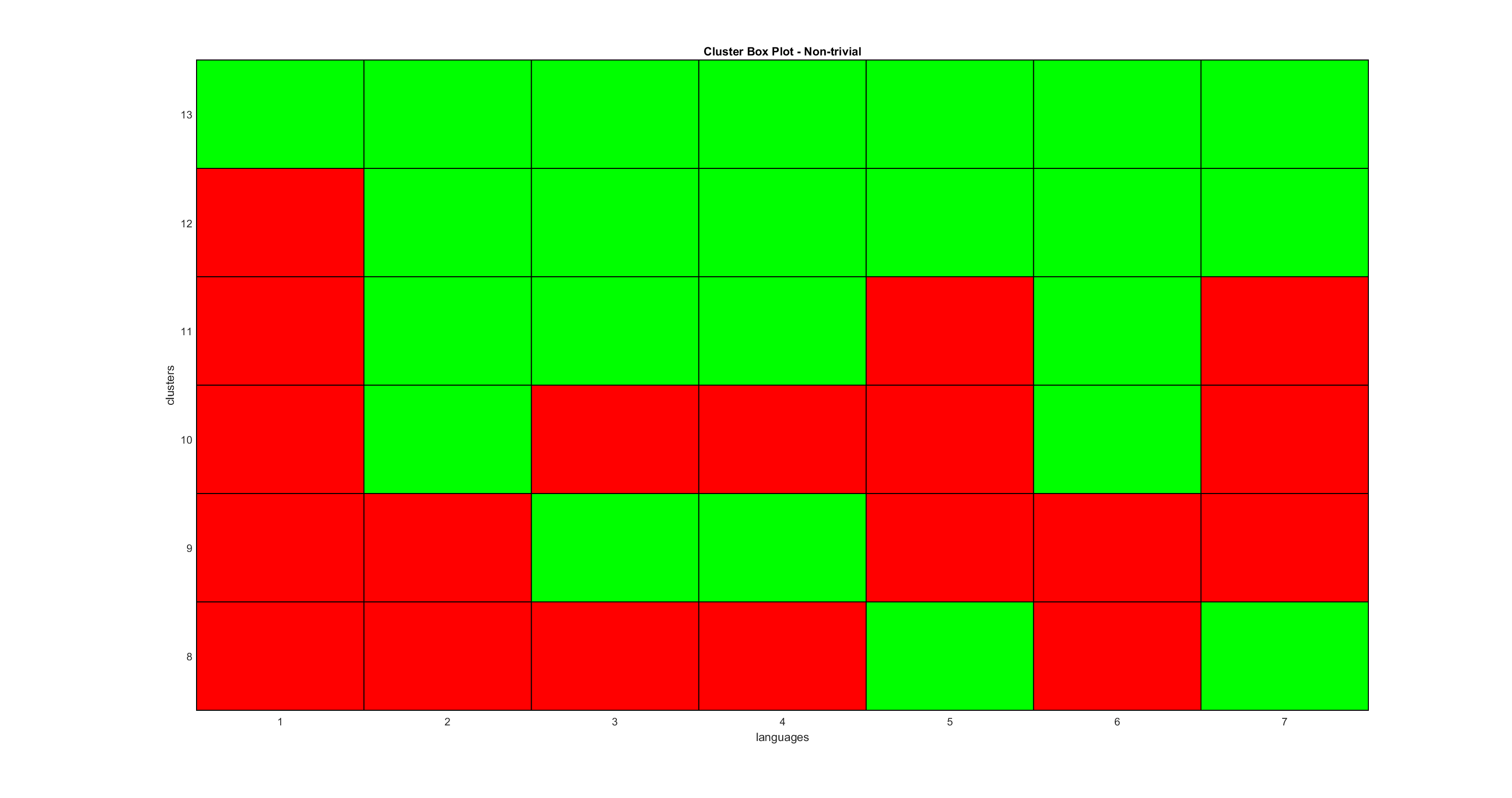}
\caption{Clusters and clustering structure in the persistent components tree for the Germanic languages with LanGeLin data, PCA $60\%$.\label{LanGeLinGermanicFiltertree2}}
\end{figure}

The position of the Germanic languages within the tree of the connected persistent components
of the LanGeLin data also presents a structure that does not entirely correspond to the expected 
phylogenetic tree of the Germanic languages. In this case, however, the problem does not
originate from the incompleteness of the data, unlike the SSWL case: it reflects an intrinsic difference between
the branching structure recorded by the persistent components and the historical phylogenetic trees. 
The tree obtained at PCA $60\%$ is given in Figure~\ref{LanGeLinGermanicFiltertree} with the
clustering structure as in Figure~\ref{LanGeLinGermanicFiltertree2}.

\smallskip

The clustering structure (from left to right in Figure~\ref{LanGeLinGermanicFiltertree}) shows
Gothic as a singleton joined at the root top of the whole tree, followed by a main split into
two clusters of pairs, N.~9 with English and Dutch, and N.~10 with Old English and Icelandic,
these two clusters merge together in cluster N.~11, which is then joined by another cluster
(N.~8) consisting of Danish and Norwegian. The clustering of these two languages correctly
reflects the North-Germanic grouping. The position of Old English, English and Dutch together 
also correctly reflects a West-Germanic subtree, but the grouping of 
Icelandic together with Old English is incorrect, since Icelandic is a 
North-Germanic language. 

\smallskip
\subsubsection{Comparison with algebro-geometric reconstructions}

The two sets (Dutch, German, English, Faroese, Icelandic, Swedish)
and (Norwegian, Danish, Icelandic, German, English, Gothic, Old English) analyzed
in \cite{OSBM} have, respectively, 90 and 68 SSWL variables that are completely mapped
for all the languages in the set. The second set of languages is represented in both
databases and was analyzed in \cite{OSBM} using separately the SSWL and the LanGeLin data.

\smallskip

When we restrict to one of these small sets of Germanic languages analyzed
in \cite{OSBM}, the set (Norwegian, Danish, Icelandic, German, English, Gothic, Old English)
which includes two ancient languages, we find that at PCA variance level $60\%$ we
obtain the tree
\begin{center}
\Tree [  [ Gothic [ Norwegian  Danish ]] [ [ Icelandic Old-English ] [ English German ]]]
\end{center}
which does not correspond to the historical tree. At PCA variance level $80\%$ we obtain
a different tree
\begin{center}
\Tree [ [ Gothic [ Icelandic  Old-English ] ] [ [ English  German ] [ Norwegian  Danish ] ] ] 
\end{center}
which is still not corresponding to the historical tree, but which appears as one of the
candidate phylogenetic trees produced by PHYLIP, the candidate $T_5(G)$, in \cite{OSBM}. 
The correct phylogenetic tree, which is also identified correctly by the phylogenetic algebraic
geometry method of \cite{OSBM}, is given by
\begin{center}
\Tree [  [ Gothic [ Icelandic [ Norwegian  Danish ]] ] [ Old-English [ English  German ] ] ]
\end{center}

\smallskip

Thus, we see that, unlike in the setting of phylogenetic reconstructions discussed in \cite{ShuMar2}
and in \cite{OSBM}, the persistent component trees continue to differ from the phylogenetic
trees even when only fully mapped syntactic variables are considered, with Icelandic located incorrectly 
in close proximity to Old-English and the grouping of languages not reflecting the North/West
Germanic split. 

\smallskip

Moreover,  this example shows that the effect of the PCA variance level on the clustering of the
persistent connected components has a detectable effect in the altering the tree topology. 
It also shows a clear case where the clustering structure of the persistent connected
components does not necessarily reflect the phylogenetic tree considered to be the correct
one by historical linguistics.

\smallskip

The algebro-geometric method of \cite{OSBM} yields the correct phylogenetic tree for the Germanic languages, 
with both the SSWL data (when only completely filled variables are retained) and for the LanGeLin data, with 
the correct subdivision into the North-Germanic and the West-Germanic branches and the position of each language 
within the branches is correct. In our case, when considering the other set of Germanic languages (Dutch, German, English, Icelandic, 
Faroese,  Swedish) discussed in \cite{OSBM} we correctly get that the closest languages to each other are Faroese to Icelandic 
and Dutch to German, but not only that we get Swedish  (incorrectly) a West Germanic language, but it also occurs as the 
farthest language (out of the ones in this list) from Faroese and Icelandic, which is also incorrect. 
		
\smallskip
\subsection{Romance language family} \label{RomanceSec}	
	
	The second subfamily of the Indo-European family that we analyze more closely is the Romance languages. 
	This subfamily contains all the modern Indo-European languages that have historically developed from Latin. 
	We find that the LanGeLin data, that includes a set of Southern Italian dialects that had historically interesting
	interactions with the Greek language family, has a persistent components tree that separates out this cluster
	of dialects from the main Romance languages (French, Italian, Spanish, Portuguese, Romanian). The persistent
	components tree of the main Romance languages does not entirely match the phylogenetic tree. When
	restricted to the part of the tree that contains only these main Romance languages and using data
	filtered by completeness, the tree obtained from the SSWL data agrees with the one obtained from the
	LanGeLin data and differs from the historical phylogenetic tree by having the positions
	of Spanish and Italian flipped.

\begin{figure} [h]  \includegraphics[width = 7.5in, angle=90]{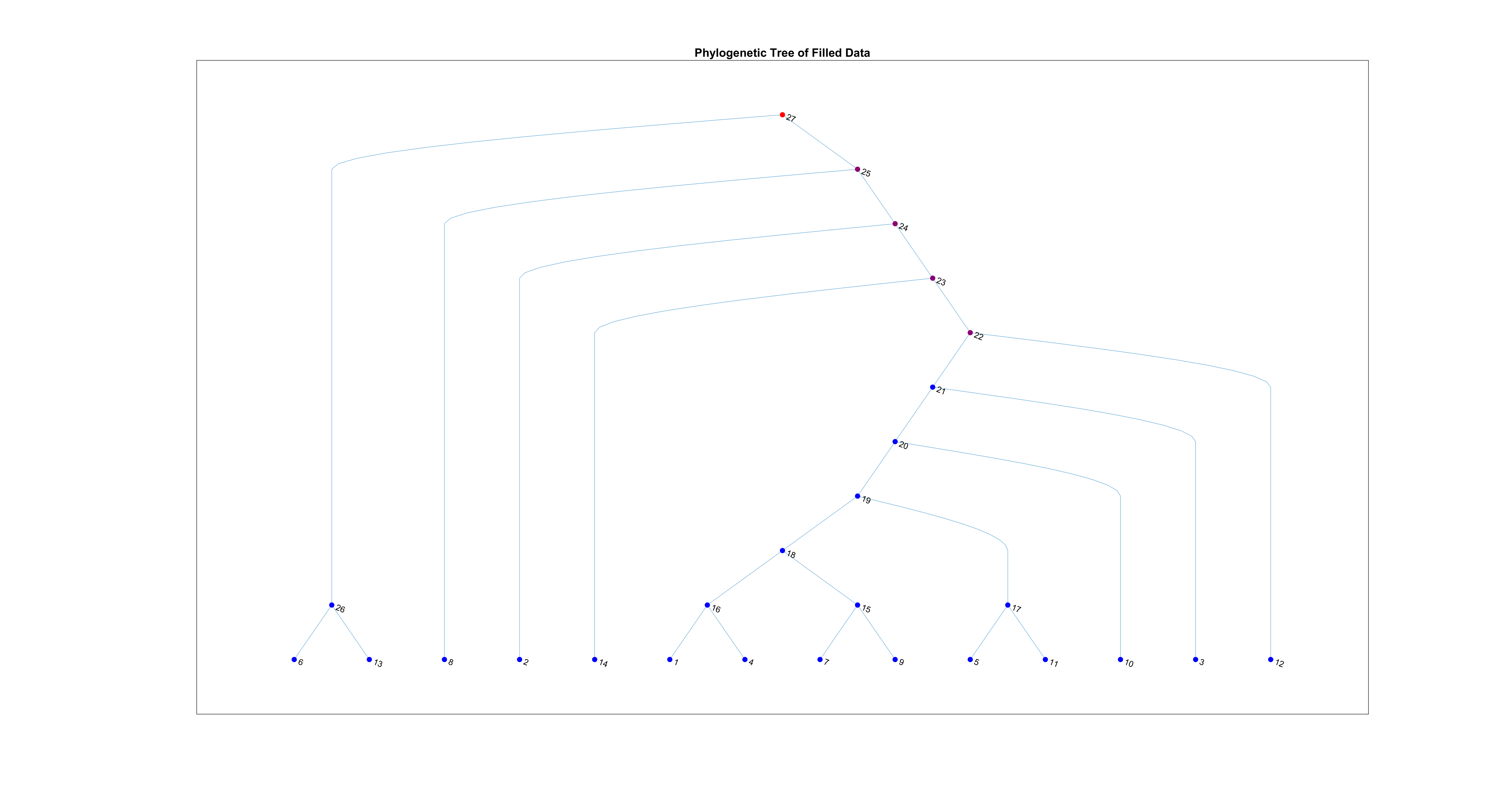}
\caption{Persistent components tree of Romance languages from filtered SSWL data, PCA $60\%$.\label{SSWLRomFiltertree}}
\end{figure}

\begin{figure} [h]  \includegraphics[width = 5.56in]{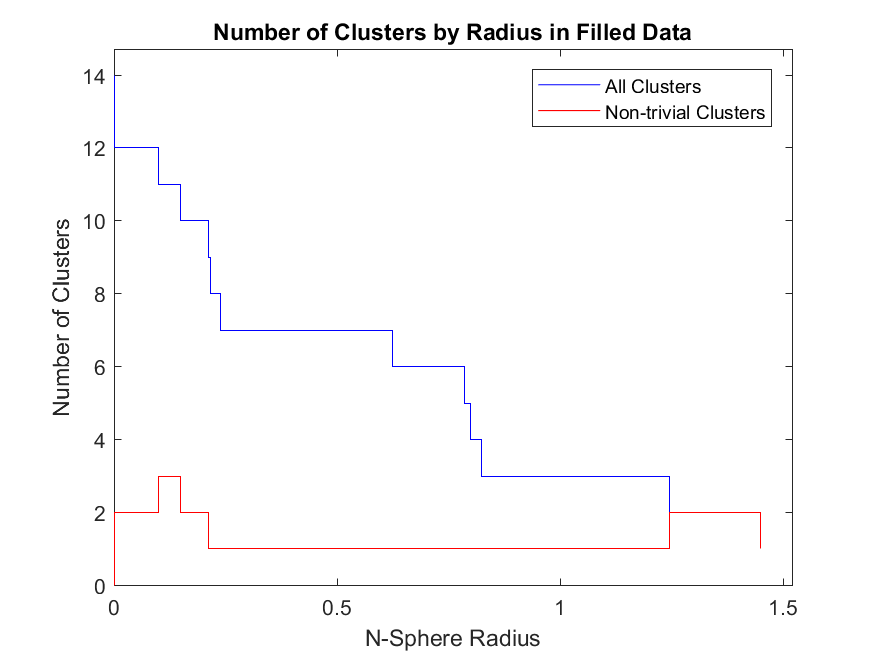}\\
\includegraphics[width = 6.25in]{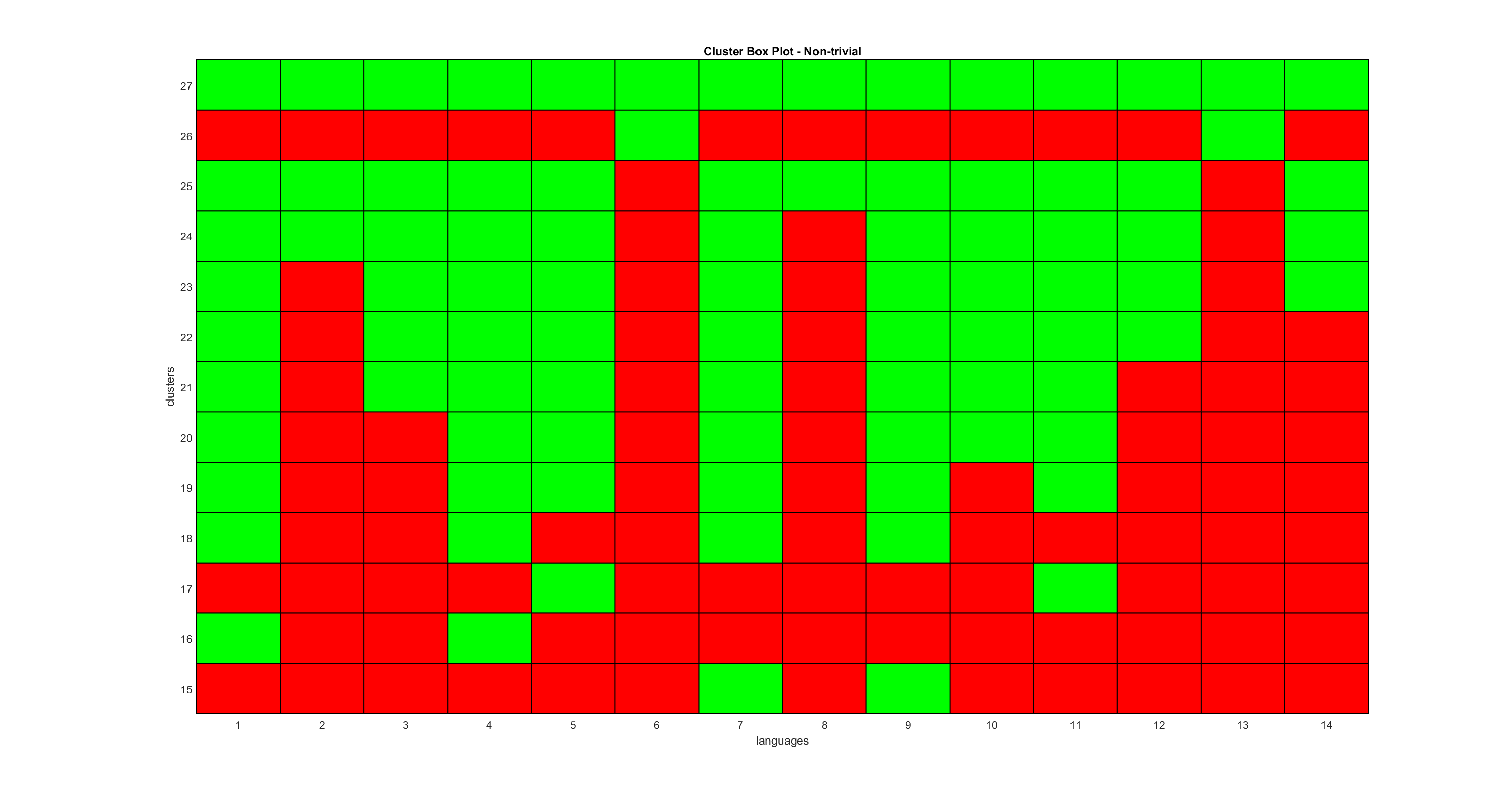}
\caption{Clusters and clustering structure in the persistent components tree for the Romance languages with filtered SSWL data, PCA $60\%$.\label{SSWLRomFiltertree2}}
\end{figure}

\smallskip
\subsubsection{Romance language family in the SSWL data}

The Romance languages are modern languages that evolved from Latin. The family is represented in the SSWL 
database by Bellinzonese,  Calabrian, Catalan, French, Galician, Italian,  various Italian dialects 
(Napoletano Antico, Old, Reggiano), Latin, Late Latin, Neapolitan, Occitan, Old French, Brazilian Portuguese, European Portuguese, Romanian, Sicilian, Spanish, Teramano. 

\smallskip

When we consider the full tree of the persistent components for the unfiltered SSWL data, 
most of the Romance languages are indeed grouped together in a subtree (which corresponds to Cluster $125$). 
This contains most of the Romance languages, including Italian, Portuguese, Spanish, Catalan, Sicilian. It also contains 
misplaced languages, such as English that is incorrectly placed outside of the German language family. 

\smallskip

Looking at this subtree we notice that, although it contains many of the Romance languages, it notably does not contain French and Romanian. 
The languages French and Old French are quite close to each other in the tree but are added later as singletons and are very far from the rest 
of the Romance languages. Romanian on the other hand is not as far from this subtree as French is, but is not close either. Another thing to note is 
that Romanian is clustered with European-Portuguese, which we would also expect to see together with the other Romance languages.  
It is known that Romanian shares grammatical features with non-Romance languages such as Greek, Bulgarian and Serbo-Croatian, and it is indeed 
placed closer to these languages in our tree. Going back to the subtree $125$ we also note that topology-wise the languages are not located accurately, 
as we would expect, for example, to have Spanish and Portuguese in closest proximity as Iberico-Romance languages. 

\smallskip

Thus, we encounter once again the same kind of problems illustrated in \cite{ShuMar2} when using the
full unfiltered SSWL data. 

\smallskip

We then consider only the subset of Romance languages that are at least $50\%$ complete
in the SSWL database and we select only the subset of SSWL syntactic variables that are
fully mapped for all of those languages. We analyze again the persistent components when
using only this set of Romance languages and only their subset of completely mapped parameters.

\smallskip

The resulting tree (Figure~\ref{SSWLRomFiltertree}) and clustering structure (Figure~\ref{SSWLRomFiltertree2})
has several interesting sub-structures. The clusters (from left to right in Figure~\ref{SSWLRomFiltertree}) show
as sub-cluster joining at the root top of the tree the correct placement of the ancient languages: cluster N.~26 
containing Latin and Late Latin. This is followed by three singleton clusters, that also join 
near the root of the tree (in descending order after the Latin cluster): N.~8 Romanian, N.~2 Northern Calabrian,
N.~14 Old French. There is then a large structure (cluster N.~22) consisting of several sub-structures.
Two two-language clusters, N.~16 with Italian and Brazilian Portuguese and N.~15 with Sicilian and 
Portuguese are joined together into cluster N.~18. This cluster then merges with another two-language
cluster, N.~17 containing French and Napoletano Antico, followed by two singleton clusters,
N.~10 Spanish and N.~3 Catalan. The clustering of Italian and the Southern Italian dialects together
with Brazilian Portuguese, Portuguese, and French, respectively appears at odds with historical phylogenetic
trees and the position of the main Romance languages also does not match the position expected
in the phylogenetic tree as we discuss in Section~\ref{RomCompareSec}.

\smallskip
\subsubsection{Romance family in the LanGeLin data} 

\begin{figure} 
	\subfloat[Romance languages tree]{\includegraphics[width = 6.3in]{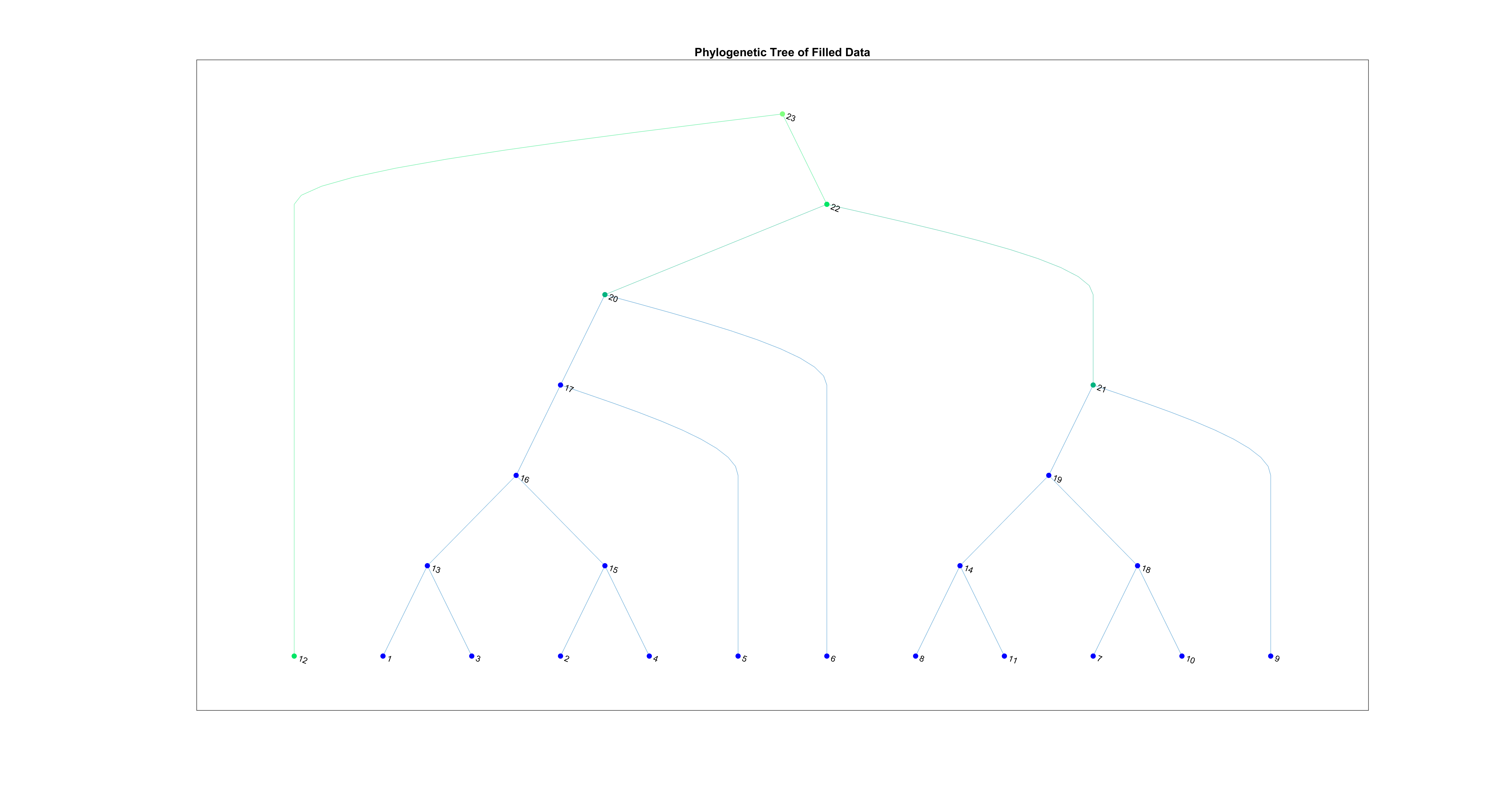}} \\
	\subfloat[Cluster N.~20]{\includegraphics[width = 5.3in]{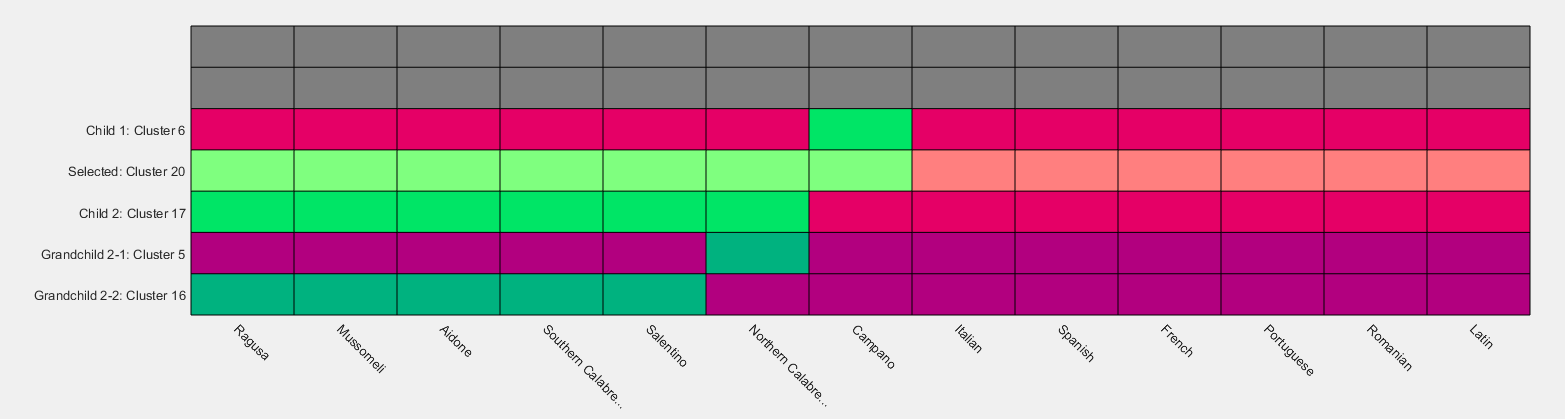}} \\
	\subfloat[Cluster N.~21]{\includegraphics[width = 5.3in]{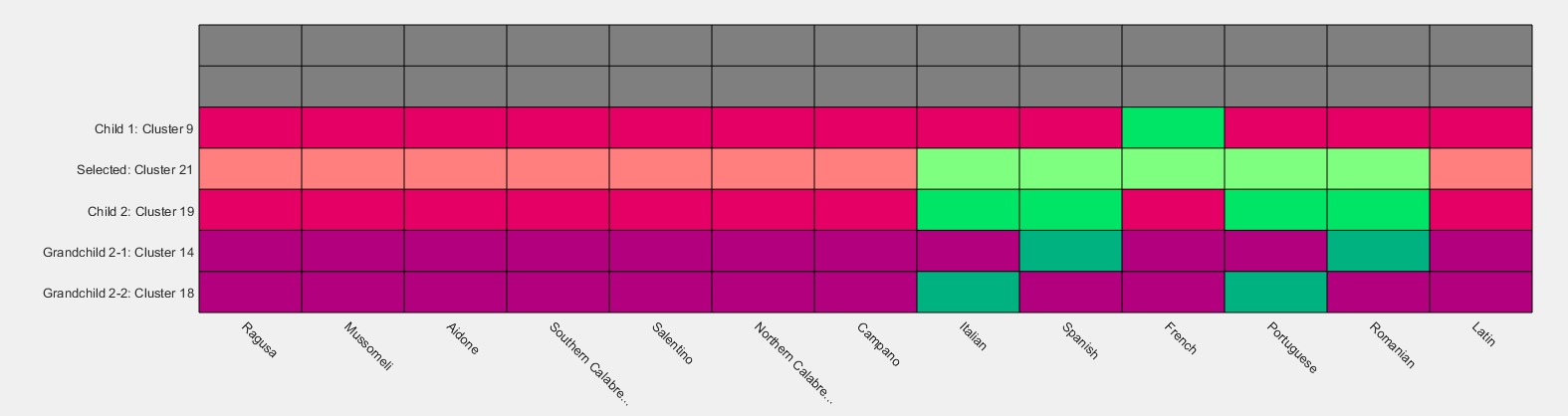}}
	
	\caption{Persistent components trees of the Romance languages from the LanGeLin data with PCA $60\%$.
	\label{RomanceLanGeLinFig}}
\end{figure}

In the case of the Romance languages, the LanGeLin data deliver a persistent components tree that is more
informative and accurate in terms of reflecting phylogenetic information than in the case of the Germanic 
languages discussed above.

\smallskip

The Romance languages represented in the LanGeLin data include Latin, the main modern Romance languages French, Spanish, Italian, Portuguese, Romanian,
and a set of Sourthern Italian dialects that were studied in \cite{Gua16} in relation to the Greek-Italian microvariations that we already mentioned. This group of
Sourthern Italian dialects is given by Ragusa, Mussomeli, Aidone, Northern Calabrese, Southern Calabrese, Salentino, and Campano. When we compute the
persistent components tree for the Romance languages in the LanGeLin data, the PCA at $60 \%$, we find the structure illustrated in Figure~\ref{RomanceLanGeLinFig}.
The tree illustrated in the figure corresponds to the structure of cluster N.~23. It contains a subcluster N.~12, which consists of Latin alone, and which should in fact
really correspond to the root of the tree, and two main subclusters, N.~20 and N.~21. The content of the two subclusters is illustrated in Figure~\ref{RomanceLanGeLinFig}:
we see that cluster N.~20 contains all the Southern Italian dialects, while cluster N.~21 contains all the main Romance languages. 

\smallskip

The persistent components tree for the cluster of the Southern Italian dialects has the form 
{\small
\begin{center}
\Tree [ Campano [ Northern-Calabrese [ [ Mussomeli Salentino ] [ Southern-Calabrese [ Aidone  Ragusa ] ] ] ] ]
\end{center}
}

\smallskip
\subsubsection{Comparison with the algebro-geometric method}\label{RomCompareSec}

In \cite{OSBM} the subset of Romance languages (Latin, Romanian, French, Spanish, Portuguese) is analyzed with the algebro-geometric method using 
a combination of the Longobardi and the SSWL databases, where in the SSWL only the parameters that were completely mapped were retained, 
and with the additional information that Latin should be regarded as the root vertex.

The topology of the tree obtained in \cite{OSBM} reflects the historical phylogenetic tree and is of the form 
\begin{center}
  \Tree [ .Latin [ Romanian [ Italian [ French [  Spanish Portuguese ] ] ] ] ]
\end{center}
The persistent component tree for the LanGeLin data analyzed above gives a tree with the inverted position of Italian and Spanish
\begin{center}
\Tree [ .Latin [ Romanian [ Spanish  [ French [ Italian Portuguese ] ] ] ] ]
\end{center}
It is interesting to notice that 
the tree topology obtained from the filtered SSWL data, when restricted only to the main 
Romance languages (Latin, Romanian, Italian, French, Spanish, Portuguese) is also of the same 
form as the one obtained from the LanGeLin data, with the positions of Italian and Spanish inverted
with respect to the historical phylogenetic tree.

\smallskip
\subsection{Hellenic language family} \label{HellSec}

The Hellenic language family in the SSWL, after filtering the data for completeness,
only consists of Ancient Greek, Homeric Greek, Medieval Greek, Modern Greek,
Cappadocian Greek, and Cypriot Greek. The associated tree has topology 
\begin{center}
\Tree [ .  [ Homeric Ancient ] [ Cappadocian [ Medieval [ Cypriot Modern ] ] ] ] 
\end{center}

The Hellenic family in the LanGeLin data is rich with the additional presence of
the Greek Southern-Italian dialects. The resulting tree topology is given by
{\small 
\begin{center}
\Tree [ .  [ Ancient New-Testament ] [ [ Calabrian-A [ Calabrian-B Salento ] ]  [ Romeyka-Pontic [ Cypriot Modern ] ] ] ] 
\end{center} }

\smallskip
\subsection{Balto-Slavic language family}\label{SlavicSec}

The Balto-Slavic language family is comprised by two main branches: the Baltic languages,
such as Lithuanian and Latvian, and the Slavic languages. In the SSWL database the
Balto-Slavic languages include Bulgarian, Croatian, Czech, Lithuanian, Polish, Russian,
Serbian, Slovenian, Ukrainian. 

\smallskip
\subsubsection{Balto-Slavic languages in the SSWL data}\label{SlavicSSWLSec}

When we work with the full unfiltered set of SSWL data the Balto-Slavic language family does not 
demonstrate a special structure and the Balto-Slavic languages are not even grouped together. 
Polish and Russian are added as singletons and are far away from the rest of the languages 
within the tree, and from each other. Two Slavic languages that are grouped together (but not in the 
closest proximity) are Serbian (West Slavic) and Bulgarian (East Slavic) and are placed closer to Lithuanian 
more than to Polish and Russian. There aren't enough languages in this group left after filtering the
SSWL data to apply the previous technique.

\smallskip
\subsubsection{Balto-Slavic family in the LanGeLin data} \label{SlavicLanGeLinSec}

	The set of Slavic languages included in the LanGeLin data is given by Bulgarian (Blg), Serb-Croatian (SC), Slovenian (Slo), Polish (Po), Russian (Rus). 
	As we already observed before, in the tree obtained from the persistent connected components 
we find the Slavic languages grouped together near the Germanic languages, in the same 
sub-cluster structure that also contains the Greek languages and Latin. An interesting phenomenon
one observes in this cluster is the fact that Bulgarian is not adjacent to the other Slavic languages
but is placed in proximity to the Greek languages. The resulting tree structure (drawn without
resolving the substructures corresponding to the Germanic and Greek languages) has the shape {\small
\begin{center}
\Tree [ [ [ [ [ [ Slovenian Russian ] Polish ] Serb-Croatian ] Germanic ] Hellenic ] Bulgarian ]
\end{center}  }
This structure appears to suggest the presence of a loop involving the Slavic and the Greek languages (possibly involving some
of the Germanic languages). The presence of a loop involving the Hellenic branch and some of the Slavic languages was
already observed in \cite{Port} (based on the analysis of a cluster in the SSWL data). We will discuss more in detail the $H_1$-structures in 
Section~\ref{H1sec} and in particular a persistent $H_1$-generator involving some Greek and Slavic languages and Gothic, which may
be another manifestation of the same linguistic phenomenon of syntactic relatedness.

\smallskip
\subsubsection{Comparison with the algebro-geometric method}

In \cite{OSBM} the set of Slavic languages (Polish, Russian, Bulgarian, Serb-Croatian, Slovenian) is analyzed using
the combined LanGeLin and SSWL data (retaining only completely mapped variables). The phylogenetic algebraic
geometry method correctly identifies the phylogenetic tree for this set of
	languages according to historical linguistics, given by 
\begin{center}
\Tree [ Polish [ Russian [ Bulgarian [ Serb-Croatian Slovenian ] ] ] ] 
\end{center}

\smallskip

By isolating the Slavic languages in the LanGeLin data persistent components tree discussed above we find a tree with topology
\begin{center}
\Tree [ Bulgarian [ Serb-Croatian [ Polish [ Russian Slovenian ] ] ] ]
\end{center}
This is not reflecting the phylogenetic structure of the Slavic languages, nor does it reflect the correct placement of
Bulgarian in the South-Slavic subbranch. Thus, this is another example where one sees that the information on
syntactic relatedness captured by the persistent connected components tree is not the same as the phylogenetic
information about historical language development, although it correlates to it in terms of grouping together (most of)
the languages within the subfamily.

\smallskip
\subsection{The hypothetical Ural-Altaic family}\label{UrAltSec}

The LanGeLin data include the Uralic languages Estonian, Finnish, Hungarian, Udmurt, Yukaghir, Khanty
and the putative Altaic languages Turkish, Buryat, Yakut, Even, Evenki, as well as Japanese and Korean
that were proposed, in the early developments of the Ural-Altaic hypothesis, as other possible members
of an Altaic family, although these two languages were later discarded from the Altaic hypothesis.

The persistent components tree for the LanGeLin data places Japanese and Korean in closest proximity to
each other, but at the outskirts of the tree, far away from the other Altaic and Uralic languages. The rest of
the Uralic and Altaic languages are all placed very closely together in cluster N.~$95$, with the exception
of the Uralic language Yukaghir that is placed together with the Indo-Iranic languages. The persistent
components tree for the Ural-Altaic languages in cluster N.~$95$ is of the form {\small
\begin{center}
\Tree  [ [ [ Finnish Estonian ] [ [ Hungarian Khanty ] [ Even Eveki ] ] ]    [ Buryat [ Udmurt [ Turkish Yakut  ] ] ]  ] 
\end{center} }
Notice that, although Udmurt is an Uralic language, it is placed in the sub-cluster with the Altaic languages ,
while the sub-cluster formed by the Altaic languages Even and Eveki is placed within a cluster with the Uralic
languages Estonian, Finnish, Hungarian, and Khanty.

\smallskip
\subsection{Examples from other language families in the SSWL data}

An extensive representation of non-Indo-European language families is only available in the SSWL data set. In the LanGeLin data set, with the exception of the Ural-Altaic hypothesis discussed above, the non-Indo-European  languages are too few to make any meaningful analysis, except for noticing the fact that they are grouped 
together in their own $2$-leaf subtrees within the outer layer of the tree, as in the case of Arabic and 
Hebrew (both Semitic languages).  In the SSWL database, languages in non-Indo-European families tend to
be less completely mapped than the Indo-European ones. Thus, the filtering process we described above (retaining only languages
that are at least $50\%$ complete and retaining only syntactic variables that are fully mapped for those languages) tends to eliminate
too many languages from these families. For this reason, we will present here the persistent components trees
for these families for the full unfiltered SSWL data. These will necessarily be less reliable as they will resent from the incompleteness of the data.

\subsubsection{Niger-Congo family}

%\begin{figure} 
%	 \includegraphics[width = 7.5in, angle=90]{SSWL-atlantic-congo_data.png} 
%	\caption{\Note{OLD:} Persistent components tree for the Niger-Congo language family, unfiltered SSWL data
%\label{NCSSWLtree}}
%\end{figure}

\begin{figure} 
	 \includegraphics[width = 7.5in, angle=90]{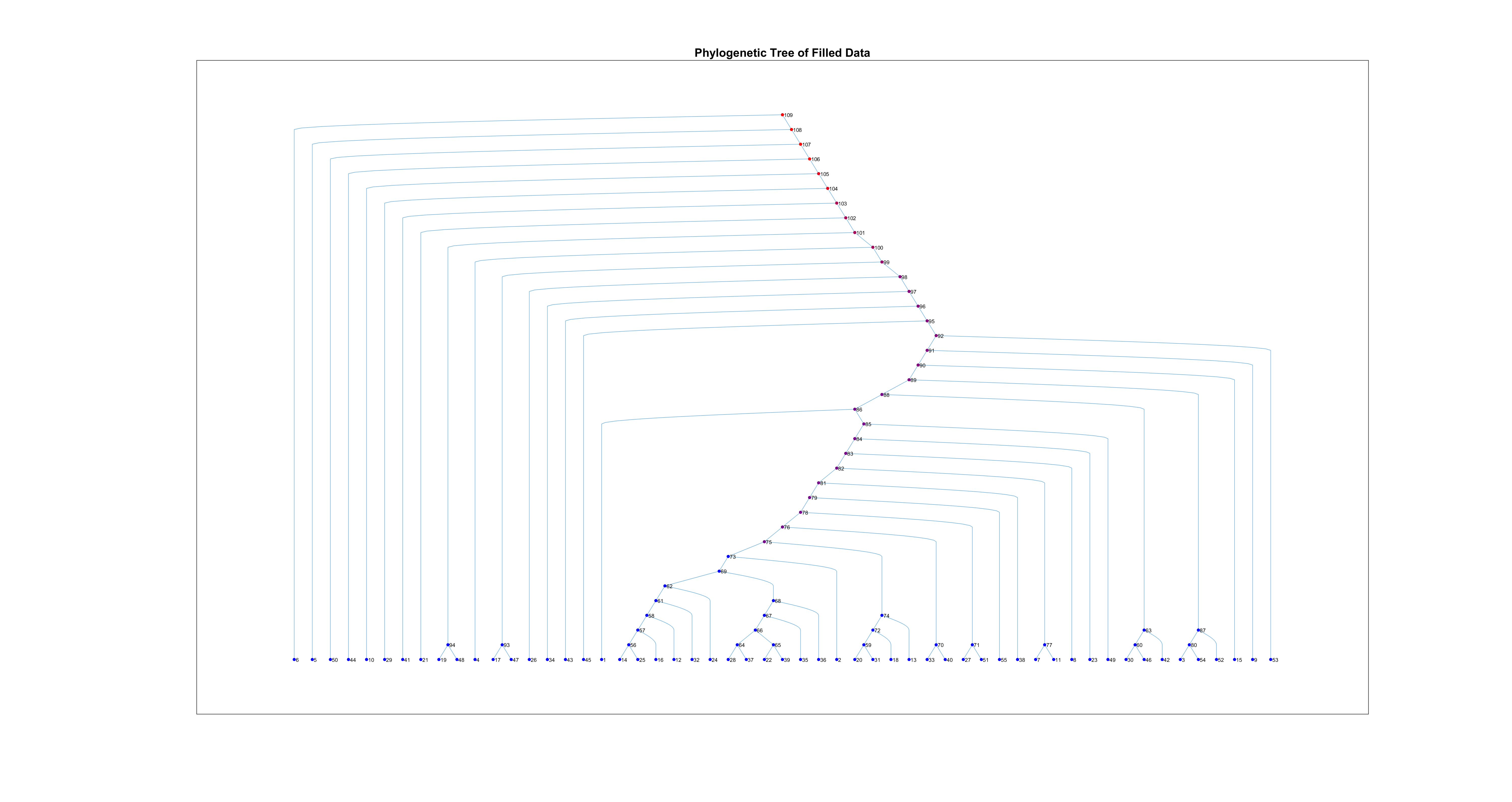} 
	\caption{Persistent components tree for the Niger-Congo language family, SSWL data.\label{NCSSWLtree}}
\end{figure}

\smallskip

The Niger-Congo languages are very well represented in the SSWL database with over sixty languages
(see the Introduction). 
When we consider the full unfiltered SSWL data, see Figure~\ref{NCSSWLtree}, 
the Niger-Congo tree of persistent compoments exhibits one main large cluster (N.~69), which
in turn splits into two main sub-clusters (N.~62 and N.~68). Cluster N.~62 contains the languages
Ewondo, Fe'efe'e, Ghomala',  Kaiama Ijo,  Kenyang and Koyo, while cluster N.~68 contains the
languages Igala, Kindendeule, Mankanya, Medumba, Naki, Ndut. Among the smaller structures 
we find historically relevant clusters N.~74, which contains the languages Farefari, Gurene, Hanga, and Konni
and N.~63 with Kom, Nweh, and Tukombo Tuki. 

\smallskip

Ewondo, Fe'efe', Ghomala, Kenyang, and Koyo are all within the Southern Bantoid languages
which include the Bantu group (Koyo is a Bantu C language), while Ijo (or Ijaw) is a group of
putative Niger-Congo languages that are generally viewed as outside of the main branched of the
Niger-Congo family. Thus, with the exception of the presence of the Kaiama Ijo language, 
the rest of cluster N.~62 can be seen as a Southern Bantoid structure. Within this cluster
Fe'efe'e and Kenyang form the deepest sub-structure, to which Ghomala, Ewondo, Koyo, 
and Kaiama Ijo are successively added, with the only non-Bantoid language added last. 

\smallskip

In cluster N.~68 we find that 
Mankanya belongs to the Bak group, and Ndut is a Senegambian Cangin language. 
Bak and Senegambian languages are considered close and often grouped together, 
so having Mankanya and Ndut in the same substructure is consistent with this proximity.
Kindendeule is again a Southern Bantoid language, Naki is an East Beboid language, which
is also part of the Southern Bantoid group, and Medumba is also Southern Bantoid (Grassfields
group). The language Igala, on the other hand, is a Volta-Niger Yoruboid language, which
belongs to yet another branching of the Niger-Congo languages. 
The Southern Bantoid family and the Bak and 
Senegambian groups do not belong to the same sub-branching of the Niger-Congo family: 
Bak and Senegambian are part of the Atlantic branch while the Southern Bantoid languages are part
of the Benue-Congo group inside the Volta-Congo branch, and Igala is part of the Volta-Niger group of 
the Volta-Congo branch.  Thus, the close proximity of the persistent
connected components of these languages in cluster N.~68 does not reflect their historical relatedness. 
The structure within this cluster shows the pairing of Kindendeule and Naki (cluster N.~64) and
Igala and Ndut (cluster N.~65): both of these mix different subfamilies of the Niger-Congo family.
These two-language clusters are then joined together (cluster N.~66) and successively joined
by the Bak language Mankanya and the South Bantoid language Medumba. 
The language Babanki, which joins just above cluster N.~69 (cluster N.73), is also a Southern Bantoid language.

\smallskip

Among the smaller clusters, cluster N.~74 is made entirely of Atlantic-Congo Gur languages, 
with Hanga and Konni placed in closest proximity, joined by Gurene and Farefari. 
Cluster N.~63 is made of Southern Bantoid languages, with Kom and Nweh in Grassfields group 
and Tuki in the Mbam group. Thus, with the exception of the mixing of different branches
that occurs in cluster N.~68, the main clusters are either Southern Bantoid or Gur structures.

\smallskip
\subsubsection{Austronesian family}

%\begin{figure} 
%	 \includegraphics[width = 7.5in, angle=90]{SSWL-austronesian_data.png}
%	\caption{\Note{OLD:} Persistent components tree for the Austroneasian language family, SSWL data.\label{AustroSSWLtree}}
%\end{figure}

\begin{figure} 
	 \includegraphics[width = 7.5in, angle=90]{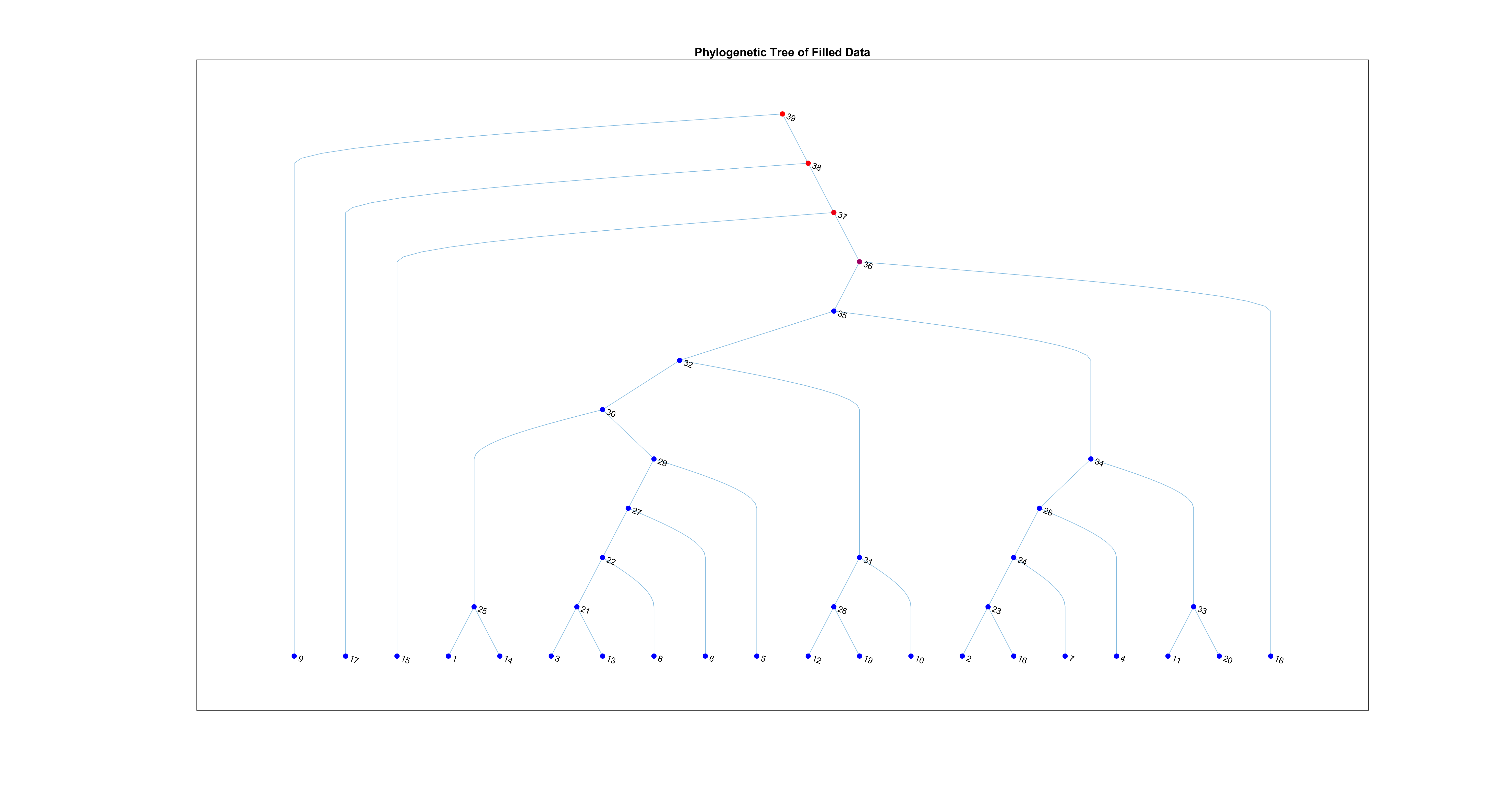}
	\caption{Persistent components tree for the Austroneasian language family, SSWL data.\label{AustroSSWLtree}}
\end{figure}

The internal structure of the Austronesian languages is complex. The family consists of many similar and closely related languages with large numbers of dialect continua, making it difficult to recognize boundaries between branches and subfamilies. 

\smallskip

The persistent components tree obtained from the SSWL data of Austronesian languages has
some large structures and only four singletons added near the root of the tree  
(see Figure \ref{AustroSSWLtree}). The large structures
consist of two main subtrees (clusters N.~32 and N.~34), with cluster N.~32 containing
a main sub-cluster N.~30. 

\smallskip

The structure of cluster N.~30 includes a pair (cluster N.~25) consisting of 
Palue, a Malayo-Polynesian Flores language and Kayan, a 
Malayo-Polynesian Borneo language. 
It also contains another two-language sub-cluster  (N.~21)
with Squliq Atayal, a Northern Formosan language, together with
the Tongic Polynesian language Niuean. Both of these languages are 
among the most poorly mapped in this family so their occurrence together 
should be regarded as an accident due to the incompleteness of the data.
Sub-cluster N.~29 of cluster N.~30 also contains Isbukun Bunun, 
which is a Formosan language, Ilokano (a Malayo-Polynesian  
Philippine language), and the Malayo-Polynesian Oceanic language Fijian.
Cluster N.~32 also contains a two-language cluster with Maori (Malayo-Polynesian 
Oceanic Eastern-Polynesian Tahitic) and Tongan (which is Malayo-Polynesian 
Oceanic Polynesian-Tongic). 
Thus, cluster N.~35 does not reflect a historical linguistic grouping,
since it includes Malayo-Polynesian languages of different groups
(Flores, Borneo, Philippine, Oceanic) as well as Formosan languages, but
it can be seen as prevalently a Malayo-Polynesian structure.

\smallskip

Cluster N.~34 contains Acehnese (a Malayo-Polynesian
Malayo-Sumbawan language),
West Coast Bajau (a Malayo-Polynesian Borneo language),
Indonesian (Malayo-Polynesian Malayic), 
Marshallese (Malayo-Polynesian Micronesian), 
Sasak (Malayo-Polynesian Malayo-Sumbawan),
Tukang Besi (Malayo-Polynesian Celebic).  Among these,
the two Malayo-Sumbawan languages Acehnese
and Sasak are grouped together in a two-language
sub-cluster (N.~23) which agrees with historical 
proximity, while the two-language cluster N.~33
with Marshallese and Tukang Besi does not 
reflect historical proximity. 

\smallskip

The two clusters N.~32 and N.~34 merge into cluster N.~35. 
The remaining singletons are Titan (a Malayo-Polynesian Oceanic Manus language) joining
just above cluster N.~35 (cluster N.~36), followed by merging with Samoan (a Malayo-Polynesian 
Polynesian language) in cluster N.~37, with Tagalog (a Malayo-Polynesian Philippine language)
and Malagasy (the Malayo-Polynesian East Barito language of Madagascar), which is the
farthest away and last to merge with the rest of the tree.  The structures seen in the
persistent component tree of the Austronesian languages are primarily
Malayo-Polynesian, but the historical linguistic subdivisions of this family into subfamilies
is not preserved by the clustering of connected components.

\smallskip
\subsubsection{Afro-Asiatic family}

\begin{figure} [!htb]  
\includegraphics[width = 7.6in, angle=90]{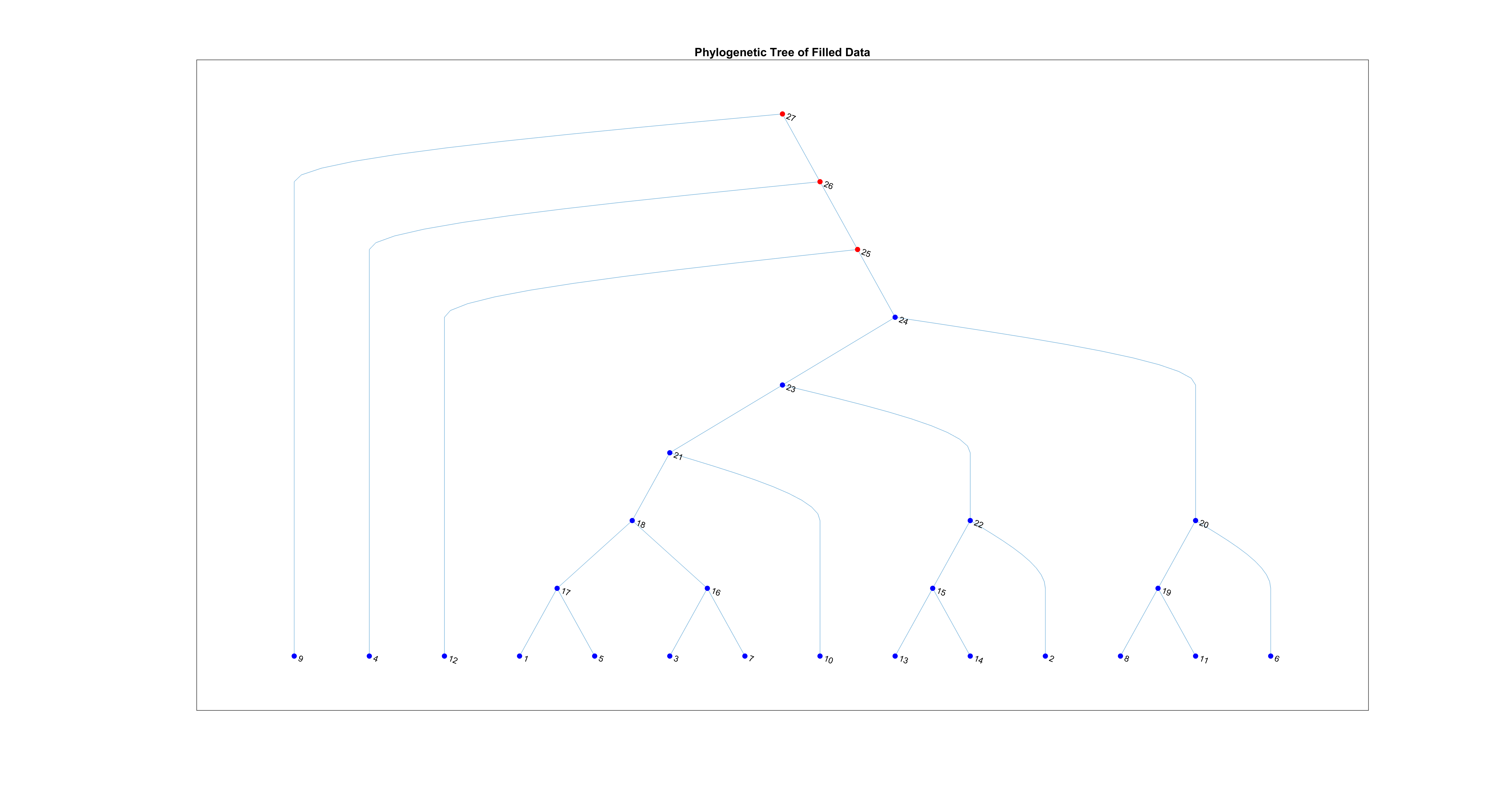}
\caption{Persistent components tree of the Afro-Asiatic language family, SSWL data, PCA $60\%$. \label{SSWL_AA_phylotree}}
\end{figure}

The persistent components tree for the Afro-Asiatic languages obtained from the SSWL data  (see
Figure~\ref{SSWL_AA_phylotree}) shows a subdivision into
different small substructures.  
We find a singleton cluster N.~9 with  Amharic which joins the tree very close to the root; 
just below comes singleton cluster N.~4 with Hebrew and singleton N.~12 with Muyang. 
We have then a main structure, cluster N.~24 with two sub-structures, N.~20 and
N.~23 (in turn split into N.~21 and N.~22). 

\smallskip

Cluster N.~20 has Bole, Hausa, Miya and Moroccan Arabic (with the last two grouped together).
The first three are all Chadic languages, and Moroccan Arabic is placed in closer proximity to 
Miya of the Chadic group of Berber languages by which is historically influenced rather than together with
the other Arabic languages in the Semitic languages cluster N.~18. This
cluster has good correlation to a historical linguistic grouping. 

\smallskip

Cluster N.~22 has Wolane, Tigre, and Senaya. The clustering together
of Senaya and Tigre is likely an accident due to them being very incompletely mapped
languages in this group. Indeed, they should not belong in the same sub-structure: Tigre is a
South Semitic Ethiopic language, which is correctly placed close to Wolane and should also be close to 
Amharic (which occurs here as a singleton joining at the top of the tree), while Senaya is a 
Central Semitic Aramaic language. 

\smallskip

Cluster N.~21 sees two language pairs, N.~17 with Biblical Hebrew and Gulf Arabic
and N.~16 with Egyptian and Lebanese Arabic, merging into a cluster N.~18 of Semitic 
languages. Note that the fact that Hebrew occurs as a separate singleton N.~4, placed 
very far away from this cluster, is not an effect created by incompleteness of the data, 
as the level of completeness of Biblical Hebrew and Hebrew in the database is comparable 
and significant. It is also not an issue introduced by the PCA since the relative positions 
of these languages remains the same with PCA $80\%$. The cluster N.~18 of Semitic 
languages is then joined by the Chadic language Mbuko (which is not placed among
the other Chadic languages in cluster N.~20). 

\smallskip

%\begin{figure} [!htb]  
%\includegraphics[width = 7.6in, angle=90]{SSWL_AA_tree.png}
%\caption{\Note{OLD:} Persistent components tree of the Afro-Asiatic language family, SSWL data, PCA $60\%$. 
%\label{SSWL_AA_phylotree}}
%\end{figure}	

There are different proposals in historical linguistics regarding the phylogenetic tree of
the Afro-Asiatic languages.
A possible historical phylogenetic tree was discussed in \cite{Militarev}.
The only languages that are in the intersection of the Afro-Asiatic languages listed in \cite{Militarev} and
those listed in the SSWL database are Miya, Hausa, Arabic, Lebanese-Arabic, Hebrew, Amharic, Wolane,
Tigre. For this subset of languages, the phylogenetic tree proposed in \cite{Militarev} has the
topology {\small 
\begin{center}
\Tree [ [ [ Miya Hausa ]  [ [ Arabic Lebanese-Arabic ] Hebrew ] ] [ [ Amharic Wolane ] Tigre ] ] 
\end{center} }

\smallskip

While the persistent component tree correctly places Miya and Hausa in the same
substructure and Arabic, Lebanese Arabic, and Biblical Hebrew also in a substructure,
it does not correctly place Amharic in the same substructure with Wolane and Tigre.
Some misplacements are likely due to
problems of incompleteness of the data and the poor mapping of some of the languages.

\medskip
\section{Persistent First Homology}\label{H1sec}

We discuss in this section the presence of non-trivial generators of the persistent first 
homology $H_1$ for the data of languages in the
SSWL and the LanGeLin database. We first observe that the behavior of persistent 
homology for these syntactic data is different from
the typical behavior of random simplicial sets. We then discuss the method we 
follow for identifying specific representatives of 
generators of the persistent first homology and we discuss an example that 
appears to be detecting syntactic homoplasy phenomena as well as another example
that instead may have a possible historical linguistic interpretation. 
We also show the barcode structure of $H_1$-generators for the
filtered SSWL data, over individual language families. 
The large set of generators over the full database is more likely to be due to  
homoplasy detection, while $H_1$-generators within language families have more
chances of representing possible historical influences between languages
across different branches that happened at the syntactic level. 

\begin{figure} [!ht]	
	\subfloat[random barcode]{\includegraphics[width = 3in]{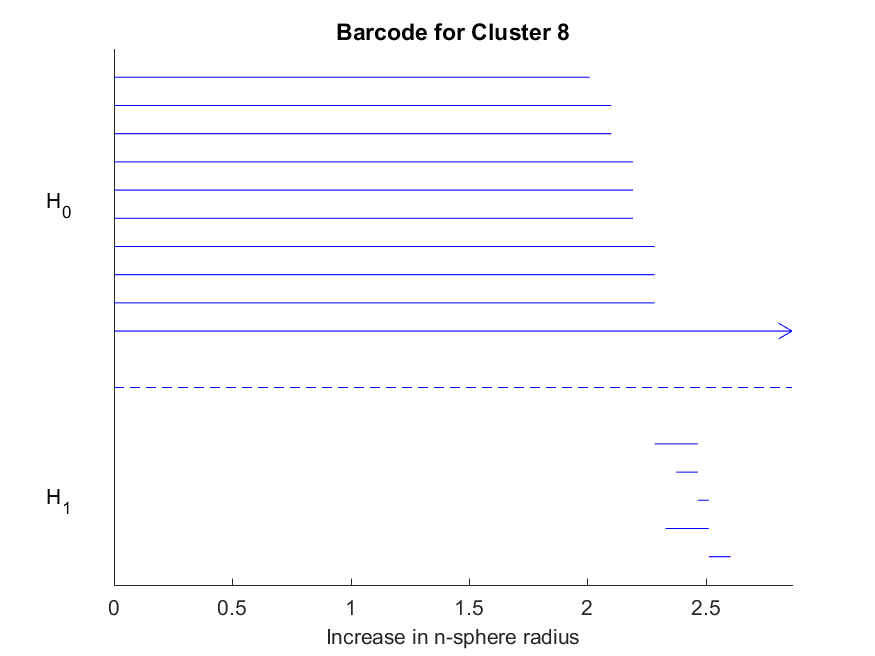}}
	\subfloat[random barcode]{\includegraphics[width = 3in]{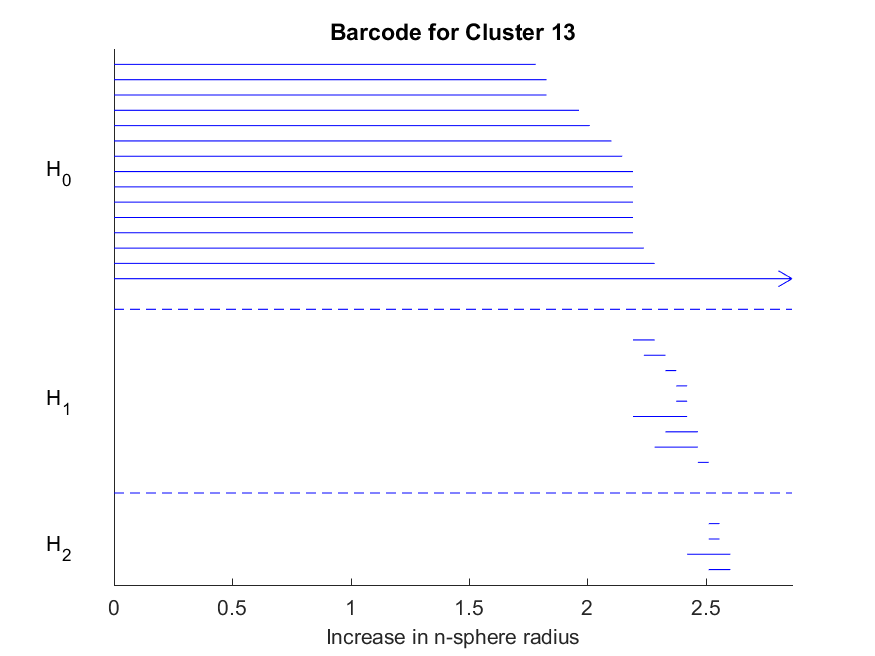}} 
	
	\caption{Barcodes of random data sets. The left barcode is of a smaller cluster and already shows several $H_1$-generators. \label{random_barcodes}}
\end{figure}

\subsection{Behavior of persistent homology on random data sets}
We ran the same analysis via persistent components and persistent homology computations 
on random data sets of binary vectors (and also for non discrete sets) with varying PCA values. 
The main differences are that in the random data sets $H_1$ appears also for smaller clusters, 
whereas in the case of the syntactic data from both the SSWL and the LanGeLin databases non-trivial 
$H_1$ generators are starting to appear only for bigger clusters (over the full database, for instance, they
are seen only in clusters containing at least 30 different languages). Moreover, the structure of the $H_1$
generators themselves is different. On the random data sets (see Figure~\ref{random_barcodes}) we always see that the $H_1$ generators are of shorter length, stacked on top of each other and less spread out. Also, there is typically a significantly larger number of generators for the $H_1$. On our data sets on the other hand, we also
observe the presence of $H_1$ generators with a longer length of persistence in the barcode diagram, and they are also typically more sparse. This suggest that in our case the $H_1$ structure is more persistent and also less coincidental, hence more likely to reveal some genuine underlying structure in the data.

\begin{figure}   [!ht]
	\subfloat[Cluster N.~102]{\includegraphics[width = 6.5in]{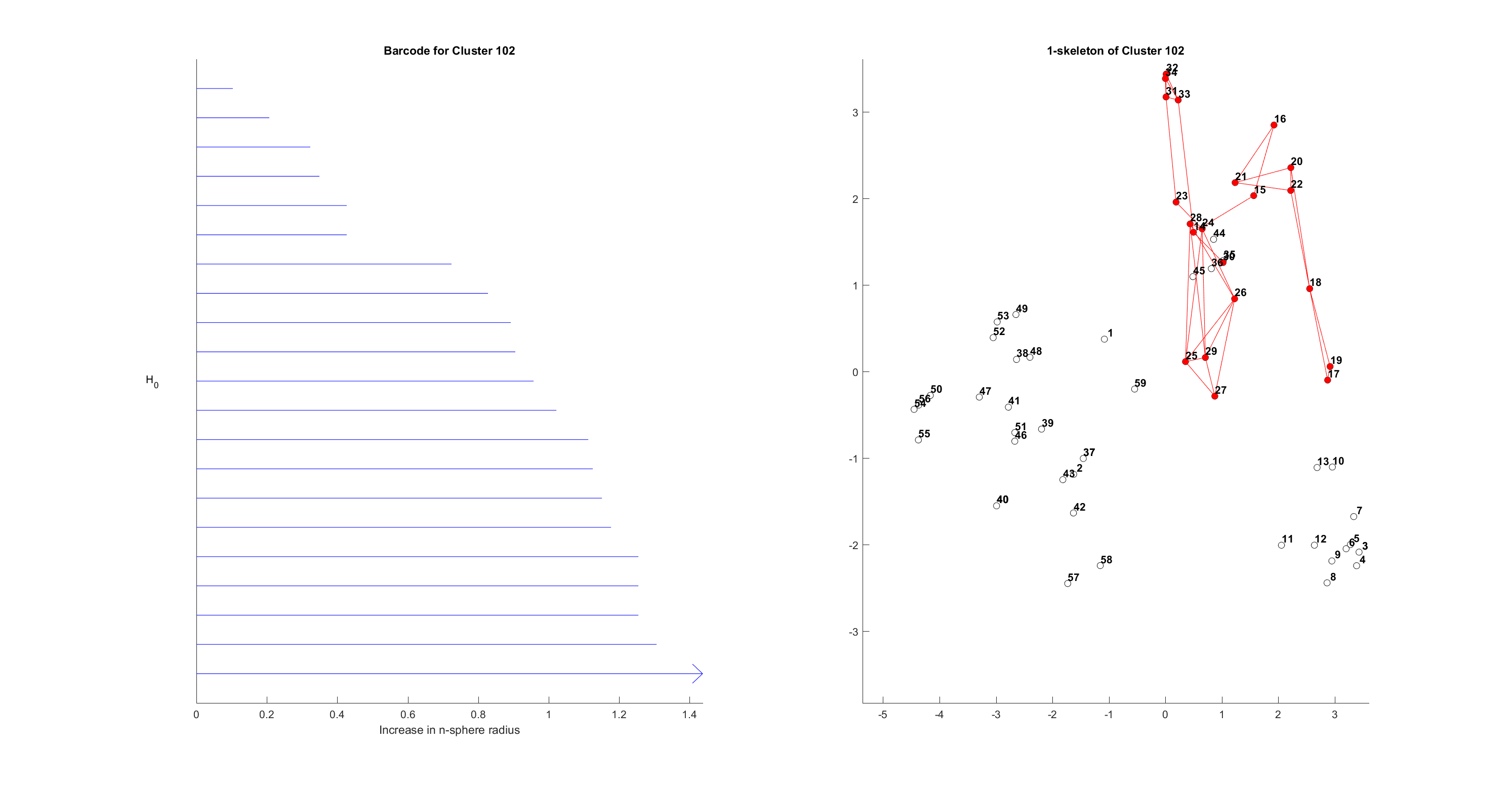}} \\
	\subfloat[Cluster N.~113]{\includegraphics[width = 6.5in]{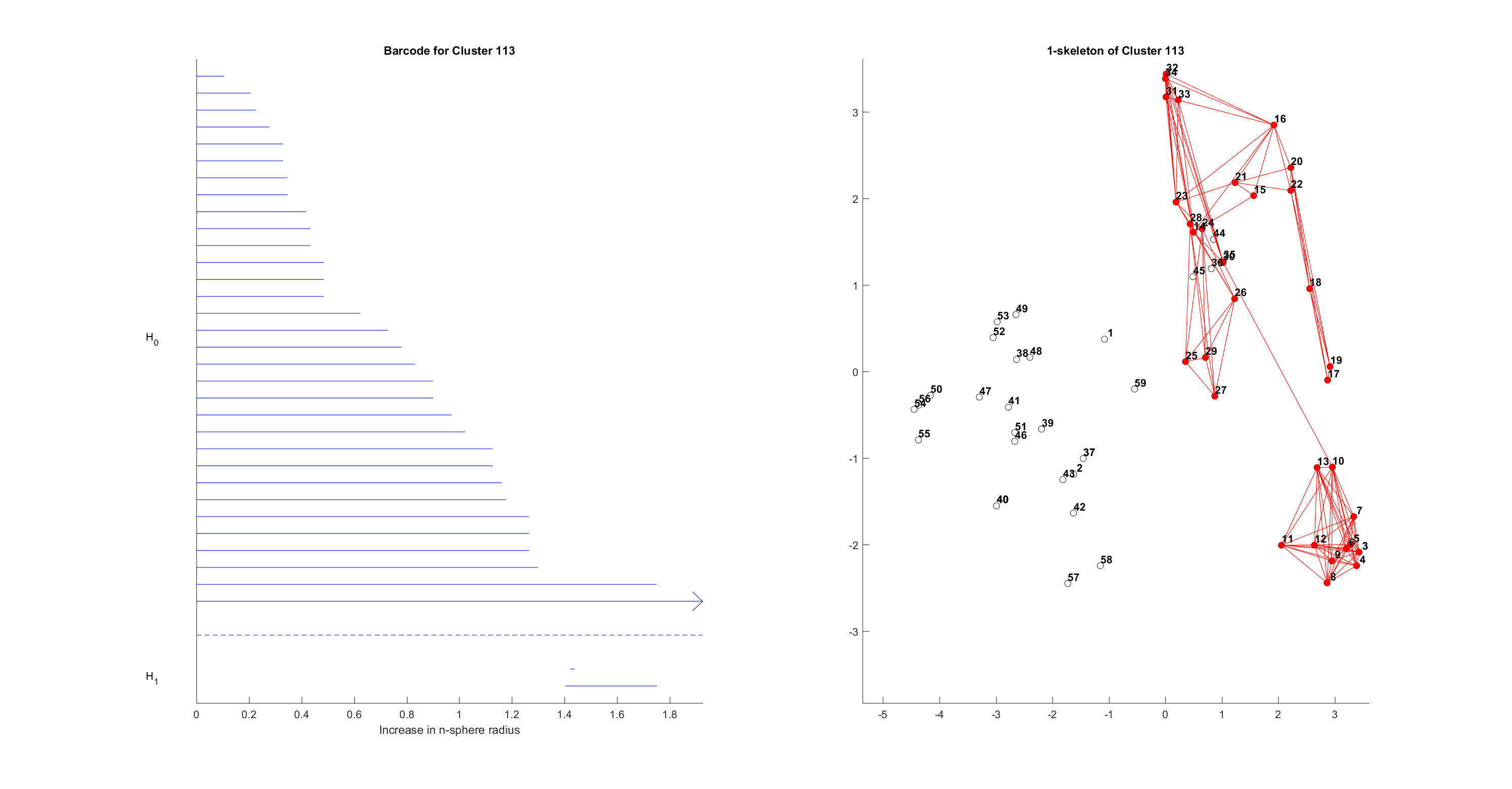}}
	\caption{Formation of a non-trivial persistent $H_1$-generator in the Indo-European languages (LanGeLin data): the Gothic--Slavic--Greek loop. \label{GothicGreekSlavicLoopFig}}
	\end{figure}	
	
	\begin{figure}   [!ht]
 	\subfloat[Cluster N.~91]{\includegraphics[width = 6.4in]{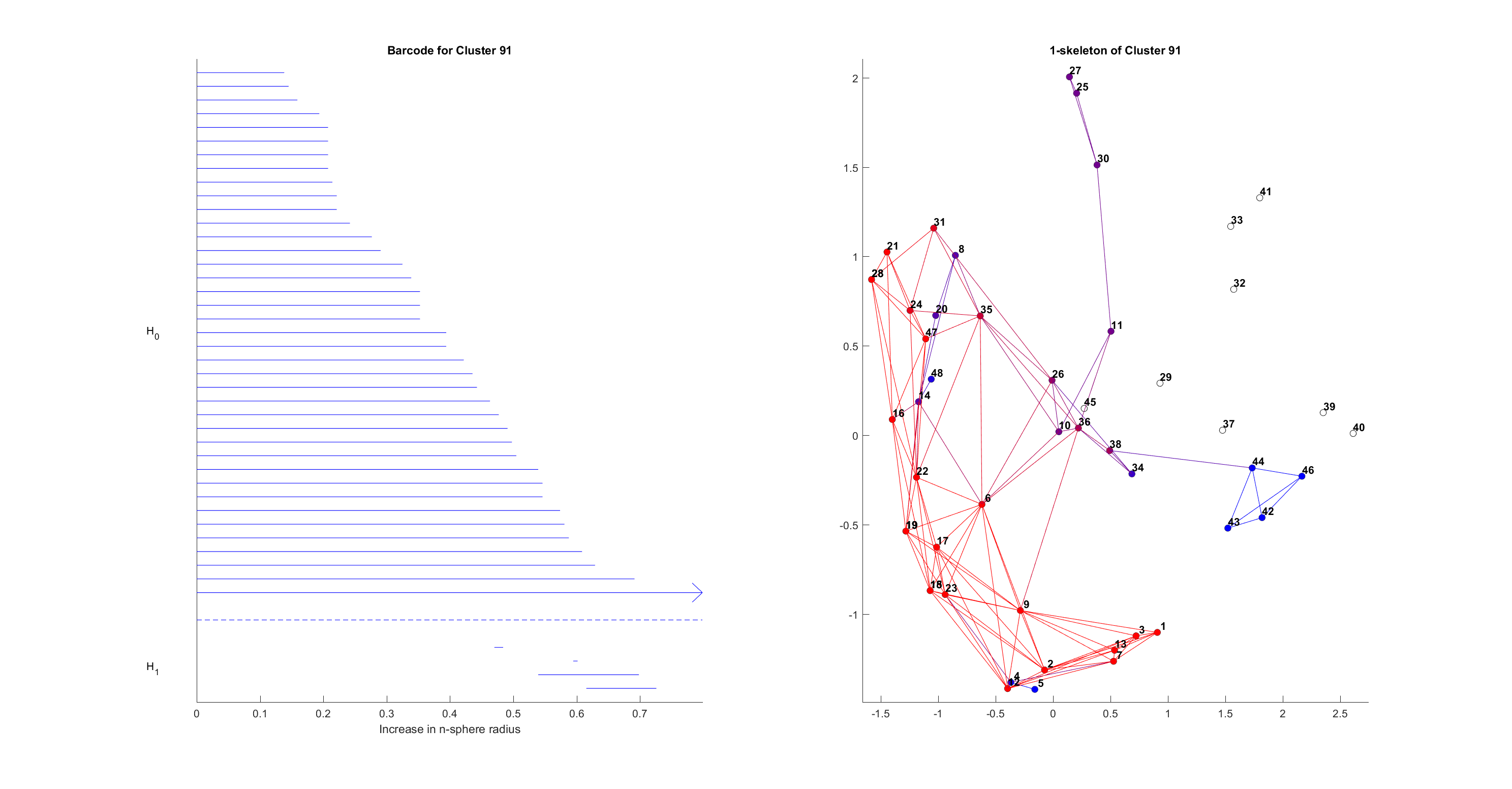}} \\
	\subfloat[Cluster N.~95]{\includegraphics[width = 6.4in]{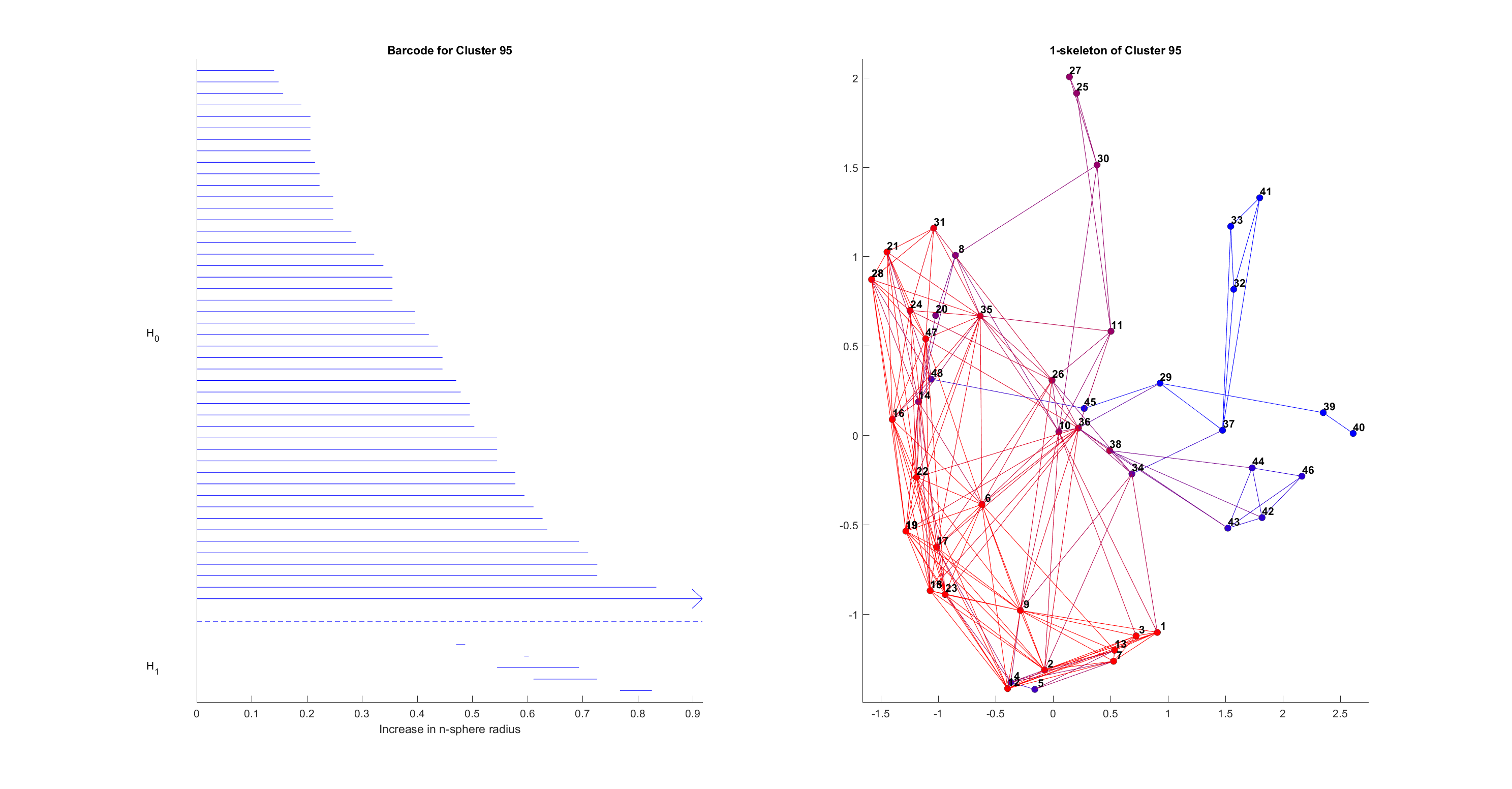}} \\
	\caption{Persistent $H_1$-structures in the filtered SSWL data (including both Indo-European and Ural-Altaic languages), PCA $60\%$, clusters
	N.~91 and N.~95. \label{IEUAH1Fig}}
\end{figure}

\begin{figure}   [!ht]
 	\subfloat[Cluster N.~75]{\includegraphics[width = 6.6in]{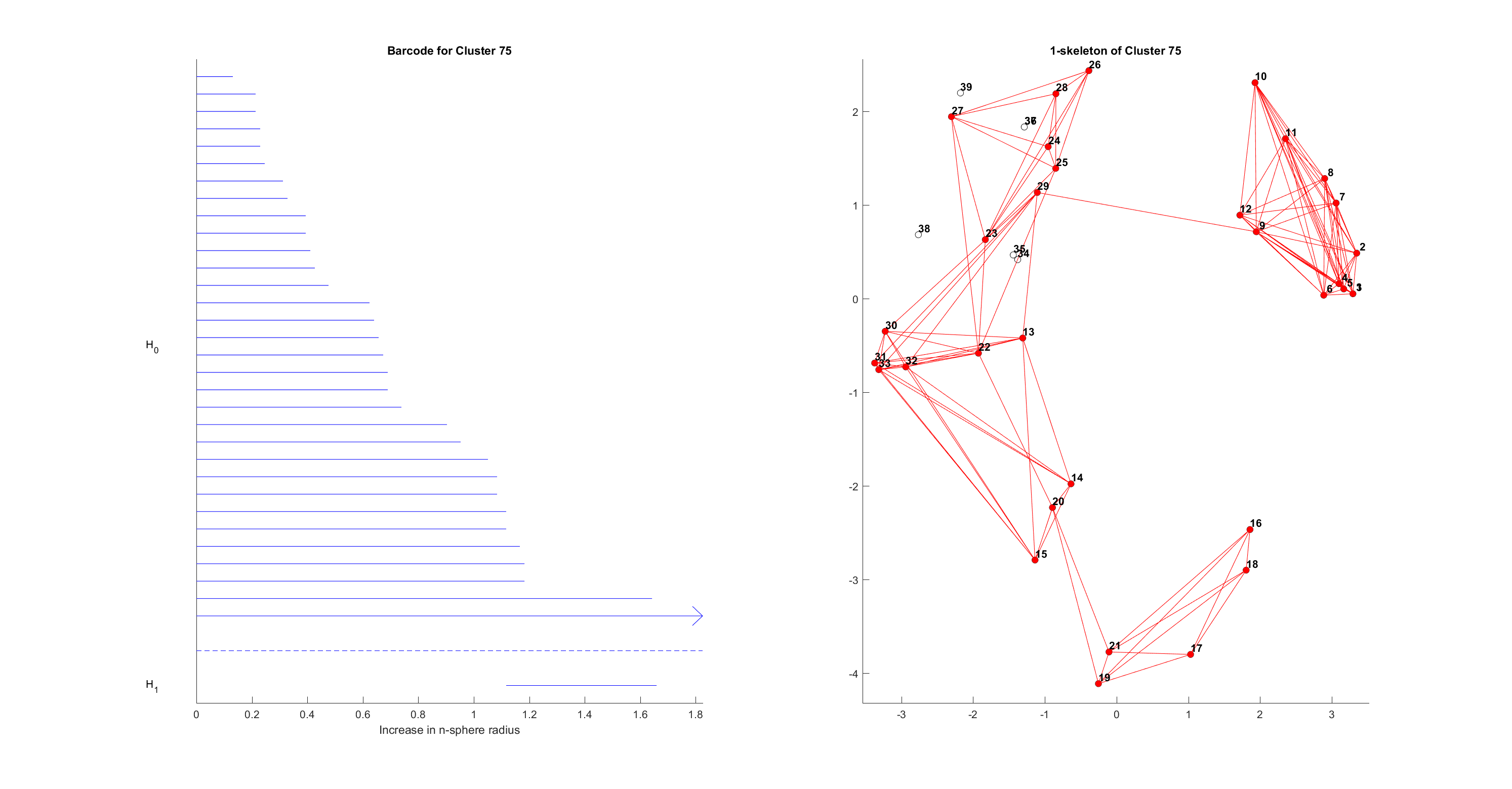}} \\
	\subfloat[Cluster N.~77]{\includegraphics[width = 6.5in]{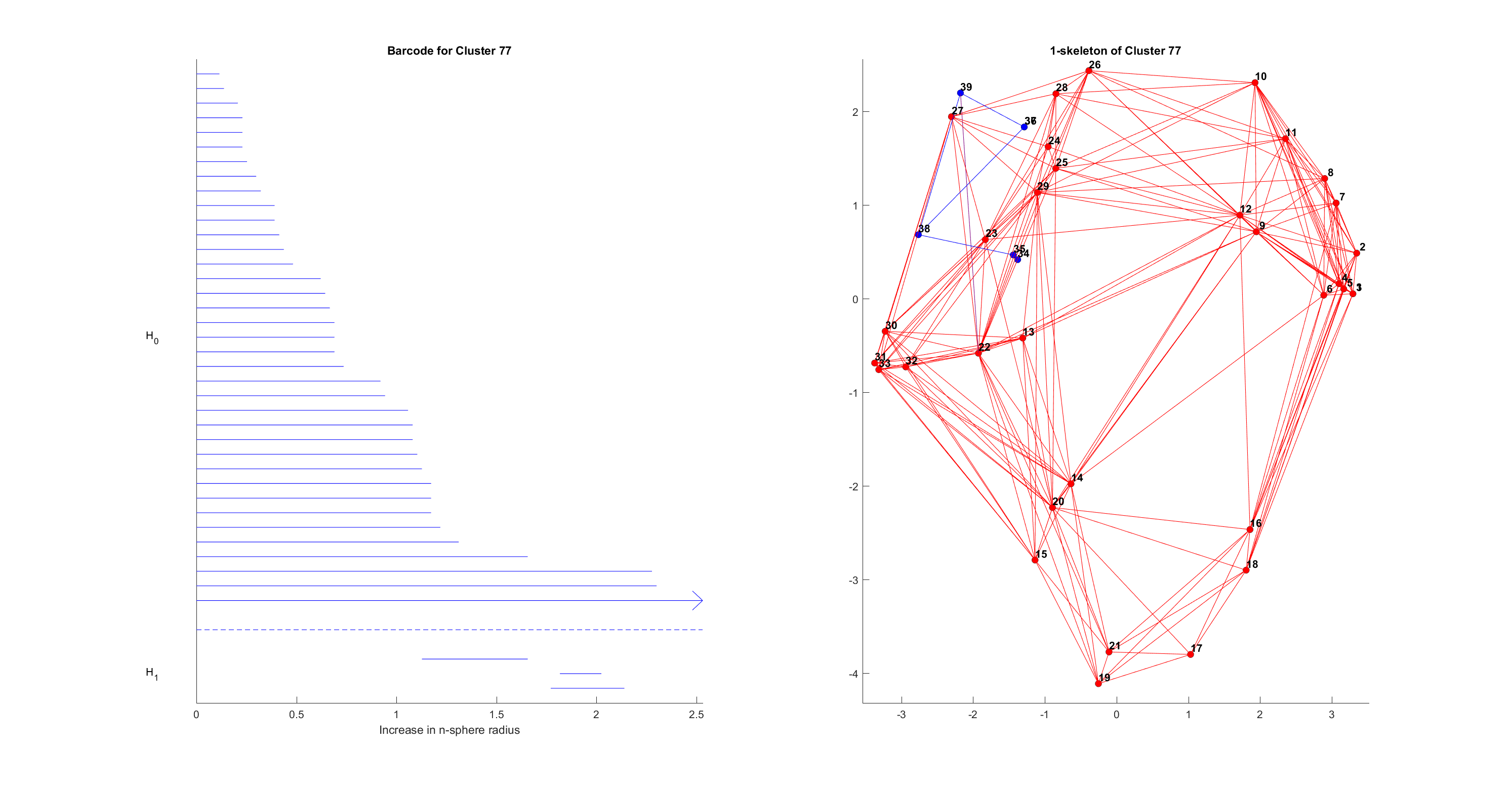}} \\
	\caption{Persistent $H_1$-structures in the LanGeLin data, PCA $60\%$, clusters
	N.~75 and N.~77. \label{IEUALanGeLinH1Fig}}
\end{figure}

\begin{figure}   [!ht]
 	\includegraphics[width = 6.5in]{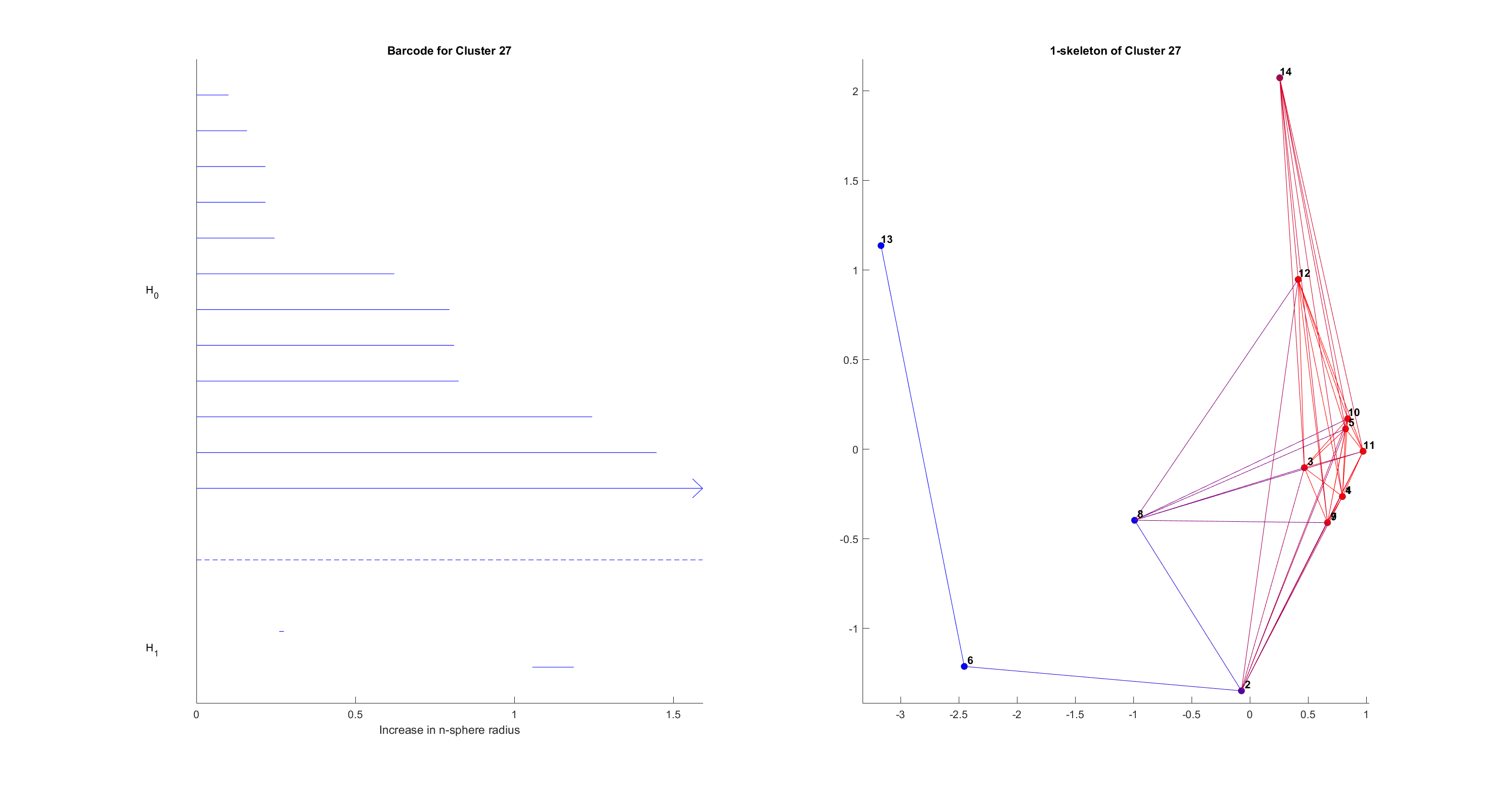} 
		\caption{Persistent $H_1$-generator in the Romance languages, LanGeLin data with 
		PCA $60\%$, cluster N.~27. \label{RomanceH1Fig}}
\end{figure}

\medskip

\subsection{Identification of data points in $H_1$ generators}

Along with computing the clusters via $H_0$ to assemble the persistent components tree, as we
discussed in the previous sections. The Perseus code also computes the persistent $H_1$ of each cluster. 
In the output we only get the birth and the death of each generator of the $H_1$ in the form of a barcode diagram, but the persistent homology computation does not provide us with a specific choice of generators themselves for the persistent $H_1$ homology at each scale radius. 

\smallskip

In particular, because no dimensional structure is detected in the data, a simple measure of the topological \textit{significance} of a cluster is provided by the sum of the lengths of its persistent $H_1$ generators. Because of the way the tree is constructed using inclusions of clusters, this measure of topological significance monotonically increases while going up the tree from child to parent. 

\smallskip

The natural question to ask then is when the first topologically significant cluster in a tree arises, that is, the first cluster that exhibits a non-trivial persistent $H_1$, and which clusters contribute to the formation of the first non-trivial $H_1$-generators. 

\smallskip

In order to compute manually the generator of the significant $H_1$ we need to identify each group of the languages in these cycles, remove each of these language groups from our data set and compute the persistent homology again to see whether the significant structure in the $H_1$ still exists. 
The method we use to identify explicit generators of the persistent $H_1$ consists of the following procedure:
\begin{enumerate}
\item identify the first cluster in the persistent components tree where the new $H_1$-generator appears,
\item list the languages that are added in passing from the previous cluster (the last one without the new
generator) to the new one,
\item identify all the new cycles that are added in the $1$-skeleton of the new cluster that were not
present in the $1$-skeleton of the previous cluster,
\item in turn remove the languages belonging to one of the new cycles and recompute the persistent
topology of the remaining set,
\item check if the new $H_1$-generator is still present after the removal or not.
\end{enumerate}
If more than one of the new cycles causes the disappearance of the $H_1$-generator then those
cycles are homologous and either one can be chosen as generator. If the removal of a cycle does
not cause the $H_1$-generator to disappear, then that cycle is a boundary in the Vietoris--Rips
complex and does not determine a non-trivial homology class.  Another possibility, if the removal of
a cycle does not eliminate the $H_1$-generator is that there are more than one homologous
cycles that represent the same $H_1$-class and eliminating one of them will still leave another
non-trivial homologous cycle. We will discuss these possibility in explicit examples below. 

\begin{figure}   [!ht]
 	\subfloat[Niger-Congo, cluster N.~85]{\includegraphics[width = 6.6in]{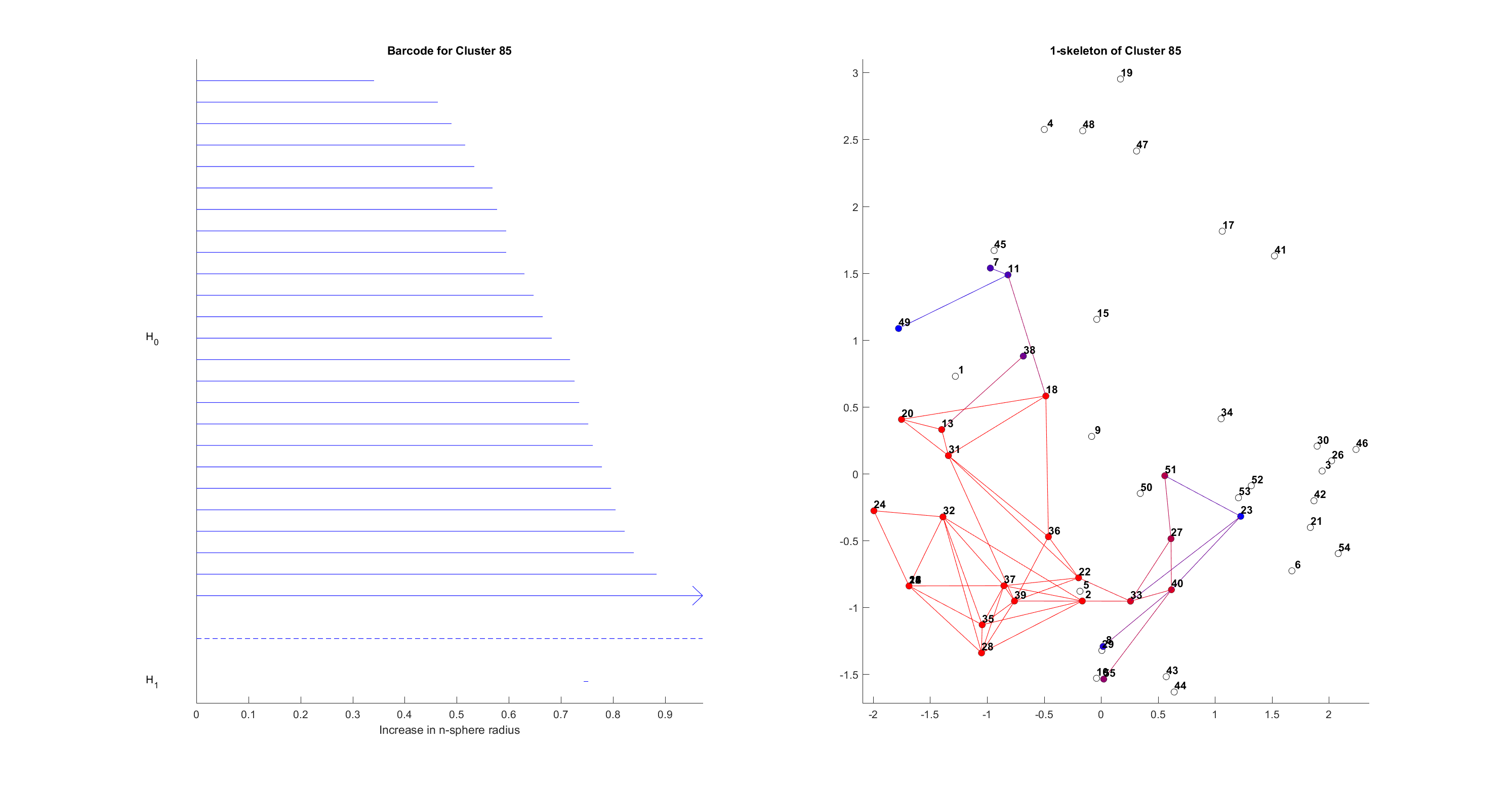}} \\
	\subfloat[Niger-Congo, cluster N.~89]{\includegraphics[width = 6.5in]{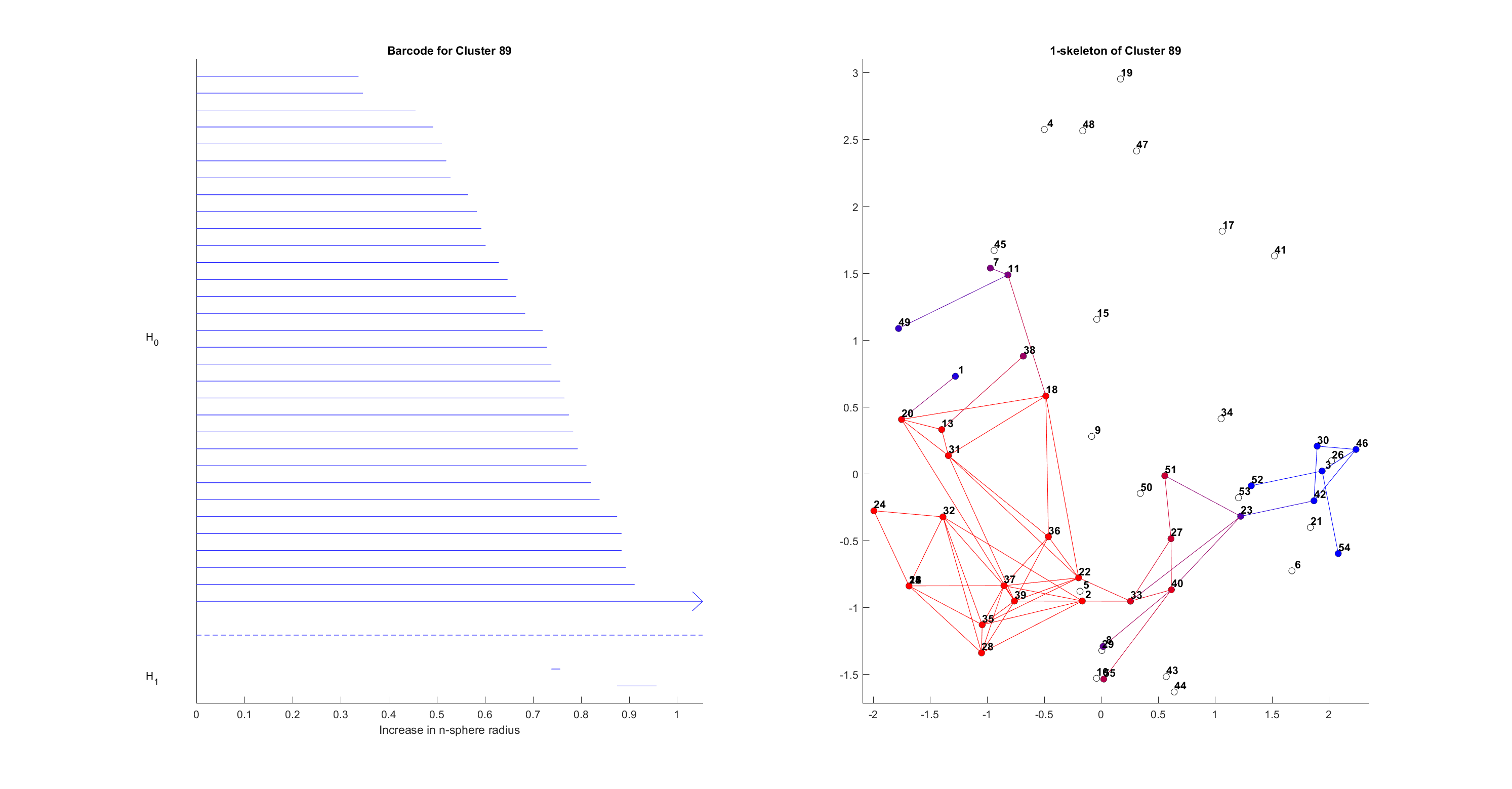}} \\
	\caption{Persistent $H_1$-structures in the Niger-Congo families, SSWL data, PCA $60\%$, clusters
	N.~85 and N.~89. \label{SSWLNCLoopFig}}
\end{figure}

%\begin{figure}   [!ht]
% 	\subfloat[Niger-Congo, cluster N.~60]{\includegraphics[width = 6.6in]{NC-figure_cluster_60.png}} \\
%	\subfloat[Niger-Congo, cluster N.~75]{\includegraphics[width = 6.5in]{NC-figure_cluster_75.png}} \\
%	\caption{Persistent $H_1$-structures in the Niger-Congo families, SSWL data, PCA $60\%$, clusters
%	N.~60 and N.~75. \label{SSWLNCLoopFig}}
%\end{figure}

\begin{figure}  	 [!ht]
	\subfloat[Niger-Congo, cluster N.~99]{\includegraphics[width = 6.5in]{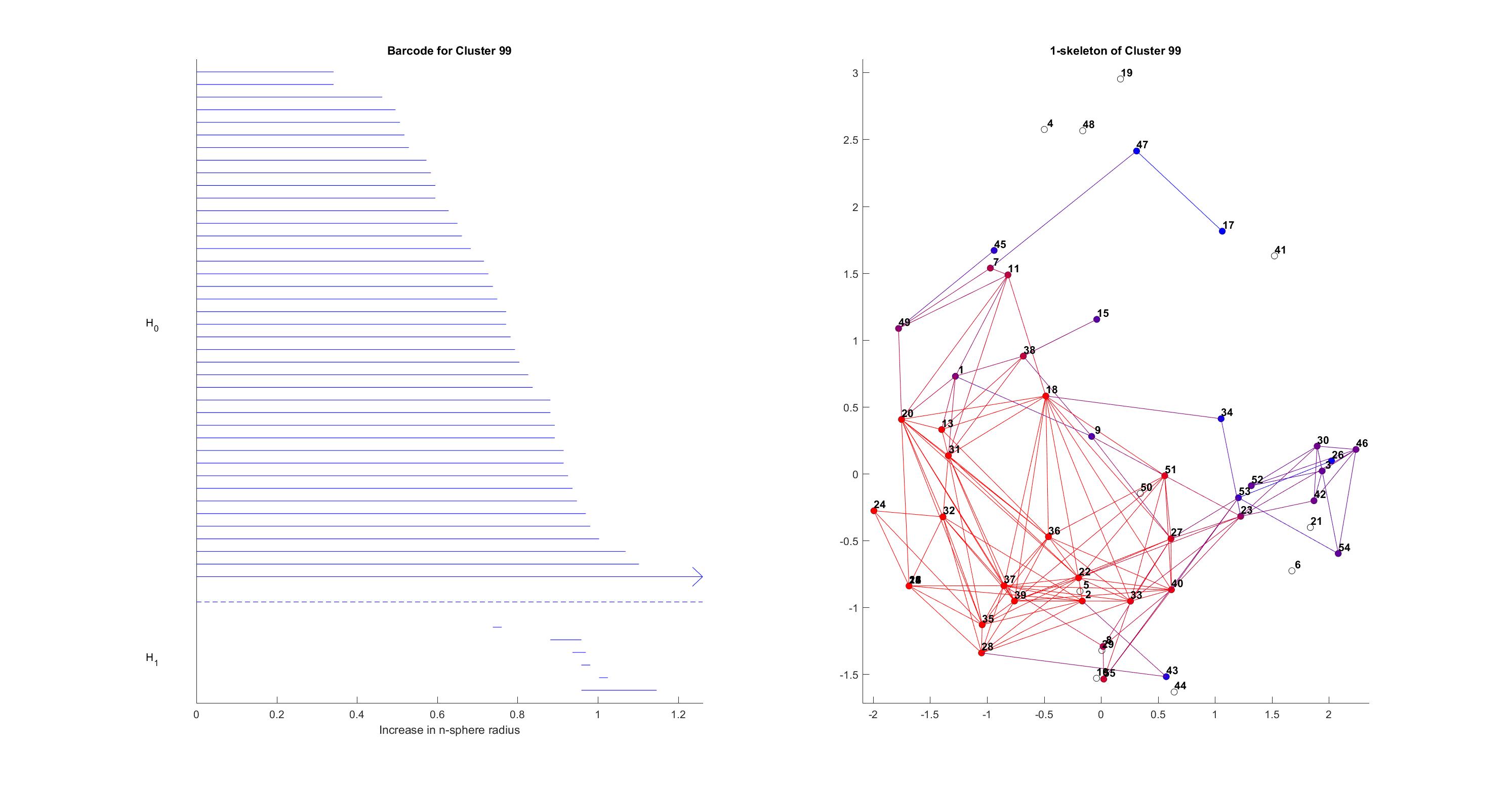}} \\
	\subfloat[Niger-Congo, cluster N.~109]{\includegraphics[width = 6.6in]{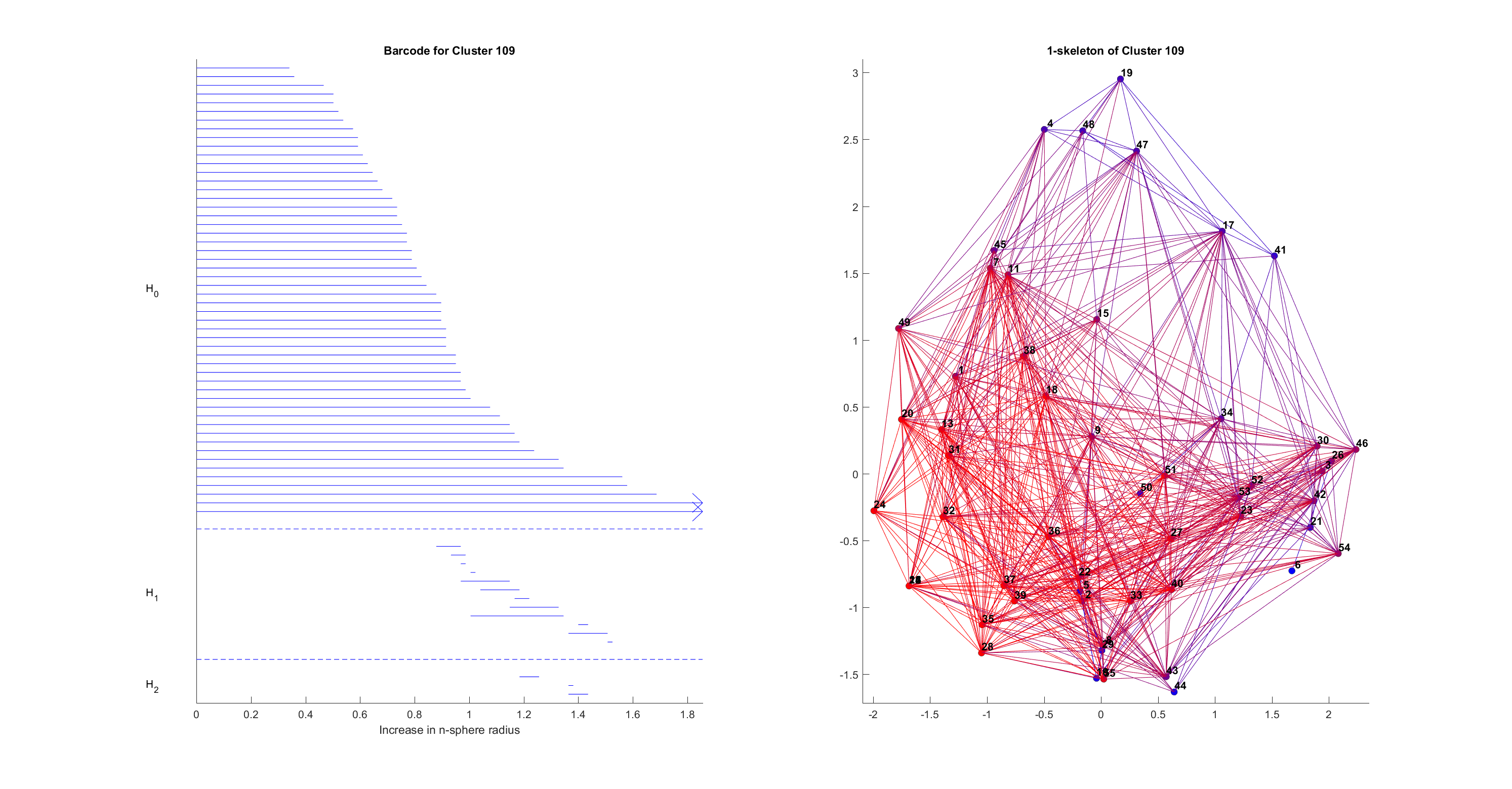}} 
	\caption{Persistent $H_1$-structures in the Niger-Congo families, SSWL data, PCA $60\%$, 
	clusters N.~99 and N.~109. \label{SSWLNCLoopFig2}}
\end{figure}
	
%\begin{figure}  	 [!ht]
%	\subfloat[Niger-Congo, cluster N.~82]{\includegraphics[width = 6.5in]{NC-figure_cluster_82.png}} \\
%	\subfloat[Niger-Congo, cluster N.~89]{\includegraphics[width = 6.6in]{NC-figure_cluster_89.png}} 
%	\caption{Persistent $H_1$-structures in the Niger-Congo families, SSWL data, PCA $60\%$, 
%	clusters N.~82 and N.~89. \label{SSWLNCLoopFig2}}
%\end{figure}

\begin{figure}  [!ht]
	\subfloat[Afro-Asiatic]{\includegraphics[width = 6.5in]{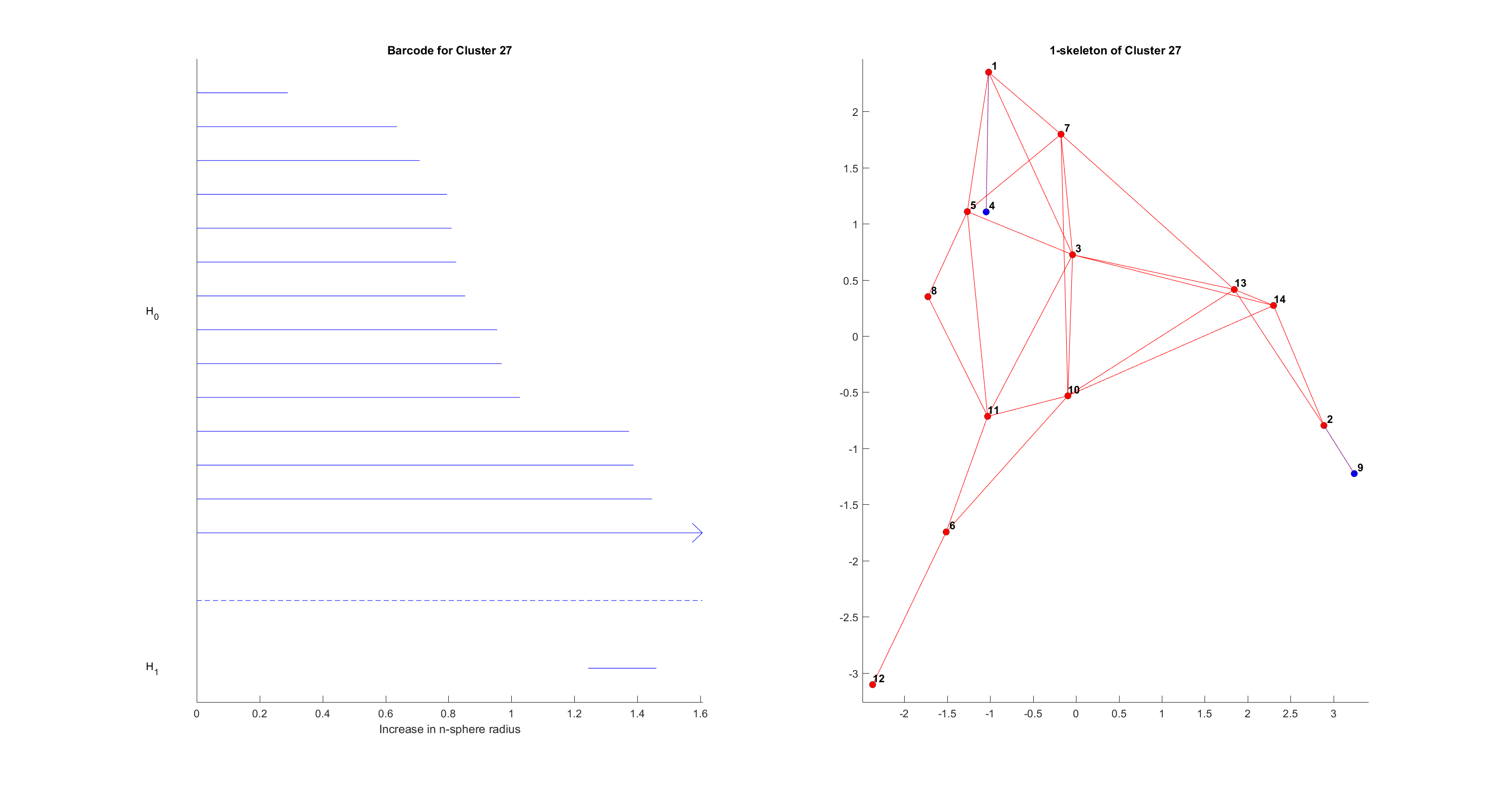}} \\
	\subfloat[Austronesian]{\includegraphics[width = 6in]{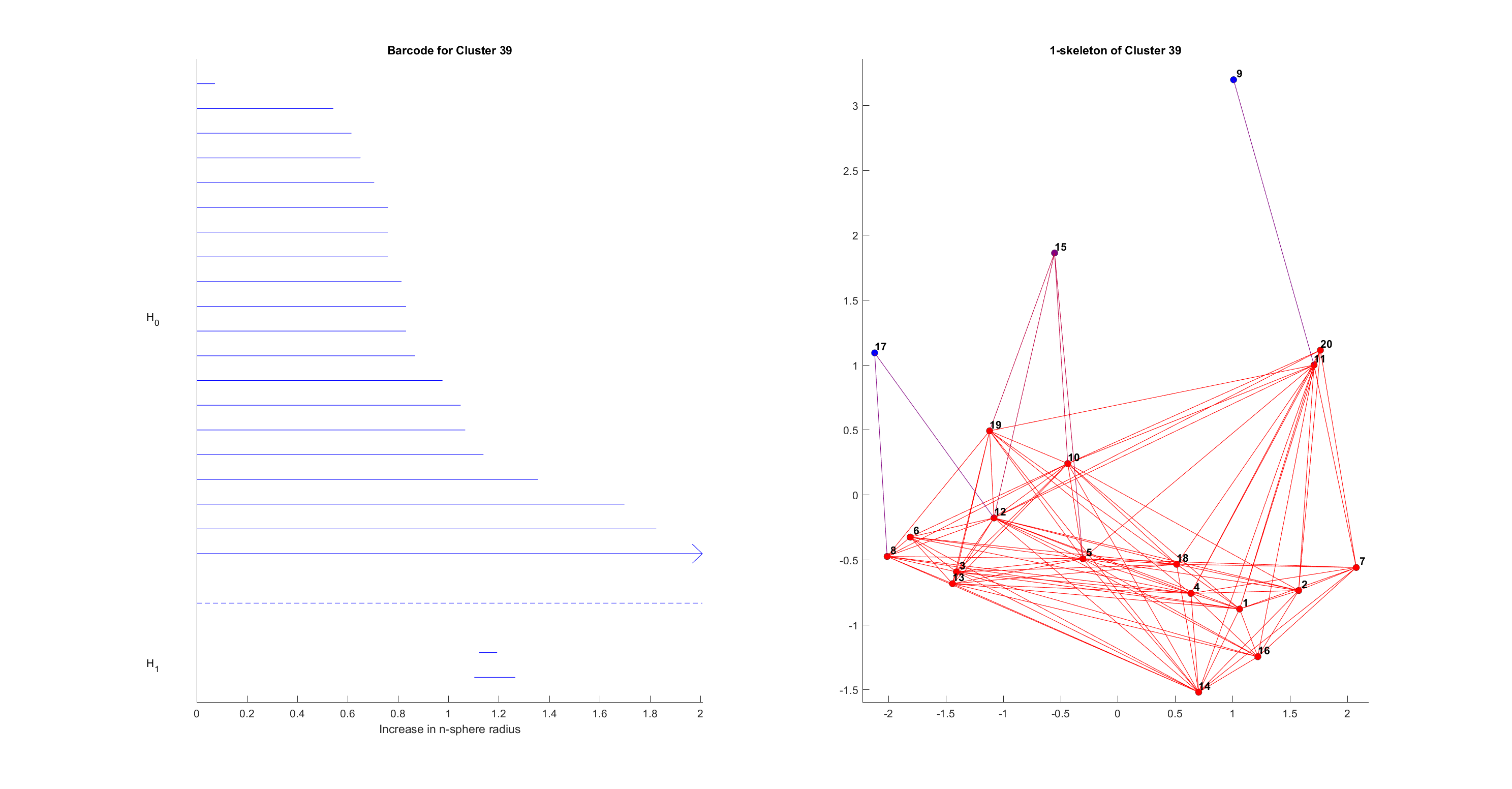}} \\
	\caption{Persistent topology in the Afro-Asiatic and Austronesian language families, SSWL data, PCA $60\%$.  \label{SSWLfamiliesLoopFig}}
	\end{figure}

%\begin{figure}  [!ht]
%	\subfloat[Afro-Asiatic]{\includegraphics[width = 6.5in]{AA-figure_cluster_29.png}} \\
%	\subfloat[Austronesian]{\includegraphics[width = 6in]{Austro-figure_cluster_35.png}} \\
%	\caption{Persistent topology in the Afro-Asiatic and Austronesian language families, SSWL data, PCA $60\%$.  \label{SSWLfamiliesLoopFig}}
%	\end{figure}

\smallskip
\subsection{Example: homoplasy phenomena}\label{HomoplSec}

One particularly interesting example is in cluster $141$ of the SSWL Indoeuropean data (see Figure~\ref{clusters_SSWL}). The large cluster N.~$140$ has no significant $H_1$ generator. However, the language Brazilian Portuguese is added in going from cluster N.~$140$ to Cluster N.~$141$, and in doing so it makes $141$ the smallest cluster with an $H_1$ generator of note. The natural question when looking at this example is what is the loop that gives the new non-trivial $H_1$-generator starting in Cluster N.~$141$ and how do we detect it. 

\smallskip

Using the method discussed above for identifying explicit $H_1$-generators, we see that
the addition of the single new language Brazilian Portuguese in the transition from cluster N.~$140$ to cluster N.~$141$ causes a change to the $1$-skeleton of the Vietoris--Rips complex, which consists of the addition of
three new cycles:  \{English, Swedish, European Portuguese\}, \{Czech, Lithuanian, Middle Dutch, Swiss German\} and \{Czech, Ukranian, Slovenian, Tocharian A\}. We remove in turn each one of these cycles from the data set and
we compute the effect on the persistent $H_1$. Only in the case of the removal of the cycle consisting of
 \{Czech, Lithuanian, Middle Dutch, Swiss German\} the $H_1$-generator disappears, while it remains unchanged
 when removing one of the other two cycles. This means that the cycles \{English, Swedish, European Portuguese\}
 and \{Czech, Ukranian, Slovenian, Tocharian A\} are boundaries of $2$-chains in the Vietoris--Rips complex,
 hence they do not define non-trivial homology classes, while the cycle \{Czech, Lithuanian, Middle Dutch, Swiss German\} is not a boundary.  This leads to the conclusion that this loop is a possible generator of the non-trivial $H_1$. 
 
 \smallskip

The Germanic languages Middle Dutch and Swiss German are closely related, but the fact that they 
occur in a non-trivial loop together with the Balto-Slavic Lithuanian and Czech seems difficult to justify
in historical linguistic terms and is more likely representing a case of homoplasy detection. 

\smallskip
\subsection{Example: the Gothic--Slavic--Greek circle}
	
	We discuss another example of explicit identification of an $H_1$-generator, with the same method
	discussed above, where the resulting generator may have a historical linguistic explanation beyond
	homoplasy. 
	
	\smallskip

The first persistent $H_1$-generator that we analyze in the Indo-European family, with the LanGeLin data, is illustrated in 
Figure~\ref{GothicGreekSlavicLoopFig} which shows the persistent generator arising between cluster N.102 and cluster N.113.
The barcodes and the $1$-skeleta are illustrated in the figure. Since only one $H_1$-generator is present most loops visible in the
one skeleton are filled by $2$-simplices from the $2$-skeleton (not shown in the figures). 

\smallskip

In order to find an explicit generator we remove some of the clusters involved and see which removals cause the 
$H_1$-generator to disappear. We observe that New Testament Greek, Romeyka Pontic Greek, and Gothic are in the loop generators, as removal of
any one of them causes the generator to disappear. In \cite{Port} we had observed an $H_1$-generator in the Indo-European language family involving
the Hellenic languages and some Slavic languages, hence we expect the Slavic languages to possibly play a role in this $H_1$-generator as well.
We observe that the removal of any individual Slavic language does not cause the $H_1$-generator to disappear but the removal of all of them (which
are very closely clustered together) destroys the persistent $H_1$-generator causing the appearance of a smaller one with a much shorter length of 
persistence in the barcode diagram. This suggests a geometry of a simplex where the three nodes New Testament Greek, Romeyka Pontic Greek, and Gothic
are connected to some of the Slavic languages, which in turn are connected among themselves via $2$-simplexes. This creates some homologous
loops, so that the removal of a single Slavic language still leaves another nontrivial homologous generator while removal of the entire cluster of the
Slavic languages removes it. The new smaller generator created by the removal of the Slavic languages accounts for the arrangement of
2-simplices around them. 

\smallskip

In terms of historical linguistics the existence of a loop involving some of the Greek languages, Gothic, and some of the Slavic languages
may be explainable in terms of a combination of historical phenomena. One is influences, here seen at the syntactic level, 
between the Greek languages and South Slavic languages, see for instance \cite{MiseTo}. Another phenomenon that this non-trivial 
persistent $H_1$-generator may be capturing is the syntactic influence of New Testament Greek on Gothic, as discussed for instance in \cite{Goth},
where several calques from Greek constructions are identified in Gothic syntax. Finally, while it is known that the Gothic influence in the
Proto-Slavic borrowing was primarily a lexical phenomenon (see \cite{Holzer}, \cite{Marty}), there is an indication of morpho-syntactic borrowing as well,
see for instance \cite{Genis}. While we cannot be sure that the structure detected by the persistent topology is indeed reflecting these
historical linguistic phenomena, this is a possible tentative explanation for the existence of this non-trivial loop and the languages involved.

\smallskip
\subsection{Additional $H_1$-structures in the Indo-European family}\label{IEH1Sec}

The filtered SSWL data we considered in Section~\ref{SSWLfilterH0Sec}, for the
Indo-European languages together with the hypothetical Ural-Altaic languages, show
the first appearance of a very small (in terms of persistence interval) 
persistent $H_1$-generator at cluster N.~78, while two much more significant 
persistent $H_1$-generators arise at cluster N.~91, followed by a third one at
the top cluster, N.~95, see Figure~\ref{IEUAH1Fig}. We will discuss elsewhere the
identification of explicit representative cycles for these $H_1$-generators.

\smallskip

By comparison, the LanGeLin data set, which comprises mostly Indo-European
and Ural-Altaic languages, also has three non-trivial $H_1$-generators in the top
cluster, N.~77, with the first non-trivial one arising at cluster N.~75, and the other
two at cluster N.~76, see Figure~\ref{IEUALanGeLinH1Fig}. In this case also we 
will return to discuss elsewhere explicit cycle generators for this $H_1$-structure,
in comparison with those of the SSWL data. 

\smallskip

The individual subfamilies of the Indo-European family do not show
$H_1$-generators with long persistence, which is an indication that
the three main persistent generators describe structures that
simultaneously involve different subfamilies. Only the Romance languages
in the filtered SSWL data show a small persistent $H_1$-generator in 
cluster N.~27, Figure~\ref{RomanceH1Fig}. An explicit cycle representing
this generator will be discussed elsewhere.

\smallskip
\subsection{$H_1$-structures in other language families}\label{OtherH1Sec}

In \cite{Port} where we only analyzed some of the sub-clusters of the various language
families, we had not found significantly persistent $H_1$-structures in the sub-clusters 
we analyzed of the Niger-Congo languages. However, in the present analysis which we 
extended to all sub-clusters, we find that a first significant non-trivial persistent $H_1$-structure 
begins to occur in the Niger-Congo family at cluster N.~89, followed by a second significant
persistent generator that arises at cluster N.~99 followed by more non-trivial $H_1$-generators at N.~109.
The $1$-skeleta and barcode diagrams for these clusters are shown in Figure~\ref{SSWLNCLoopFig}
and Figure~\ref{SSWLNCLoopFig2}. A more in depth analysis of the $H_1$-structures in the
Niger-Congo language family, identifying explicit generators and investigating their possible
linguistic significance, will be conducted elsewhere. 

\medskip

The Austronesian language family exhibits two non-trivial $H_1$-generators in cluster N.~39, 
see Figure~\ref{SSWLfamiliesLoopFig}. The Afro-Asiatic family shows only some occurrence 
of a single persistent $H_1$-generator which only arises in cluster N.~27, see Figure~\ref{SSWLfamiliesLoopFig}.

\smallskip
	
	In these examples, at the stage where a new persistent $H_1$ generator arises, typically a few (sometimes just one) new languages
	are added to the cluster. This creates several new edges, which in turn add several 
	new cycles to the $1$-skeleton. We will return to
 a more detailed analysis of these cycles and an identification of an explicit generator among them. Note that, 
 as in the case discussed in Section~\ref{HomoplSec}, the new languages added at the level where the new $H_1$ generator
 appears need not themselves be part of a cycle representative of the persistent $H_1$-generator.

%\newpage

\end{document}